\documentclass[]{article}

\usepackage[sort,nocompress]{cite}
\usepackage{hyperref}
\hypersetup{
	colorlinks   = true,
	linkcolor    = blue,
	citecolor    = green
}

\usepackage[cmex10]{amsmath}
\usepackage{algorithmic}
\algsetup{linenosize=\footnotesize}

\usepackage[tight,footnotesize]{subfigure}
\usepackage{url}
\usepackage{color}
\usepackage{verbatim}
\usepackage{gensymb}
\usepackage{caption}
\usepackage{xcolor}

\usepackage{graphicx}
\usepackage{multirow}

\usepackage[ruled, vlined]{algorithm2e}
\usepackage{amssymb}
\usepackage{cleveref}

\usepackage{textcomp}
\usepackage{csquotes}

\usepackage{fullpage}

\newcommand {\uu}  { {\bf u} }

\newcommand {\bgg}  { {\bf g} }

\newcommand {\xx}  { {\bf x} }
\renewcommand {\aa}  { {\bf a} }

\newcommand {\rr}  { {\bf r} }

\newcommand {\qq}  { {\bf q} }

\newcommand {\pp}  { {\bf p} }

\newcommand {\vv}  { {\bf v} }
\newcommand {\ww}  { {\bf w} }

\newcommand {\BB}  { {\bf B} }

\newcommand {\bb}  { {\bf b} }

\newcommand {\ee}  { {\bf e} }
\newcommand {\sa}  { {\bf s} }

\renewcommand{\vec}[1]{\ensuremath{\mathbf{#1}}}

\newcommand{\A}{\vec{A}}
\renewcommand{\H}{\vec{H}}

\newcommand{\V}{\vec{V}}
\newcommand{\W}{\vec{W}}  
\newcommand{\X}{\vec{X}}

\newcommand{\BInd}{\vec{BInd}}
\newcommand{\Hq}{\vec{Hq}}
\newcommand{\Q}{\vec{Q}}

\newcommand{\defeq}{\mathrel{\mathop:}=}

\renewcommand{\Pr}{\hbox{\bf{Pr}}}

\definecolor{forestgreen}{rgb}{0.13, 0.55, 0.13}

\newcounter{comment}

\newcommand{\mP}{\text{maxPart}}
\newcommand{\xP}{\text{sumExpPart}}
\newcommand{\logP}{\text{logPart}}
\newcommand{\linP}{\text{linearPart}}

\newtheorem{remark}{Remark}

\newcommand*\lin[1]{\langle #1\rangle}

\author{
Sudhir B. Kylasa
\thanks{Elec. and Comp. Engg. Dept
Purdue Univ., W. Lafayette,
Indiana 47907, US
\tt skylasa@purdue.edu}
\and
Farbod Roosta-Khorasani
\thanks{School of Mathematics and
 Physics, Univ. of Queensland 
St Lucia, QLD 4072, Australia
\tt fred.roosta@uq.edu.au}
\and
Michael W. Mahoney
\thanks{ICSI and Department of Statistics 
Univ. of California at Berkeley 
Berkeley, CA 94720, US
\tt mmahoney@stat.berkeley.edu}
\and
Ananth Grama
\thanks{Comp. Sci. Dept
Purdue Univ., W. Lafayette,
Indiana 47907, US 
\tt ayg@cs.purdue.edu}
}

\begin{document}

\title{GPU Accelerated Sub-Sampled Newton\textsf{'}s Method}

\maketitle

\begin{abstract}
First order methods, which solely rely on gradient information, are commonly used in diverse machine learning (ML) and data analysis (DA) applications. This is attributed to the simplicity of their implementations, as well as low per-iteration computational/storage costs. However, they suffer from significant disadvantages; most notably, their performance degrades with increasing problem ill-conditioning. Furthermore, they often involve a large number of hyper-parameters, and are notoriously sensitive to parameters such as the step-size. 
By incorporating additional information from the Hessian, second-order methods, have been shown to be resilient to many such adversarial effects.
However, these advantages of using curvature information come at the cost of higher per-iteration costs, which in \enquote{big data} regimes, can be computationally prohibitive.   

In this paper, we show that, contrary to conventional belief, second-order methods, when implemented appropriately, can be more efficient than first-order alternatives in many large-scale ML/ DA applications. 
In particular, in convex settings, we consider variants of classical Newton\textsf{'}s method in which the Hessian and/or the gradient are randomly sub-sampled. We show that by effectively leveraging the power of GPUs, such randomized Newton-type algorithms can be significantly accelerated, and can easily outperform state of the art implementations of existing techniques in popular ML/ DA software packages such as TensorFlow. Additionally these randomized methods incur a small memory overhead compared to first-order methods.
In particular, we show that for million-dimensional problems, our GPU accelerated sub-sampled Newton\textsf{'}s method achieves a higher test accuracy in milliseconds as compared with tens of seconds for first order alternatives.
\end{abstract}

\section{Introduction}
\label{sec:intro}
%
%

Optimization techniques are at the core of many ML/DA applications. First-order methods that rely solely on gradient of the objective function, have been methods of choice in these applications. The scale of commonly encountered problems in typical applications necessitates optimization techniques that are \emph{fast}, i.e., have low per-iteration cost and require few overall iterations, as well as \emph{robust} to adversarial effects such as problem ill-conditioning and hyper-parameter tuning. First-order methods such as stochastic gradient descent (SGD) are widely known to have low per-iteration costs. However, they often require many iterations before suitable results are obtained, and their performance can deteriorate for moderately to ill-conditioned problems. Contrary to popular belief, ill-conditioned problems often arise in machine learning applications. For example, the ``vanishing and exploding gradient problem'' encountered in training deep neural nets \cite{bengio1994learning}, is a well-known and important issue. What is less known is that this is a consequence of the highly ill-conditioned nature of the problem. Other examples include low-rank matrix approximation and spectral clustering involving radial basis function (RBF) kernels when the scale parameter is large~\cite{gittens2016revisiting}.
A subtle, yet potentially more serious, disadvantage of most first-order methods is the large number of hyper-parameters,
as well as their high sensitivity to parameter-tuning, which can significantly slow down the training procedure and often necessitate many trial and error steps~\cite{xuNonconvexEmpirical2017,berahas2017investigation}. 

Newton-type methods use curvature information in the form of the Hessian matrix, in addition to the to gradient. This family of methods has not been commonly used in the ML/ DA community because of their high per-iteration costs, in spite of the fact that second-order methods offer a range of benefits. 
Unlike first-order methods, Newton-type methods have been shown to be highly resilient to increasing problem ill-conditioning~\cite{roosta2016sub_global,roosta2016sub_local,xu2016sub}. 
Furthermore, second-order methods typically require fewer parameters (e.g., inexactness tolerance for the sub-problem solver or line-search parameters), and are less sensitive to their specific settings~\cite{berahas2017investigation,xuNonconvexEmpirical2017}. 
By incorporating curvature information at each iteration, Newton-type methods scale the gradient such that it is a more suitable direction to follow. Consequently, although their iterations may be more expensive than those of the first-order counterparts, second-order methods typically require much fewer iterations. 

In this context, by reducing the cost of each iteration through efficient approximation of
curvature, coupled with hardware specific acceleration, one can obtain methods that are \emph{fast} and \emph{robust}.
\emph{In most ML applications, this typically translates to achieving a high test-accuracy early on in the iterative process and without significant parameter tuning}; see Section~\ref{sec:results}. This is in sharp contrast with slow-ramping trends typically observed in training with first-order methods, which is often preceded by a lengthy trial and error procedure for parameter tuning. 
Indeed, the aforementioned properties, coupled with efficiency obtained from algorithmic innovations and implementations that effectively utilize all available hardware resources, hold promise for significantly changing the landscape of optimization techniques used in ML/DA applications. 



With the long-term goal of achieving this paradigm shift,  we focus on the commonly encountered finite-sum optimization problem
\vspace{-2mm}
\begin{align}
\label{eq:obj}
\min_{\xx \in \mathbb{R}^d} F(\xx) \triangleq \sum_{i=1}^n f_i(\xx),
\end{align}
where each $f_{i}(\xx)$ is a smooth convex function, representing a loss (or misfit) corresponding to $i^{th}$ observation (or measurement)~\cite{friedman2001elements, bottou2016optimization, sra2012optimization}. 
In many ML applications, $F$ in eq.~\eqref{eq:obj} corresponds to the \emph{empirical
risk}~\cite{shalev2014understanding}, and the goal of solving eq.~\eqref{eq:obj} is
to obtain a solution with small generalization error, i.e., high predictive accuracy
on ``unseen'' data. 
We consider eq.~\eqref{eq:obj} at scale, where the values of $n$ and $d$ are large -- millions and beyond. In such
settings, the mere computation of the Hessian and the gradient of $ F $ increases
linearly in $n$. Indeed, for large-scale problems, operations on the Hessian,
e.g., matrix-vector products involved in the (approximate) solution of
the sub-problems of most Newton-type methods, typically constitute the main computational 
bottleneck.
In such cases, randomized sub-sampling has been shown to be highly successful in reducing computational and memory costs to be effectively \emph{independent} of $n$. 
For example, a simple instance of eq.~\eqref{eq:obj} is when the functions $f_i$'s are quadratics, in which case one has an over-constrained least squares problem.  For these problems, randomized numerical linear algebra (RandNLA) techniques rely on random sampling, which is used to compute a data-aware or data-oblivious subspace embedding that preserves the geometry of the entire subspace \cite{mahoney2011randomized}. 
Furthermore, non-trivial practical implementations of algorithms based on these ideas have been shown to beat state-of-the-art numerical techniques~\cite{avron2010blendenpik,meng2014lsrn,yang2016implementing}.  
For more general problems, theoretical properties of sub-sampled Newton-type methods, for both convex and non-convex problems of the form in eq.~\eqref{eq:obj}, have been recently studied in a series of efforts~\cite{roosta2016sub_global,roosta2016sub_local, xuNonconvexTheoretical2017, xu2016sub, bollapragada2016exact,byrd2012sample, erdogdu2015convergence}. 
\emph{However, for real ML/ DA applications beyond least squares, practical and hardware-specific implementations that can effectively draw upon all available computing resources, are lacking}. 

%
%


%
%
\vspace{1mm}
\noindent \textbf{Contributions:}
Our contributions in this paper can be summarized as follows: 
\emph{
Through a judicious mix of statistical techniques, algorithmic innovations, and highly optimized GPU implementations, we  develop an accelerated variant of the classical Newton's method that has low per-iteration cost, fast convergence, and minimal memory overhead. 
In the process, we show that, for solving eq.~\eqref{eq:obj}, our accelerated randomized method significantly outperforms state of the art implementations of existing techniques in popular ML/DA software packages such as TensorFlow~\cite{abadi2016tensorflow}, in terms of improved training time, generalization error, and robustness to various adversarial effects. 
}



%
%

This paper is organized as follows. Section \ref{sec:related_work} provides an overview of related literature. Section~\ref{sec:theory} presents technical background regarding sub-sampled Newton-type methods, Softmax classifier as a practical instance of eq.~\eqref{eq:obj}, along with a description of the algorithms and their implementation. Section~\ref{sec:results} compares and contrasts GPU based implementations of sub-sampled Newton-type methods with first order methods available in TensorFlow. Conclusions and avenues for future work are presented in Section~\ref{sec:conclusions}.

\section{Related Work}
\label{sec:related_work}

%
%

The class of first-order methods includes a number of techniques that are commonly used in
diverse ML/DA applications. Many of these techniques have been efficiently implemented in popular software packages. For example, TensorFlow,~\cite{abadi2016tensorflow},
has enjoyed considerable success among ML practitioners. Among
first-order methods implemented in TensorFlow for solving~\eqref{eq:obj} are
Adagrad~\cite{duchi2011adaptive}, RMSProp~\cite{tijmen2012rmsprop}, Adam~\cite{kingma2014adam},
Adadelta~\cite{zeiler2012adadelta}, and SGD with/
without momentum~\cite{sutskever2013importance}. Excluding SGD, the rest of these
methods are adaptive, in that they incorporate prior gradients to 
choose a preconditioner at each gradient step.
Through the use of gradient history from previous iterations, these adaptive methods
non-uniformly scale the current gradient to obtain an update direction that takes
larger steps along the coordinates with smaller derivatives and, conversely,
smaller steps along those with larger derivatives. At a high level, these methods
aim to capture non-uniform scaling of Newton's method, albeit, using 
limited curvature information.

Theoretical properties of a variety of randomized Newton-type methods, for both
convex and non-convex problems of the form eq.~\eqref{eq:obj}, have been recently studied
in a series of results, both in the context of ML applications~\cite{roosta2016sub_global,
roosta2016sub_local, xu2016sub, xuNonconvexEmpirical2017, xuNonconvexTheoretical2017,
bollapragada2016exact, byrd2012sample, erdogdu2015convergence}, 
as well as scientific computing applications~\cite{rodoas1, rodoas2, doas12}. 

GPUs have been successfully used in a variety of ML applications to speed up computations~\cite{coates2009scalable,
raina2009large,coates2013deep,ngiam2011optimization}. In particular, Raina et al.~\cite{raina2009large}
demonstrate that modern GPUs can far surpass the computational capabilities of multi-core CPUs,
and have the potential to address many of the computational challenges encountered
in training large-scale learning models. Most relevant to this paper, Ngiam et al.~\cite{ngiam2011optimization} show that off-the-shelf optimization methods such as Limited memory BFGS (L-BFGS) and Conjugate Gradient (CG), have the potential to
outperform variants of SGD in deep learning applications. It was further demonstrated
that the difference in performance between LBFGS/CG and SGD is more pronounced if
one considers hardware accelerators such as GPUs. Extending similar results to full-fledged second-order algorithms, 
such Newton's method, is a major motivating factor for our work here.

\section{Theory, Algorithms and Implementation Details}
\label{sec:theory}

\subsection{Notation}
\label{sec:notation}

Vectors, $ \vv $, and matrices, $ \V $, are denoted by bold lower and upper case letters, respectively. $\nabla f(\xx)$
and $\nabla^{2} f(\xx)$ represent the gradient and the Hessian of $f$ at $\xx$,
respectively. The superscript, e.g., $\xx^{(k)}$, denotes iteration count.
$\mathcal{S}$ denotes a collection of indices drawn from the set $\{1,2,\cdots,n\}$, with
potentially repeated items, and its cardinality is denoted by $|\mathcal{S}|$. Following  \texttt{Matlab} notation, $ [\vv;\ww] \in \mathbb{R}^{2p}$ denotes vertical stacking of two column vectors $ \vv,\ww \in \mathbb{R}^{p} $, whereas $ [\vv,\ww] \in \mathbb{R}^{p \times 2 }$ denotes a $ p \text{ by } 2$ matrix whose columns are formed from the vectors $ \vv $ and $ \ww $. Vector $ \ell_{2} $ norm is denoted by $ \| \xx \| $. For a boolean variable, $ x \in \{\text{True},\text{False}\} $, the indicator function $ \mathbf{1}(x) $ evaluates to one if $ x = \text{True} $, and zero otherwise. 
$ < \uu, \vv > = \uu^{T} \vv $ denotes the dot product of vectors $ \uu $ and $ \vv $, and $ \A \odot \BB $ represents 
element-wise multiplication of matrices $ \A $ and $ \BB $.

\subsection{Sub-Sampled Newton's Method}

For the optimization problem eq.~\eqref{eq:obj}, in each iteration, consider selecting two sample sets of indices from $\{1,2,\ldots,n\}$, uniformly at random \textit{with} or \textit{without} replacement. Let $\mathcal{S}_{\bgg}$ and $\mathcal{S}_{\H}$ denote the sample collections, and define $ \bgg $ and $ \H $ as
\begin{subequations}
\begin{align}
\bgg(\xx) &\triangleq \frac{n}{|\mathcal{S}_{\bgg}|} \sum_{j \in \mathcal{S}_{\bgg}} \nabla f_{j}(\xx),	\label{subsampled_G} \\	
\H(\xx) &\triangleq \frac{n}{|\mathcal{S}_{\H}|} \sum_{j \in \mathcal{S}_{\H}} \nabla^{2} f_{j}(\xx),
\label{subsampled_H}
\end{align}
\end{subequations}
to be the sub-sampled gradient and Hessian, respectively. 


It has been shown that, under
certain bounds on the size of the samples, $ |\mathcal{S}_{\bgg}| $ and
$ |\mathcal{S}_{\H}| $, one can, with high probability, ensure that $ \bgg $ and $ \H $ are ``suitable'' approximations to the full gradient and Hessian, in an
algorithmic sense~\cite{roosta2016sub_global,roosta2016sub_local}. For each iterate $ \xx^{(k)} $, using the corresponding sub-sampled
approximations of the full gradient,
$\bgg(\xx^{(k)})$, and the full Hessian, $\H(\xx^{(k)})$, we consider \emph{inexact} Newton-type iterations of the form 
\begin{subequations}
\label{eq:newton_cg_iterations}
\begin{align}
\label{eq:update}
\xx^{(k+1)} = \xx^{(k)} + \alpha_{k} \pp_{k},
\end{align}
where $ \pp_{k} $ is a search direction satisfying
\begin{align}
\label{eq:inexact}
\| \H(\xx^{(k)})\pp_{k} + \bgg(\xx^{(k)})\| \leq \theta \|\bgg(\xx^{(k)})\|,
\end{align}
for some inexactness tolerance $ 0 < \theta < 1 $ and $\alpha_{k}$ is the largest $\alpha \leq 1$ such that
\begin{align}
F(\xx^{(k)} + \alpha \pp_{k}) \leq F(\xx^{(k)}) + \alpha \beta \pp_{k}^{T} \bgg(\xx^{(k)}),
\label{eq:armijo}
\end{align}
for some $\beta \in (0,1)$. 
\label{eq:inexact_newton_itrs}	
\end{subequations}
The requirement in eq.~\eqref{eq:armijo} is often referred to as Armijo-type
line-search~\cite{nocedal2006numerical}, and \cref{eq:inexact} is the
$\theta$-relative error approximation condition of the exact solution to the linear system
\begin{align}
\label{eq:exact}
\H(\xx^{(k)}) \pp_{k} &= -\bgg(\xx^{(k)}),
\end{align} 
which is similar to that arising in classical Newton's Method. Note that in (strictly)
convex settings, where the sub-sampled Hessian matrix is symmetric positive definite (SPD),
conjugate gradient (CG) with early stopping can be used to obtain an approximate solution
to eq.~\eqref{eq:exact} satisfying eq.~\eqref{eq:inexact}. It has also been shown~\cite{roosta2016sub_global, roosta2016sub_local},  that to inherit the convergence properties of the, rather expensive, algorithm that employs the exact solution to eq.~\eqref{eq:exact}, the inexactness tolerance, $ \theta $, in eq.~\eqref{eq:inexact} can only be chosen in the
order of the inverse of the \emph{square root} of the problem condition number. As a
result, even for ill-conditioned problems, only a relatively moderate tolerance for
CG ensures that we indeed maintain
convergence properties of the exact update (see also examples in Section~\ref{sec:results}).
Putting all of these together, we obtain Algorithm~\ref{alg:ssn}, which under specific
assumptions, has been shown~\cite{roosta2016sub_global, roosta2016sub_local} to be globally
linearly convergent\footnote{It converges linearly to the optimum starting from any
initial guess $ \xx^{(0)} $.} with problem-independent local convergence rate
\footnote{If the iterates are close enough to the optimum, it converges with a constant linear rate independent of the problem-related quantities.}.

\begin{algorithm} [!htb]
	\caption{Sub-Sampled Newton Method}
	\label{alg:ssn}
	\SetAlgoLined
	\SetKwInOut{Input}{Input}
	\SetKwInOut{Parameter}{Parameters}
	\Input{Initial iterate, $\xx^{(0)}$}
	\Parameter{$0 < \epsilon, \beta, \theta < 1$}
	\lnl{}\ForEach{$k = 0,1,2,\ldots$}{
		\lnl{} Form $\bgg(\xx^{(k)})$ as in eq.~\eqref{subsampled_G} \\
		\lnl{} Form $\H(\xx^{(k)})$ as in eq.~\eqref{subsampled_H} \\
		\lnl{} \If{$\|\bgg(\xx^{(k)})\| < \epsilon$}{
			STOP
		}
		\lnl{} Update $\xx^{(k+1)}$ as in eq.~\eqref{eq:inexact_newton_itrs}\\
	}
	\label{alg:cg}
\end{algorithm}

\subsection{Multi-Class classification}
\label{sec:multi_class}
For completeness, we now briefly review multi-class classification using softmax and cross-entropy loss function,
as an important instance of the problems of the form described in eq.~\eqref{eq:obj}. Consider a $ p $
dimensional feature vector $ \aa $, with corresponding labels $ b $, which can belong
to one of $C$ classes. In such a classifier, the probability that $\aa$ belongs to a
class $c \in \{1,2,\ldots,C\}$ is given by $ \Pr\left(b = c \mid \aa,\ww_{1},\ldots, \ww_{C}\right) = {e^{\lin{\aa,\ww_{c}}}}/{\sum_{c' = 1}^{C} e^{\lin{\aa, \ww_{c'}}}} $,
where $ \ww_{c} \in \mathbb{R}^{p}$ is the weight vector corresponding to class $ c $.
Since probabilities must sum to one, there are in fact only $ C-1 $ degrees of freedom. Consequently, by defining $ \xx_c \triangleq \ww_c - \ww_C, \; c=1,2,\ldots,C-1 $, for training data
$\{\aa_{i},b_{i}\}_{i=1}^{n} \subset \mathbb{R}^{p} \times \{1,\ldots,C\}$,
the cross-entropy loss function for $ \xx = [\xx_{1}; \xx_{2}; \ldots; \xx_{C-1}] \in \mathbb{R}^{(C-1)p} $
can be written as
\begin{align}
\label{eq:softmax_log_likelihood}
F(\xx) \triangleq & F(\xx_{1},\xx_{2},\ldots,\xx_{C-1}) \nonumber \\
= &\sum_{i=1}^{n} \left(\log \left(1+\sum_{c' = 1}^{C-1} e^{\lin{\aa_{i}, \xx_{c'}}}\right)  - \sum_{c = 1}^{C-1}\mathbf{1}(b_{i} = c) \lin{\aa_{i},\xx_{c}}\right).
\end{align} 
Note that here, $ d = (C-1) p $. It then follows that the full gradient of $F$ with respect to $\xx_c$ is
\begin{align}
\label{eq:softmax_gradient}
\nabla_{\xx_{c}} F(\xx) = \sum_{i=1}^{n} \left(\frac{e^{\lin{\aa_{i}, \xx_{c}}}}{1+\sum_{c' = 1}^{C-1} e^{\lin{\aa_{i}, \xx_{c'}}}}  - \mathbf{1}(b_{i} = c) \right) \aa_{i}.
\end{align}
Similarly, for the full Hessian of $ F $, we have
\begin{subequations}
	\label{eq:softmax_hessian}
	\begin{flalign}
	& \nabla^{2}_{\xx_{c},\xx_{c}} F = \nonumber \\ 
	& \sum_{i=1}^{n} \left(\frac{e^{\lin{\aa_{i}, \xx_{c}}}}{1+\sum_{c' = 1}^{C-1} e^{\lin{\aa_{i}, \xx_{c'}}}} - \frac{e^{2\lin{\aa_{i}, \xx_{c}}}}{\left(1+\sum_{c' = 1}^{C-1} e^{\lin{\aa_{i}, \xx_{c'}}}\right)^{2}} \right) \aa_{i} \aa_{i}^{T},  \label{eq:softmax_hessian_wc_wc}
	\end{flalign}
	and for $ \hat{c} \in \{1,2,\ldots,C-1\} \setminus \{c\} $, we get
	\begin{flalign}
	& \nabla^{2}_{\xx_{c},\xx_{\hat{c}}} F = \sum_{i=1}^{n} \left(- \frac{e^{\lin{\aa_{i}, \xx_{\hat{c}} + \xx_{c}}}}{\left(1+\sum_{c' = 1}^{C-1} e^{\lin{\aa_{i}, \xx_{c'}}}\right)^{2}} \right) \aa_{i} \aa_{i}^{T} 	\label{eq:softmax_hessian_wc_wb}. 
	\end{flalign}
\end{subequations}
Sub-sampled variants of the gradient and Hessian are obtained similarly. 
Finally, after training phase, a new data $ \aa $ is classified as 
\begin{align*}
b = \arg \max &\left\{\left\{\frac{e^{\lin{\aa,\xx_{c}}}}{\sum_{c' = 1}^{C-1} e^{\lin{\aa, \xx_{c'}}}}\right\}_{c=1}^{C-1}, 1- \frac{e^{\lin{\aa,\xx_{1}}}}{\sum_{c' = 1}^{C} e^{\lin{\aa, \xx_{c'}}}}  \right\}.
\end{align*}
\subsubsection{Numerical Stability}
\label{sec:numerical_stability}

To avoid over-flow in the evaluation of exponential functions in~\eqref{eq:softmax_log_likelihood}, we use the ``Log-Sum-Exp'' trick~\cite{murphy2012machine}. Specifically, for each data point $\aa_{i}$, we first find
the maximum value among $\lin{\aa_{i},\xx_{c}}, \; c=1,\ldots, C-1$. Define
\begin{align}
\label{eq:max_x}
M(\aa) = \max \Big\{ 0, \lin{\aa,\xx_{1}}, \lin{\aa,\xx_{2}}, \ldots, \lin{\aa,\xx_{C-1}} \Big\}, 
\end{align}
and 
\begin{align}
\label{eq:log_sum_trick}
\alpha(\aa) \defeq e^{-M(\aa)}+\sum_{c' = 1}^{C-1} e^{\lin{\aa, \xx_{c'}} - M(\aa)}.
\end{align}
Note that $M(\aa) \geq0, \alpha(\aa) \geq 1$. Now, we have $ 1+\sum_{c' = 1}^{C-1} e^{\lin{\aa_{i}, \xx_{c'}}}  = e^{M(\aa_{i})} \alpha(\aa_{i})$.
For computing~\eqref{eq:softmax_log_likelihood}, we use $\log \left(1+\sum_{c' = 1}^{C-1} e^{\lin{\aa_{i}, \xx_{c'}}}\right) = M(\aa_{i}) + \log \big( \alpha(\aa_{i}) \big)$. 
Similarly, for~\eqref{eq:softmax_gradient} and~\eqref{eq:softmax_hessian}, we use
\begin{align*}
\frac{e^{\lin{\aa_{i}, \xx_{c}}}}{1+\sum_{c' = 1}^{C-1} e^{\lin{\aa_{i}, \xx_{c'}}}} 
&= \frac{e^{\lin{\aa_{i}, \xx_{c}}-M(\aa_{i})}}{\alpha(\aa_{i})}.
\end{align*}

Note that in all these computations, we are guaranteed to have all the exponents appearing
in all the exponential functions to be negative, hence avoiding numerical over-flow.

\subsubsection{Hessian Vector Product}
\label{sec:MVP}

Given a vector $\vv \in \mathbb{R}^{d}$, we
can compute the Hessian-vector product without explicitly forming the Hessian.
For notational simplicity, define 
\begin{align*}
h(\aa,\xx) \defeq \frac{e^{\lin{\aa, \xx}-M(\xx)}}{\alpha(\aa)}, 
\end{align*}
where $M(\xx)$ and $ \alpha(\xx) $ were defined in eqs.~\eqref{eq:max_x} and~\eqref{eq:log_sum_trick}, respectively.
Now using matrices 
\begin{align*}
\stepcounter{equation}\tag{\theequation}\label{eq:matrix-a}
\V = \begin{bmatrix}
\lin{\aa_{1},\vv_{1}} & \lin{\aa_{1},\vv_{2}} & \dots  & \lin{\aa_{1},\vv_{C-1}} \\
\lin{\aa_{2},\vv_{1}} & \lin{\aa_{2},\vv_{2}} & \dots  & \lin{\aa_{2},\vv_{C-1}} \\
\vdots & \vdots & \ddots & \vdots \\
\lin{\aa_{n},\vv_{1}} & \lin{\aa_{n},\vv_{2}} & \dots  & \lin{\aa_{n},\vv_{(C-1)}}
\end{bmatrix}_{n \times (C-1)}, 
\end{align*}
and
\begin{align*}
\stepcounter{equation}\tag{\theequation}\label{eq:matrix-b}
\W = \begin{bmatrix}
h(\aa_{1},\xx_{1}) & h(\aa_{1},\xx_{2}) & \dots  & h(\aa_{1},\xx_{C-1}) \\
h(\aa_{2},\xx_{1}) & h(\aa_{2},\xx_{2}) & \dots  & h(\aa_{2},\xx_{C-1}) \\
\vdots & \vdots & \ddots & \vdots \\
h(\aa_{n},\xx_{1}) & h(\aa_{n},\xx_{2}) & \dots  & h(\aa_{n},\xx_{C-1}) \\
\end{bmatrix}_{n \times (C-1)},
\end{align*}
we compute
\begin{align*}
\stepcounter{equation}\tag{\theequation}\label{eq:matrix-c}
\vec{U} = \V \odot \W - \W \odot \Big( \big(\left(\V \odot \W\right) \ee \big) \ee^{T} \Big), 
\end{align*}
to get
\begin{align*}
\stepcounter{equation}\tag{\theequation}\label{eq:hessianvec}
\H\vv = \text{vec} \left( \A^{T} \vec{U} \right), 
\end{align*}
where $ \vv = [\vv_{1};\vv_{2};\ldots;\vv_{C-1}] \in \mathbb{R}^{d}$, $ \vv_{i} \in \mathbb{R}^{p}, i=1,2,\ldots,C-1 $, $\ee \in \mathbb{R}^{C-1}$ is a vector of all $1$'s, and each row of the matrix
$ \A \in \mathbb{R}^{n \times p}$ is a row vector corresponding to the $i^{th}$ data point,
i.e, $\A^{T} = \begin{bmatrix} \aa_{1}, \aa_{2}, \ldots, \aa_{n} \end{bmatrix}$.

\begin{remark}
Note that the memory overhead of our accelerated randomized sub-sampled Newton's method is determined by matrices  
$ \vec{U} $, $ \V$, and $ \W $, whose sizes are dictated by the Hessian sample size, $ |\mathcal{S}_{\H}| $, which is much less than $n$. This small memory overhead enables our Newton-type method to \textit{scale} to large problems, inaccessible to traditional second order methods.
\end{remark}

\subsection{Implementation Details}



We present a brief overview of the algorithmic machinery involved 
in the implementation of iterations described in eq.~\eqref{eq:newton_cg_iterations} and applied to the function defined in eq.~\eqref{eq:softmax_log_likelihood} with an added $ \ell_{2} $ regularization term, i.e., $ F(\xx) + \lambda \|\xx\|^{2}/2 $. Here, $ \lambda $ is the regularization parameter. We note that for all the algorithms in this section, we assume that matrices are stored in column-major ordering. 

\paragraph{Conjugate Gradient}

\begin{algorithm} [!htb]
\caption{Conjugate-Gradient}
\label{alg:cg}
	\SetAlgoLined
	\SetKwInOut{Input}{Input}
	\SetKwInOut{Parameter}{Parameters}
	\Input{\\
		 $ H(.) $ - Pointer to Algorithm~\ref{cuda:hx} to compute \\ Hessian-vector product, $ H(\vv) = \H \vv $ \\ 
			$ \bgg $ - Gradient}
	\Parameter{
		\\ $\theta$ - Relative residual tolerance \\
				$T$ - Maximum no. of iterations }
	\KwResult{$ \pp_{\text{best}}$, an approximate solution to $ \H \pp = -\bgg $ }
	\lnl{cg:1}$ \pp_{0} = 0 $ \\
	\lnl{cg:2}$ \rr_{0} = -\bgg$ \tcp{initial residual vector}
	\lnl{cg:3}$ \sa_{0} = \rr_{0} $ \tcp{initial search direction}
	\lnl{cg:4}$\pp_{\text{best}} = \sa_{0}$ \tcp{best solution so far}
	\lnl{cg:5}$\rr_{\text{best}} = \rr_{0} $ \\
	\lnl{cg:6}\ForEach{$k = 0,1,\ldots,T$}{
			\lnl{cg:7} $ \alpha_k = \rr_k^T \rr_k/\sa_k^T H(\sa_k)$ \\
			\lnl{cg:8} $ \pp_{k+1} = \pp_{k} + \alpha_k \sa_k $ \\
			\lnl{cg:9} $ \rr_{k+1} = \rr_{k} - \alpha_k  H(\sa_k) $ \\
			\lnl{cg:10}\If{$ \|\rr_{k+1}\|\le  \|\rr_{\text{best}}\|$}{
				$ \rr_{\text{best}} =  \rr_{k+1}$ \\
				$ \pp_{\text{best}} = \pp_{k+1} $\\
			}
			\lnl{cg:11}\If{ $ \|\rr_{k+1}\| \le \theta\|\bgg\|$ }{
				break \\
			}
			\lnl{cg:12} $ \sa_{k+1} = \rr_{k+1} +  \frac{\parallel \rr_{k+1} \parallel_{2}^{2}}{\parallel \rr_{k} \parallel_{2}^{2}} \sa_{k} $ \\
	}
\end{algorithm}

For the sake of self-containment, in Algorithm \ref{alg:cg}, we depict a slightly modified implementation of the classical CG, to approximately solve the linear system in eq.~\eqref{eq:exact}, i.e., $ \H \pp = -\bgg $, to satisfy eq.~\eqref{eq:inexact}. This routine takes a function (pointer), $H(.)$, which computes the matrix-vector product as $ H(\vv) = \H \vv $, as well as
the right-hand side vector, $\gg$. Lines~\ref{cg:2}, and~\ref{cg:3} initializes
the residual vector $ \rr $, and search direction $ \sa $, respectively,
while the best residual is initialized on line~\ref{cg:5}. Iterations start on line~\ref{cg:6},
which maintains a counter for maximum allowed iterates to compute. Step-size $\alpha$ for CG iterations is
computed on line~\ref{cg:7}, which is used to update the solution vector, $\pp$ and residual
vector, $\rr$. The minor modification comes from line~\ref{cg:10}, which stores the best solution vector thus far. The termination
condition eq.~\eqref{eq:inexact} is evaluated on line~\ref{cg:11}. Finally, the search direction, $ \sa $,
is updated in line~\ref{cg:12}. 

\paragraph{Line Search method}

\begin{algorithm} [!htb]
\caption{Line Search}
\label{alg:linesearch}
	\SetAlgoLined
	\SetKwInOut{Input}{Input}
	\SetKwInOut{Parameter}{Parameters}
	\Input{  \\
		$ \xx $ - Current point \\
	         $ \pp $ - Newton's direction \\
			 $F(.) $ - Function pointer\\
			 $\bgg(\xx)$ - Gradient
			 }
	\Parameter{\\
			$\alpha$ - Initial step size \\
			$0< \beta < 1$ - Cost function reduction constant\\
			$0< \rho < 1$ - back-tracking parameter \\
			$ i_{\max} $ - maximum line search iterations 
			}
	\lnl{ls:1}$\alpha = 1$ \\
	\lnl{ls:3} $ i = 0 $ \\
	\lnl{ls:4}\While{ $ F ( \xx + \alpha  \pp) > F ( \xx ) + \alpha \beta  \pp^{T} \bgg(\xx) $ }{
		\lnl{ls:5}\If{ $ i > i_{\max} $}{
			\lnl{ls:6} break \\
		}
		\lnl{ls:7}$ i = i + 1 $\\
		\lnl{ls:8}$ \alpha \leftarrow \rho \alpha $	}
\label{alg-ls}
\end{algorithm}

We use a simple back-tracking line search, shown in Algorithm \ref{alg:linesearch} for computing the step size 
in eq.~\eqref{eq:armijo}. Step size, $\alpha$, is initialized in line~\ref{ls:1}, which is typically set to the ``natural'' step-size of Newton's method, i.e., $ \alpha = 1 $. Iterations start at line~\ref{ls:4} by checking 
the exit criteria, and if required, successively decreasing the step size until the ``loose"
termination condition is met. In each of these iterations, if the objective function does
not reduce by a specified amount, $\beta$, step size is reduced by a fraction, $\rho$, of its
current value, until the termination condition is met or specified iterations have been exceeded. It
has been shown~\cite{roosta2016sub_global} that this process will terminate after a
certain number of iterations, i.e., we are always guaranteed to have $ \alpha \geq \alpha_{0} > 0 $
for some fixed $ \alpha_{0} $.
\paragraph{CUDA utility functions}

\begin{algorithm} [!htb]
\caption{ComputeExp}
\label{cuda:compute-exp}
	\SetAlgoLined
	\SetKwInOut{Input}{input}
	\SetKwInOut{Output}{output}
	\SetKwData{IDX}{idx}
	\SetKwData{WarpID}{warp-id}
	\SetKwData{LaneID}{lane-id}
	\SetKwData{BlockID}{block-id}
	\SetKwData{IPart}{\linP}
	\SetKwData{MA}{\mP}
	\SetKwData{EXP}{\xP}
	\Input{
		$\hat{\A}$ - where  $\hat{\A}_{i,j}$ = $ \aa_{i}^{T} \xx_{j}, \forall i \in \{1 \ldots n \}, \forall j \in \{1 \ldots C-1 \}$ \\
		$ \bb $ - Training classes \\
		\mP - memory pointer to store eq.~\eqref{m-part} \\
		\xP - memory pointer to store eq.~\eqref{exp-part} \\
		\linP - memory pointer to store eq.~\eqref{i-part}\\
		n - no. of rows in $\hat{\A}$ \\
		C - no. of classes 
	}
	\Output{
		\MA, \EXP, \IPart
	}
	\BlankLine
	\lnl{compute-exp:1}Init. \IDX \tcp*{thread-id}
	\If{\IDX $<$ n}{
		\lnl{compute-exp:2} i $\leftarrow$ \IDX \% n \tcp*{row no.}
		\lnl{compute-exp:3} $\MA_{i}$ = $\IPart_{i}$ = $\EXP_{i}$ = 0 \\
		\lnl{compute-exp:4} \ForEach{ $j$ in  $ 1 : C-1$}{
			\If {$\MA_{i} < \hat{\A}_{i,j}$ }{
				$\MA_{i}$ = $\hat{\A}_{i,j}$ \\
			}
		 }
		\lnl{compute-exp:5} \ForEach{ $j$ in  $ 1 : C-1$}{
			\lnl{compute-exp:6}\If{$\bb_{i} ==$ j }{
				$\IPart_{i}$ = $\hat{\A}_{i,j}$  \\
			}
			\lnl{compute-exp:7}$\EXP_{i}$ += exp ($\hat{\A}_{i,j}$ - $\MA_{i}$ ) \\
		 }
	}	
\end{algorithm}

Bulk of the work in evaluating the softmax function is done by ~\textit{ComputeExp} subroutine, 
shown in Algorithm~\ref{cuda:compute-exp}. This function takes a matrix, as an input,
and computes the following data structures: 
``$\text{\mP}_{i}$'' stores the maximum component in each of the rows of the input matrix,
``$\text{\linP}_{i}$'' stores the partial summation of the term 
$\Sigma_{j = 1}^{C-1} \mathbf{1}(\bb_{i} = j) ( \aa_{i}^{T} \xx_{j}) $, and 
``$\xP$'' stores the summation in eq.~\eqref{exp-part}.
Input matrix, $ \textit{$\hat{\A}$} \in \mathbb{R}^{n \times (C-1)} $ , is the product of $ \A $ and $ \X $ matrices, where $ \X \in \mathbb{R}^{p \times (C-1)} $ is a matrix whose $ i^{\text{th}} $ column is $ \xx_i \in \mathbb{R}^{p} $, i.e., $ \X  =  [ \xx_1, \xx_2, \ldots, \xx_{C-1}]$, and $ \A $ is as in eq.~\eqref{eq:hessianvec}.
Line \ref{compute-exp:1} initializes the \textit{idx}, thread-id of a given thread. 
In the for loop in line \ref{compute-exp:4}, we compute the maximum coordinate per row
of the input matrix, and the result is stored in array ``\mP''. Line \ref{compute-exp:5} computes
``\linP'' and ``\xP'' arrays, which are later used by functions
invoking this algorithm.

\paragraph{Softmax function evaluation}

\begin{algorithm} [!htb]
\caption{ComputeFX}
\label{cuda-fx}
	\SetAlgoLined
	\SetKwInOut{Input}{input}
	\SetKwInOut{Output}{output}
	\SetKwData{IDX}{idx}	
	\SetKwData{IPart}{\linP}
	\SetKwData{MX}{\mP}
	\SetKwData{EXP}{\xP}
	\Input{
		\A - Training features \\
		$\bb$ - Training classes \\
		$\xx$ - Weights vector \\
		$\lambda$ - Regularization \\
		n - no. of rows in \A \\
		p - no. of cols in \A \\
		C - no. of classes 
	}
	\Output{
		$F ( \xx )$ - Objective function evaluated at $ \xx $
	}
	\BlankLine
	\BlankLine
	\lnl{fx:1}Initialize \MX, \IPart, \EXP to store eqs.~\eqref{m-part}--\eqref{i-part}, \\
	\lnl{fx:1} Form $\X =  [\xx_1, \xx_2, \ldots, \xx_{C-1}]_{p \times (C-1)} $ \\
	\lnl{fx:2}$\hat{\A} = \A \times \X$ \tcp*{matrix-matrix multiplication}
	\lnl{fx:3}ComputeExp( $\hat{\A}$, \bb, \MX, \EXP, \IPart, n, C) \\
	\lnl{fx:4}Reduce( \IPart, \text{pLin}, n, $ t(z) = z$ ) \\
	\lnl{fx:5}Reduce( \MX, \text{pMax}, n, $ t(z) = z$ ) \\
	\lnl{fx:6}Reduce( \EXP, pExp, n, $ t(z) = z$ ) \\
	\lnl{fx:7}temp $\leftarrow$ \MX + \EXP \\
	\lnl{fx:8}Reduce( temp, \text{pLog}, n, $ t(z) = log(z)$ ) \\
	\lnl{fx:9}$F( \xx )$ $\leftarrow$ (\text{pMax} + \text{pLog} - \text{pLin} ) + $\lambda \parallel \xx \parallel^{2}/2$
\end{algorithm}

Subroutine~\textit{ComputeFX}, shown in Algorithm~\ref{cuda-fx}, 
describes the evaluation of objective function at a given point, $ \xx  =  [ \xx_1; \xx_2; \ldots; \xx_{C-1}] \in \mathbb{R}^{ d} $.
Line~\ref{fx:1} initializes the memory to store partial results, and line~\ref{fx:2} computes the 
matrix-matrix product between training set, \A, and weight matrix, \X. By invoking the CUDA function, 
\textit{ComputeExp}, we compute the partial results, \textit{\mP, \xP, \linP}, as described in 
~\cref{exp-part,m-part,i-part}. Lines~\ref{fx:4},~\ref{fx:5} and,~\ref{fx:6} compute the sum of the temporary 
arrays, and store the partial results in \textit{\text{pLin}, \text{pMax}, pExp}, respectively. \textit{Reduce} 
operation takes a transformation function, $ \textit{t(.)} $, which is applied to the input 
argument before performing the summation. \textit{Reduce}  is a well known function and many highly optimized
implementations are readily available. We use a variation of the algorithm described in~\cite{puremd-gpu}.
\textit{\text{pLog}} is computed at line~\ref{fx:8}. Finally, the objective function value is computed at line~\ref{fx:9},
by adding intermediate results, \textit{\text{pLin}, \text{pMax}, pExp, \text{pLog}} and the regularization term, i.e., 
\begin{align*}
F(\xx) &= (\text{pMax} + \text{pLog} - \text{pLin} ) +  \frac{\lambda}{2} \|\xx \|^{2}  \\
& =  \sum_{i=1}^{n} \left( \text{\mP}_{i} + \text{\logP}_{i} - \text{\linP}_{i} \right) + \frac{\lambda}{2} \|\xx \|^{2} ,
\end{align*}
where
\begin{align}
& \text{\mP}_{i} =  M(\aa_{i})  \quad \quad \quad (\text{cf.\ \cref{eq:max_x}})\label{m-part},\\
& \text{\xP}_{i} = \sum_{c = 1}^{C-1} e^{\lin{\aa_{i}, \xx_{c}} - \text{\mP}_{i}}   \label{exp-part}, \\
& \text{\linP}_{i} =  \sum_{c = 1}^{C-1}\mathbf{1}(\bb_{i} = c) \lin{\aa_{i},\xx_{c}}    \label{i-part}, \\
& \text{\logP}_{i} = \log \left(e^{-\text{\mP}_{i}}+ \text{\xP}_{i} \right) \label{log-part}.
\end{align}

\paragraph{Softmax gradient evaluation}

\begin{algorithm} [!htb]
\caption{Compute $\nabla F$}
\label{cuda:gx}
	\SetAlgoLined
	\SetKwInOut{Input}{input}
	\SetKwInOut{Output}{output}
	\Input{
		\A - Training features \\
		$ \bb $ - Training classes \\
		$ \xx $ - Weights vector \\
		$\lambda$ - Regularization
	}
	\Output{
		$\nabla F( \xx )$ - gradient evaluated at $ \xx $
	}
	\lnl{gx:1} Initialize $\BInd_{(n \times C-1)}$ \\
	\lnl{gx:2} Form $\X =  [\xx_1, \xx_2, \ldots, \xx_{C-1}]_{p \times (C-1)} $ \\
	\BlankLine	
	\lnl{gx:3} Compute $\BInd_{i,c} = \frac{e^{\lin{\aa_{i}, \xx_{c}}}}{1+\sum_{z = 1}^{C-1} e^{\lin{\aa_{i}, \xx_{z}}}}  - \mathbf{1}(\bb_{i} = c)$, similar to Alg.~\ref{cuda:compute-exp}
	\BlankLine	
	\lnl{gx:4}$\nabla F( \xx) \leftarrow \text{vec}( \A^{T} $ \BInd + $\lambda$ \X )
\end{algorithm}

Subroutine~\textit{Compute $\nabla F$}, shown in Algorithm~\ref{cuda:gx}, 
 describes the computation of $\nabla F(x)$. Line~\ref{gx:1} initializes the memory
to store temporary results. Algorithm~\ref{cuda:compute-exp} can be easily modified to compute \BInd. 
Line~\ref{gx:4} computes the gradient of the objective function by matrix multiplication and addition of the
regularization term. 

\paragraph{Softmax Hessian-vector evaluation}


\begin{algorithm} [!htb]
\caption{Compute Hessian-Vector Product, $\nabla^{2} F( \xx ) \qq$}
\label{cuda:hx}
	\SetKwInOut{Input}{input}
	\SetKwInOut{Output}{output}
	\SetKwData{IDX}{idx}	
	\Input{
		\A - Training dataset \\
		$\lambda$ - Regularization \\
		$ \xx $ - Weights vector \\
		$ \qq $ - Vector to compute $\nabla^{2}F( \xx ) \qq $\\
		n - no. of sample points \\
		p - no. of features \\
		C - no. of classes
	}
	\Output{
		\Hq : $\nabla^{2}F( \xx ) \qq $, Hessian-vector product
	}
	\BlankLine
	\lnl{hx:1}Init. \IDX \tcp*{thread-id}
	\lnl{hx:2} Form $\Q =  [\qq_1, \qq_2, \ldots, \qq_{C-1}]_{p \times (C-1)} $ \\
	\lnl{hx:3}$ \V = \A \times \Q $ \\
	\lnl{hx:4} $\W \leftarrow$ compute as shown in~\eqref{eq:matrix-b}, similar to kernel Alg.\ref{cuda:compute-exp} \\	
	\lnl{hx:5}$\vec{U} \leftarrow$ ComputeU (\V, \W, n, p, C ) \\
	\lnl{hx:6}$\Hq \leftarrow$ vec( $\A^{T}$\vec{U} + $\lambda$\Q )
\end{algorithm}

\begin{algorithm} [!htb]
\caption{ComputeU}
\label{cuda:compute-c}
	\SetAlgoLined
	\SetKwInOut{Input}{input}
	\SetKwInOut{Output}{output}
	\SetKwData{IDX}{idx}	
	\Input{
		\V - matrix V as in eq.~\eqref{eq:matrix-a} \\
		\W - matrix W as in eq.~\eqref{eq:matrix-b} \\
		n - no. of sample points \\
		p - no. of features \\
		C - no. of classes
	}
	\Output{
		\vec{U} : matrix \vec{U} as shown in~\eqref{eq:matrix-c}
	}
	\BlankLine
	Initialize \IDX \tcp*{thread-id}
	sum = 0 \\
	\If{ \IDX $<$ n} {
		i = \IDX \% n \tcp*{row no.}
		\ForEach{$j$ in $1:C-1$}{
			sum += $\V_{i,j} \times \W_{i,j}$\;
		}
		\ForEach{$j$ in $1:C-1$}{
			$\vec{U}_{i,j}$ = $\V_{i,j} \times \W_{i,j} - \W_{i,j} \times $ sum\; 
		}
	}
\end{algorithm}

For a given vector, $ \qq $, Algorithm~\ref{cuda:hx}, computes the \textit{Hessian-vector} product, $\nabla^{2} F( \xx ) \qq$. Algorithm~\ref{cuda:hx} is heavily used in CG to solve the linear system $\H \xx = - \bgg $. Line~\ref{hx:1} computes $ \V $, as shown in eq.~\eqref{eq:matrix-a}, a matrix multiplication operation. Line~\ref{hx:4} computes $ \W $ using a function similar to Algorithm~\ref{cuda:compute-exp}, and $ \vec{U} $ is computed using Alg.~\ref{cuda:compute-c} at line~\ref{hx:5}. Finally $ \Hq$ is computed by multiplying $ \A^T $ and $ \vec{U} $, and adding the regularization term in line~\ref{hx:6}.

\section{Experimental Results}
\label{sec:results}

%
%
We present a comprehensive evaluations of the performance of Newton-type methods presented in this paper. We compare our methods to various first-order methods -- SGD with momentum (henceforth referred to as Momentum) \cite{sutskever2013importance}, Adagrad \cite{duchi2011adaptive}, Adadelta \cite{zeiler2012adadelta}, Adam \cite{kingma2014adam} and RMSProp \cite{tijmen2012rmsprop} as implemented in Tensorflow \cite{abadi2016tensorflow}.  We describe our benchmarking setup, 
software used for development, and provide a detailed analysis of the results. The code 
used in this work along with the processed datasets are publicly available~\cite{newton-cg-download}. 
Additionally, raw datasets are also available from the UCI Machine Learning Repository~\cite{uci-repository}.

%
%

\begin{table*}
	\centering
	\caption{Description of the datasets.}
	\label{table:datasets}
	\scalebox{0.75}{
	\begin{tabular}{|c|c|c|c|c|c|c|} \hline
		Classification 	& Dataset	& Train Size ($ n $)	&Test Size 	& No. of Features ($ p $) & No. of Classes ($ C $)	& Lipschitz Const. ($ L $) \\  \hline
		\multirow{4}{*}{Multi-Class}		& Covertype			& 450000		& 131012		& 54		& 7		& 1.92	\\
		& Drive Diagnostics		& 50000		& 8509		& 48		& 11		&  3.95	\\
		& MNIST				& 38000		& 38000		& 785	& 10		&  28.67	\\	
		& CIFAR-10			& 50000		& 10000		& 3072	& 10		& 534.92	\\
		& Newsgroups20 		& 10142		& 1127		& 53975	& 20		& 128.79 	\\ \hline
		\multirow{2}{*}{Binary}			& Gisette				& 6000		& 6500 		& 5000	 & 2		& 751.19	\\
		& Real-Sim			& 65078		& 7231		& 20958	& 2		& 206.76		\\ \hline
	\end{tabular}
	}
\end{table*}

%
%
\subsection{Experimental Setup and Data}

Newton-type methods are implemented in C/C++ using CUDA/8.0 toolkit. 
For matrix operations, 
matrix-vector, and matrix-matrix operations, we use cuBLAS and cuSparse libraries.
First order-methods are implemented using Tensorflow/1.2.1 python scripts.
All results are generated using an Ubuntu server with 256GB RAM, 48-core Intel
Xeon E5-2650 processors, and Tesla P100 GPU cards. For all of our experiments, we consider the $\ell_{2}$-regularized objective $ F(\xx) + \lambda \|\xx\|^{2}/2$, where $ F $ is as in eq.~\eqref{eq:softmax_log_likelihood} and $ \lambda $ is the regularization parameter. 
Seven real datasets are used for performance comparisons. 
Table~\ref{table:datasets} presents the datasets used, along with the \textit{Lipschitz} continuity constant of $ \nabla F(\xx) $, denoted by $ L$. Recall that, an (over-estimate) of the \textit{condition-number} of the problem, as defined in~\cite{roosta2016sub_global}, can be obtained by $ (L+\lambda)/\lambda $.
As it is often done in practice, we first normalize the datasets such that each column of the data matrix $ \A \in \mathbb{R}^{n \times p}$ (as defined in Section~\ref{sec:multi_class}), has Euclidean norm one. This helps with the conditioning of the problem. The resulting dataset
is, then, split into training and testing sets, as shown in the Table~\ref{table:datasets}.

\subsection{Parameterization of Various Methods}

The Lipschitz constant, $ L $, is used to estimate the
learning rate (step-size) for first order methods.
For each dataset, we use a range of learning rates from $10^{-6}/L$ to $10^{6}/L$, in increments of $ 10 $, a total of 13 step sizes, to determine the best performing
learning rate (one that yields the maximum test accuracy). Rest of the hyper-parameters required by first-order methods
are set to the default values, as recommended in Tensorflow. Two batch sizes
are used for first-order methods: a small batch size of 128 (empirically, it has been argued that smaller batch sizes might lead to better performance~\cite{le2004large}),
and a larger batch size of 20\% of the dataset. For Newton-type methods, when the gradient is sampled, its sample size is set to $ |\mathcal{S}_{\bgg}| = 0.2 n $.

We present results for two implementations of second-order methods: (a) \textit{FullNewton}, the classical 
Newton-CG algorithm~\cite{nocedal2006numerical}, which uses the exact gradient and Hessian,
and (b) \textit{SubsampledNewton}, sub-sampled variant of Newton-CG using uniform sub-sampling for gradient/Hessian approximations. 
When compared with first-order methods that use batch size of 128, \textit{SubsampledNewton} uses full gradient and 5\% for Hessian 
sample size, referred to as \textit{SubsampledNewton-100}. When first-order methods' batch size is set to 20\%, \textit{SubsampledNewton} uses 20\% for gradient and 5\% for Hessian sampling, referred to as \textit{SubsampledNewton-20}. CG-tolerance is set to $ 10^{-4} $. Maximum CG iterations is 10 for all of the datasets except 
\textit{Drive Diagnostics} and \textit{Gisette}, for which it is 1000. $ \lambda $ is set to $ 10^{-3} $ and we perform 100 iterations (epochs) for each dataset.


\subsection{Computing Platforms}

For benchmarking first order methods with batch size 128, we use CPU-cores only and for the larger
batch size 1-GPU and 1-CPU-core are used. For brevity we only present the best performance
results (lowest time-per-epochs); 
see~\ref{sec:platform-cpu-gpu} for more detailed
discussion on performance results on various compute platforms. Newton-type methods always use
1-GPU and 1-CPU-core for computations.


%
%
%
%
%

%
%
%
%

\subsection{Performance Comparisons}

\begin{table*}
\centering
\caption{Performance comparison between first-order and second-order methods. First order methods, with batch size 128, are compared 
with \textit{SubsampledNewton} using full gradient and a Hessian sample size of 5\%. First order methods, with batch size 20\%, are compared 
with \textit{SubsampledNewton} using sample sizes of 20\% and 5\% for gradient and Hessian, respectively. \textit{FullNewton}
uses the entire dataset for gradient and Hessian evaluations.} 
\label{table-normalized-uniform}
\scalebox{0.9}{
\begin{tabular}
      {cccc} \hline 
      Time vs. Accuracy & Time vs. Misfit  & Time vs. Accuracy & Time vs. Misfit \\
      \multicolumn{2}{c}{First Order Batch Size = 128} & \multicolumn{2}{c}{First Order Batch Size = 20\%} \\       
      \multicolumn{2}{c}{Alg.~\ref{alg:ssn} Gradient Sample Size = 100\%} & \multicolumn{2}{c}{Alg.~\ref{alg:ssn} Gradient Sample Size = 20\%} \\       
      \multicolumn{2}{c}{Alg.~\ref{alg:ssn} Hessian Sample Size = 5\%} & \multicolumn{2}{c}{Alg.~\ref{alg:ssn} Hessian Sample Size = 5\%} \\       
      \hline
      \multicolumn{4}{c}{  	\parbox[c]{6in}{
      	\includegraphics[width=6in, height=0.5in]{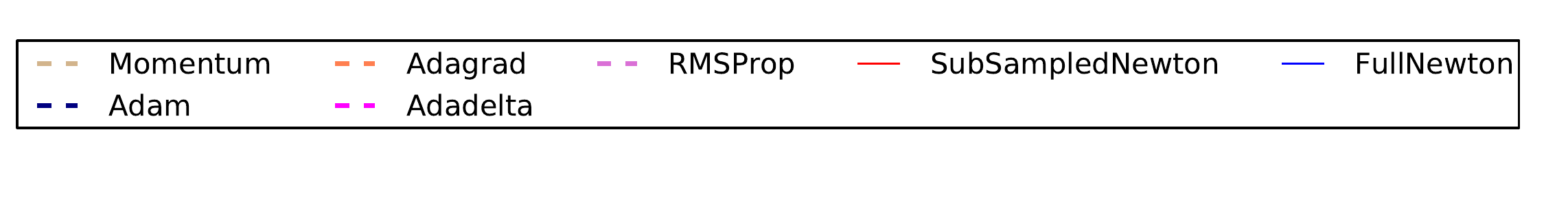}
      } }	\\
 	\parbox[c]{1.5in}{
      	\includegraphics[width=1.4in, height=0.9in]{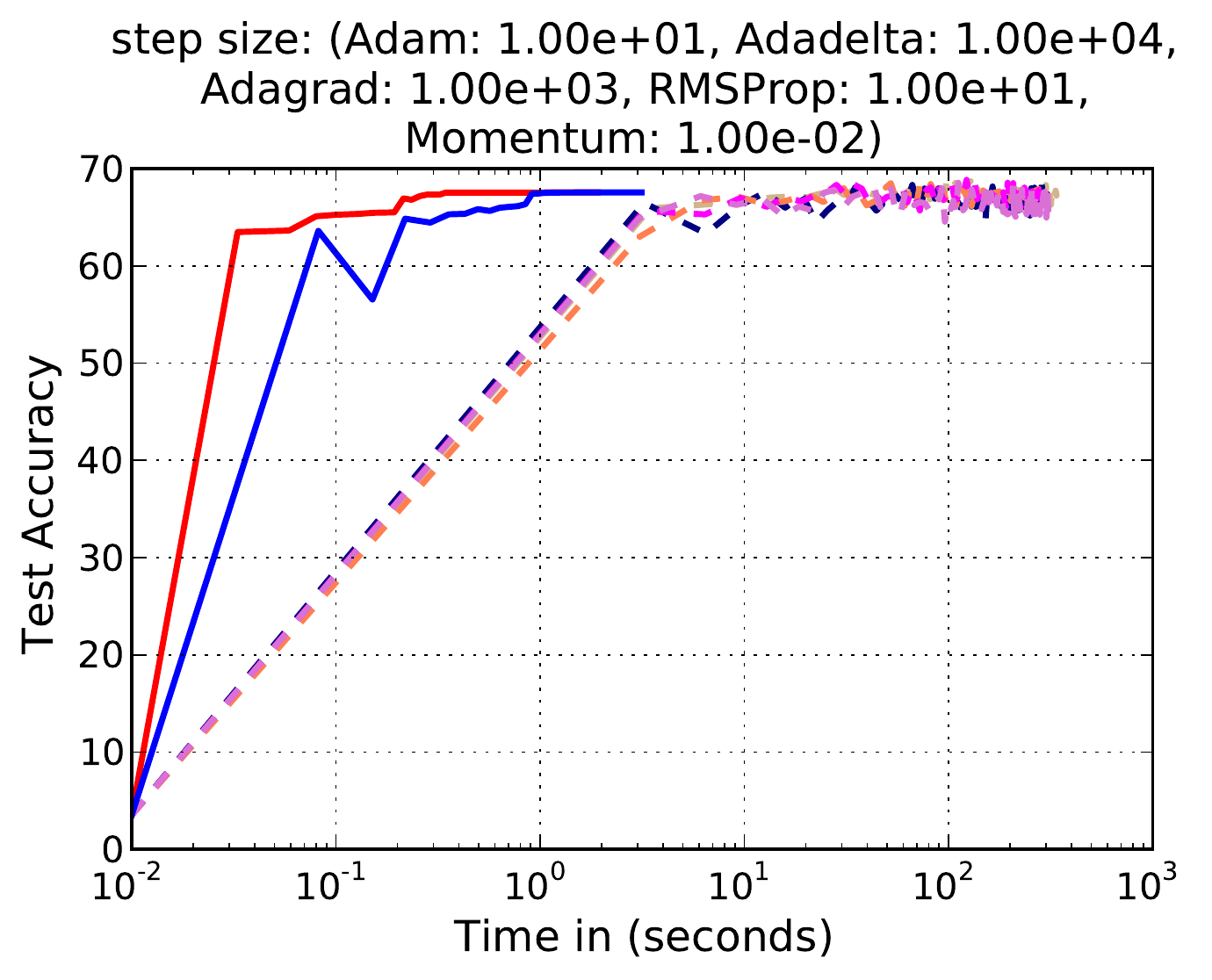}
      } & 
 	\parbox[c]{1.5in}{
      	\includegraphics[width=1.4in, height=0.9in]{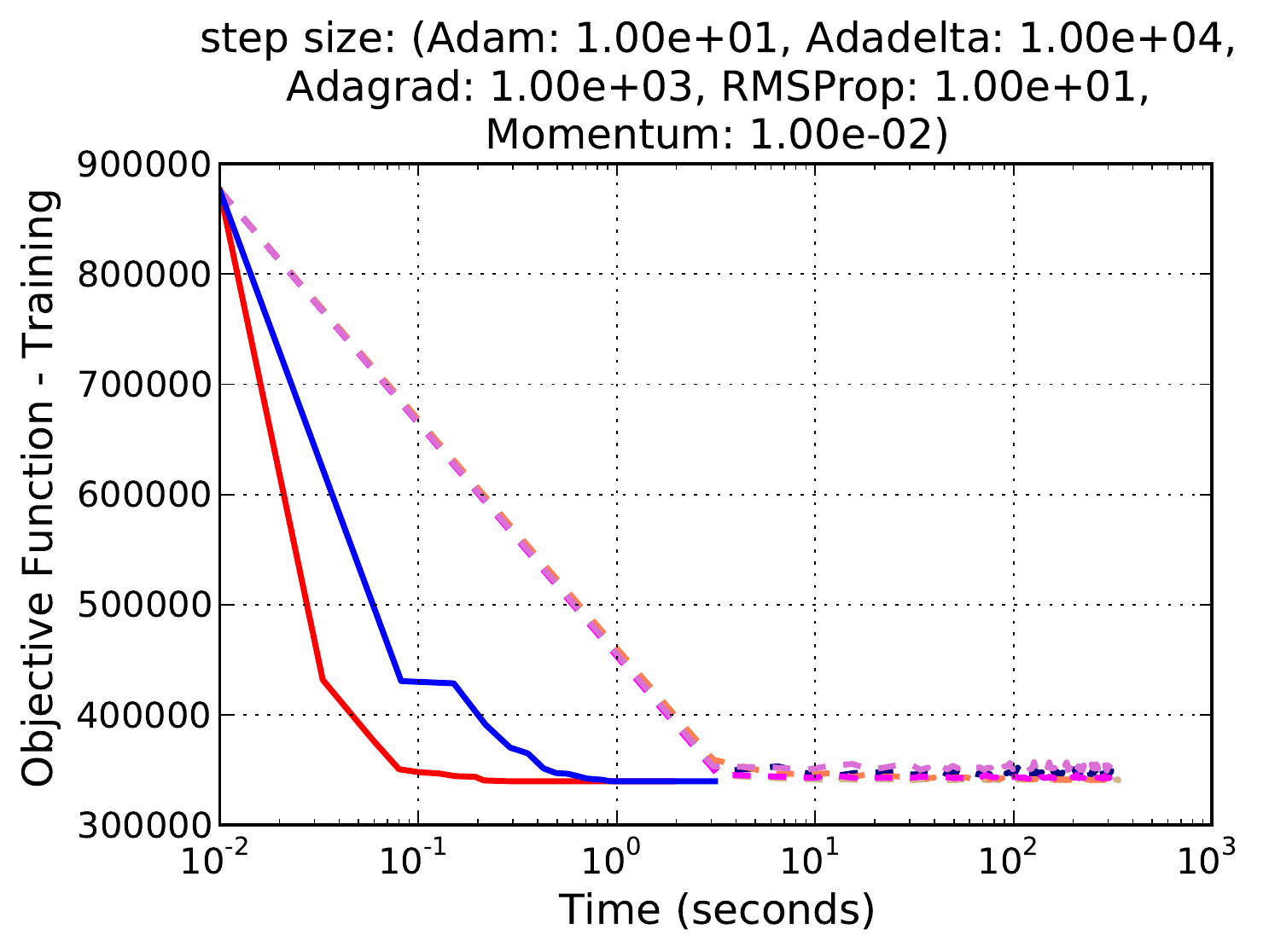}
      } & 
 	\parbox[c]{1.5in}{
      	\includegraphics[width=1.4in, height=0.9in]{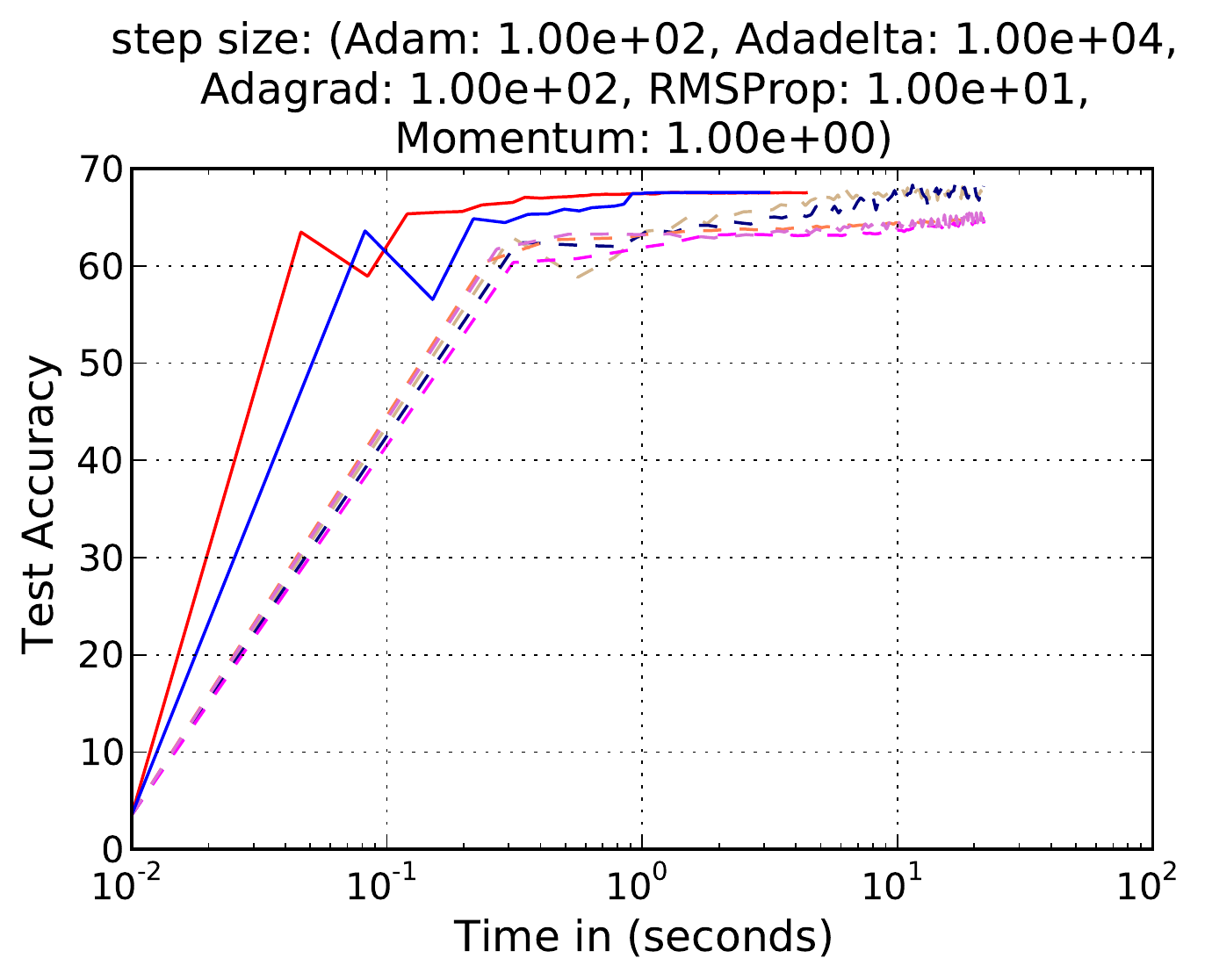}
      } & 
 	\parbox[c]{1.5in}{
      	\includegraphics[width=1.4in, height=0.9in]{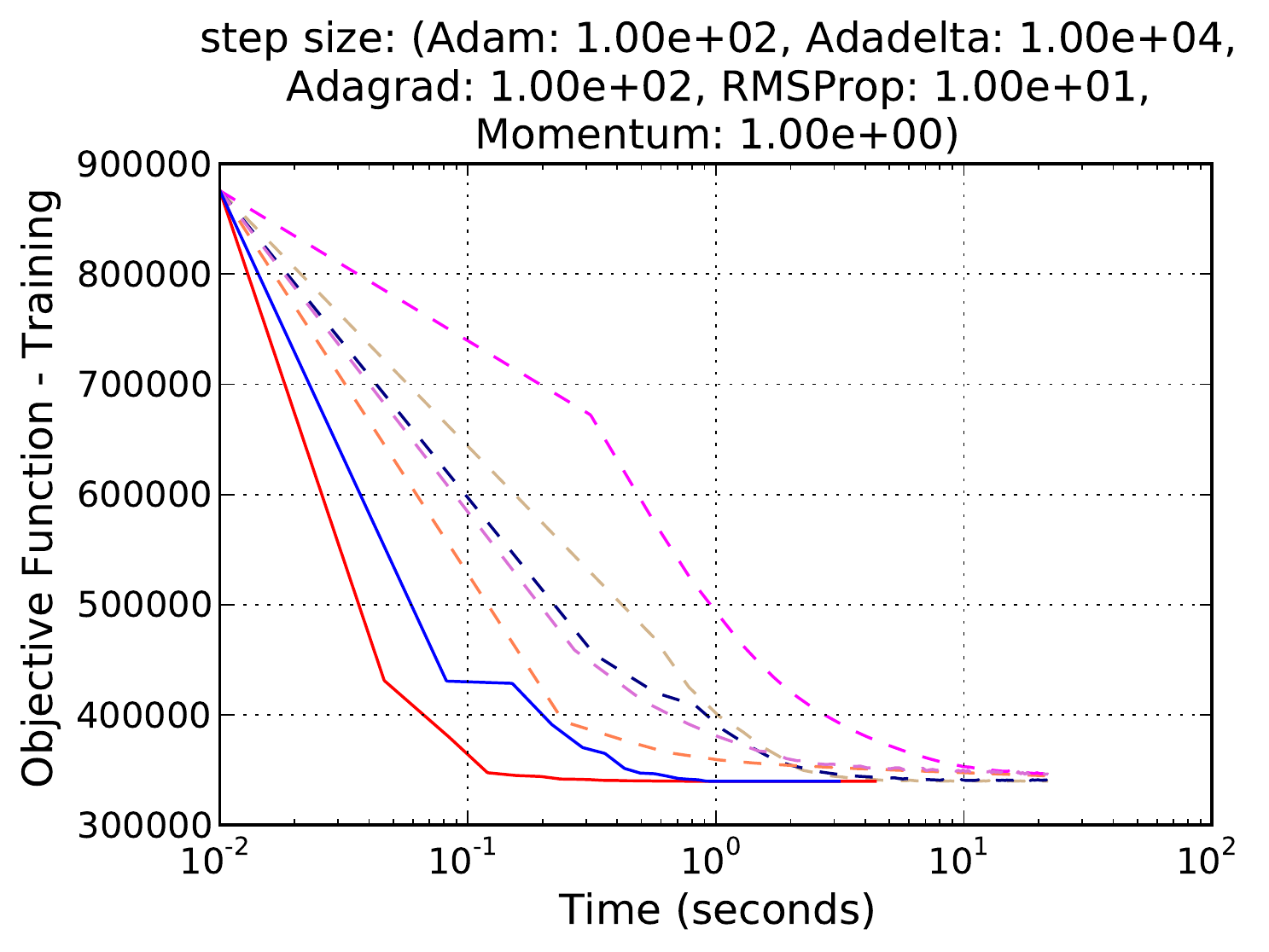}
      } \\
      \multicolumn{4}{c}{Covertype} \\
 	\parbox[c]{1.5in}{
      	\includegraphics[width=1.4in, height=0.9in]{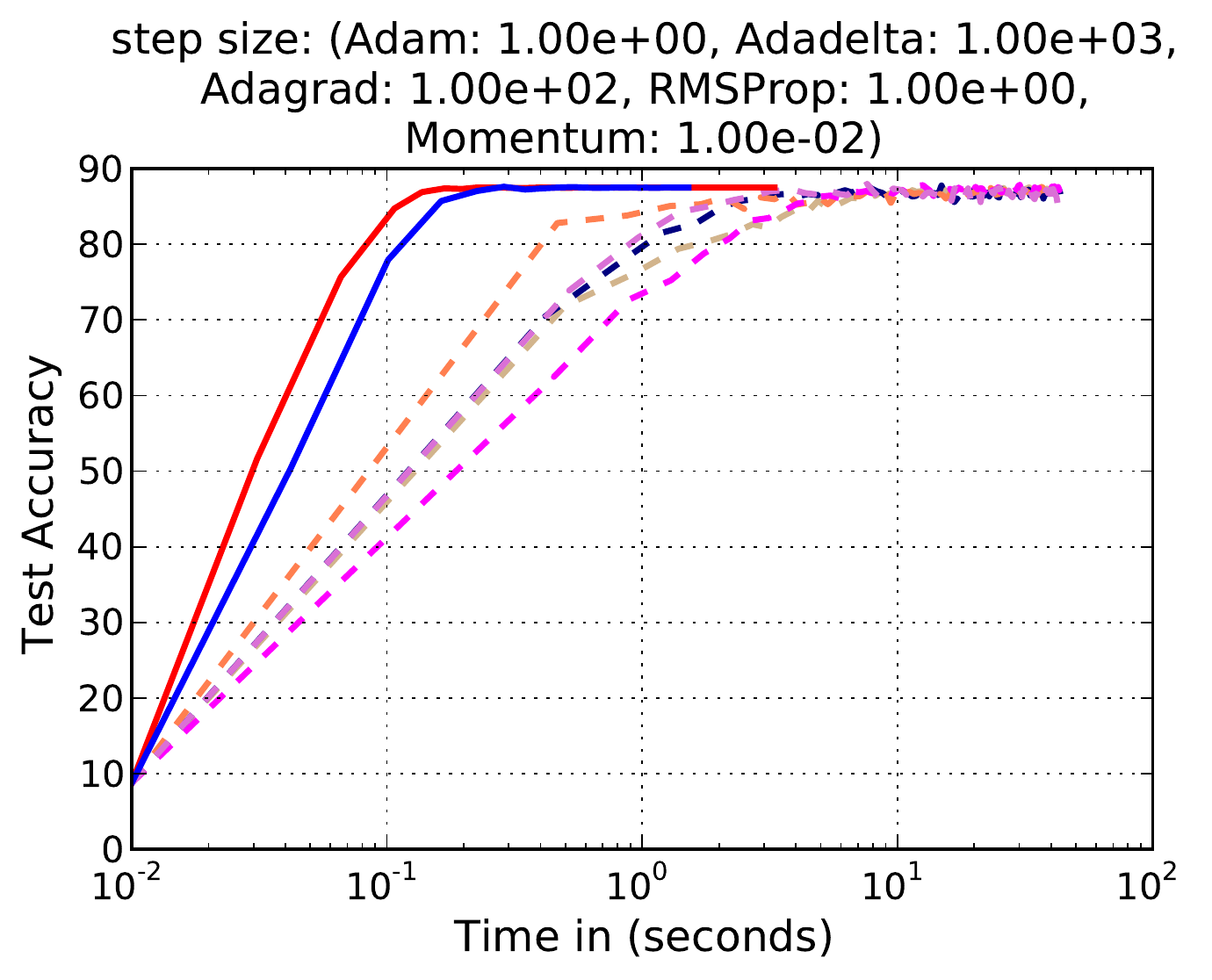}
      } & 
 	\parbox[c]{1.5in}{
      	\includegraphics[width=1.4in, height=0.9in]{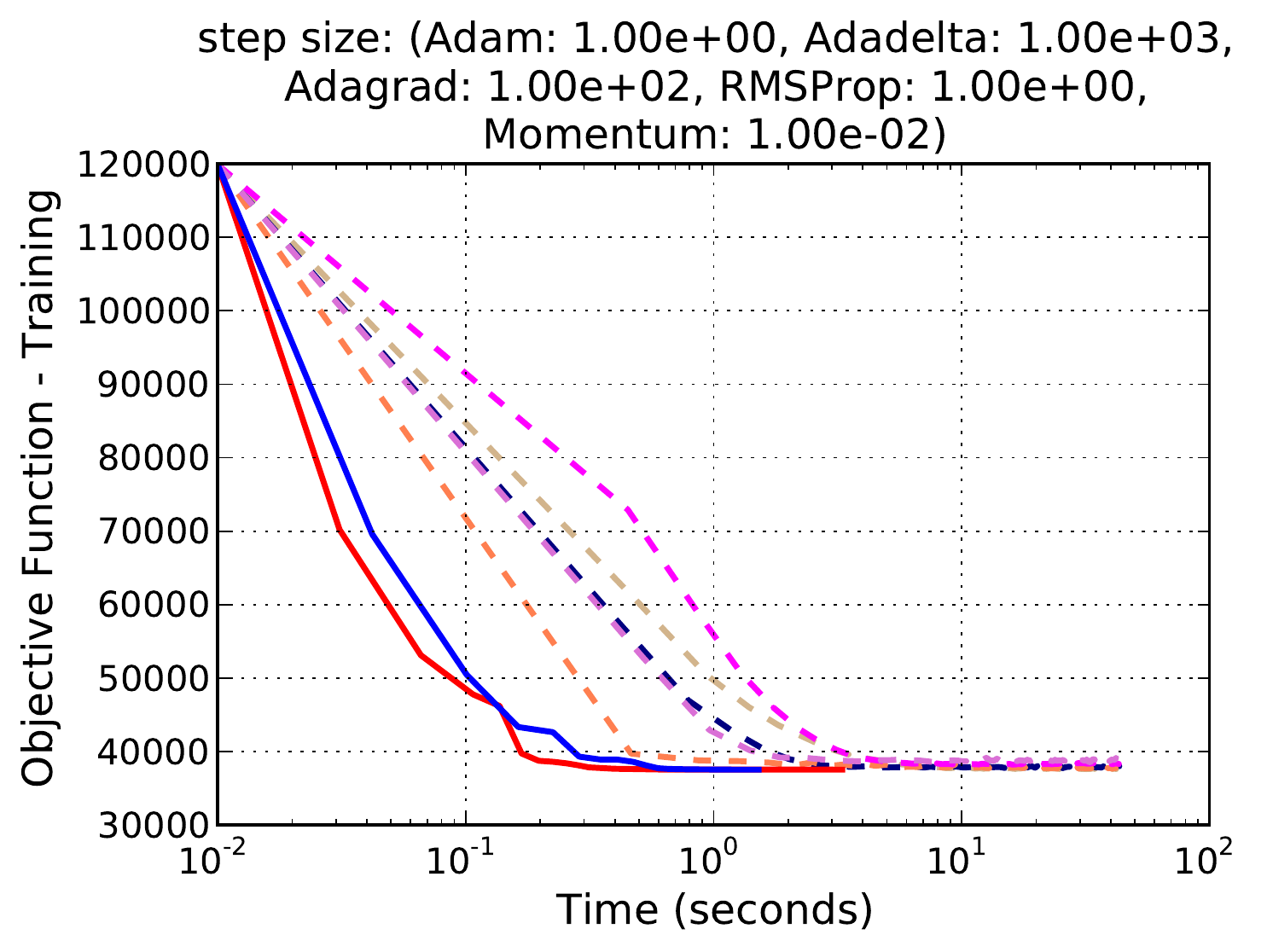}
      } & 
 	\parbox[c]{1.5in}{
      	\includegraphics[width=1.4in, height=0.9in]{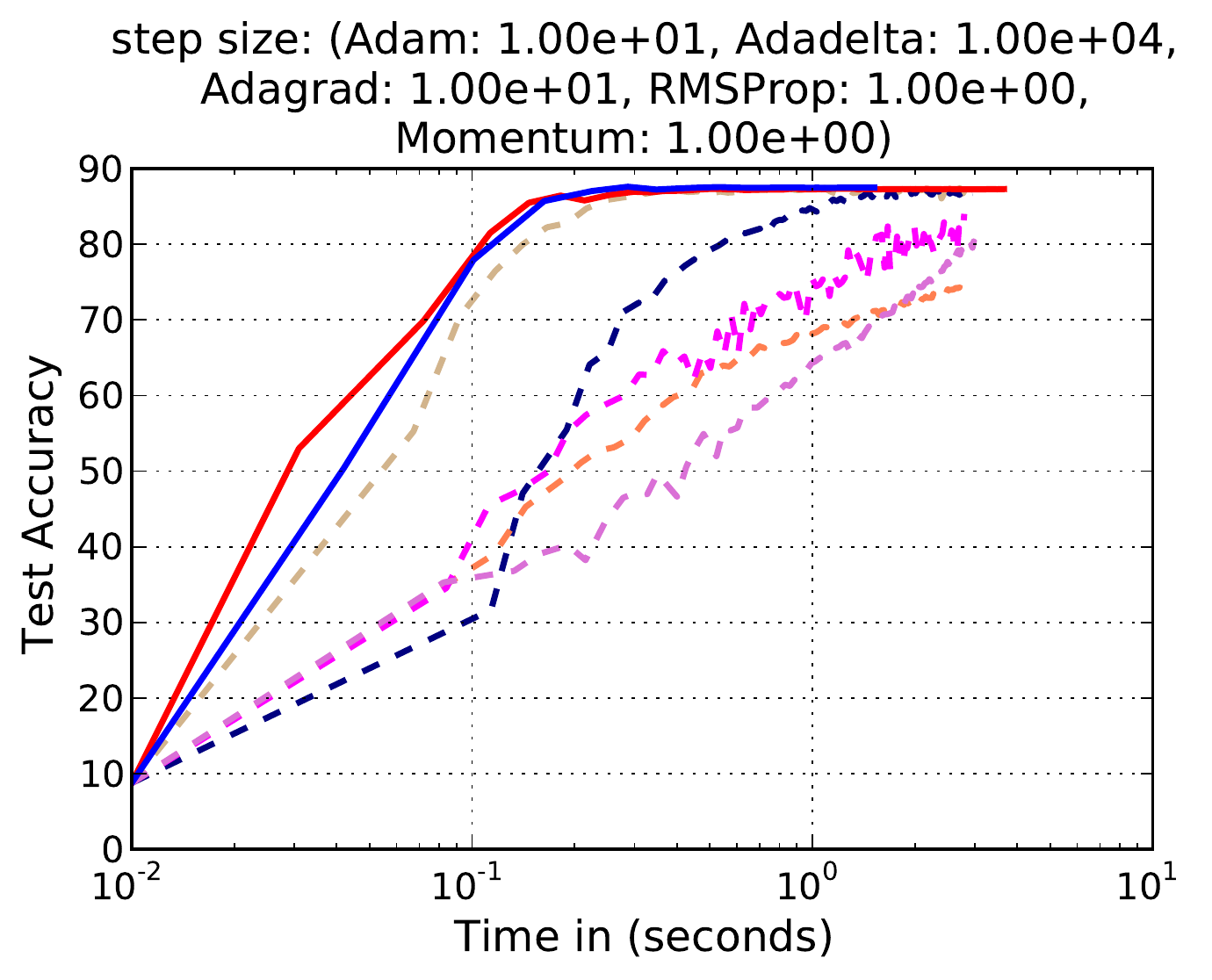}
      } & 
 	\parbox[c]{1.5in}{
      	\includegraphics[width=1.4in, height=0.9in]{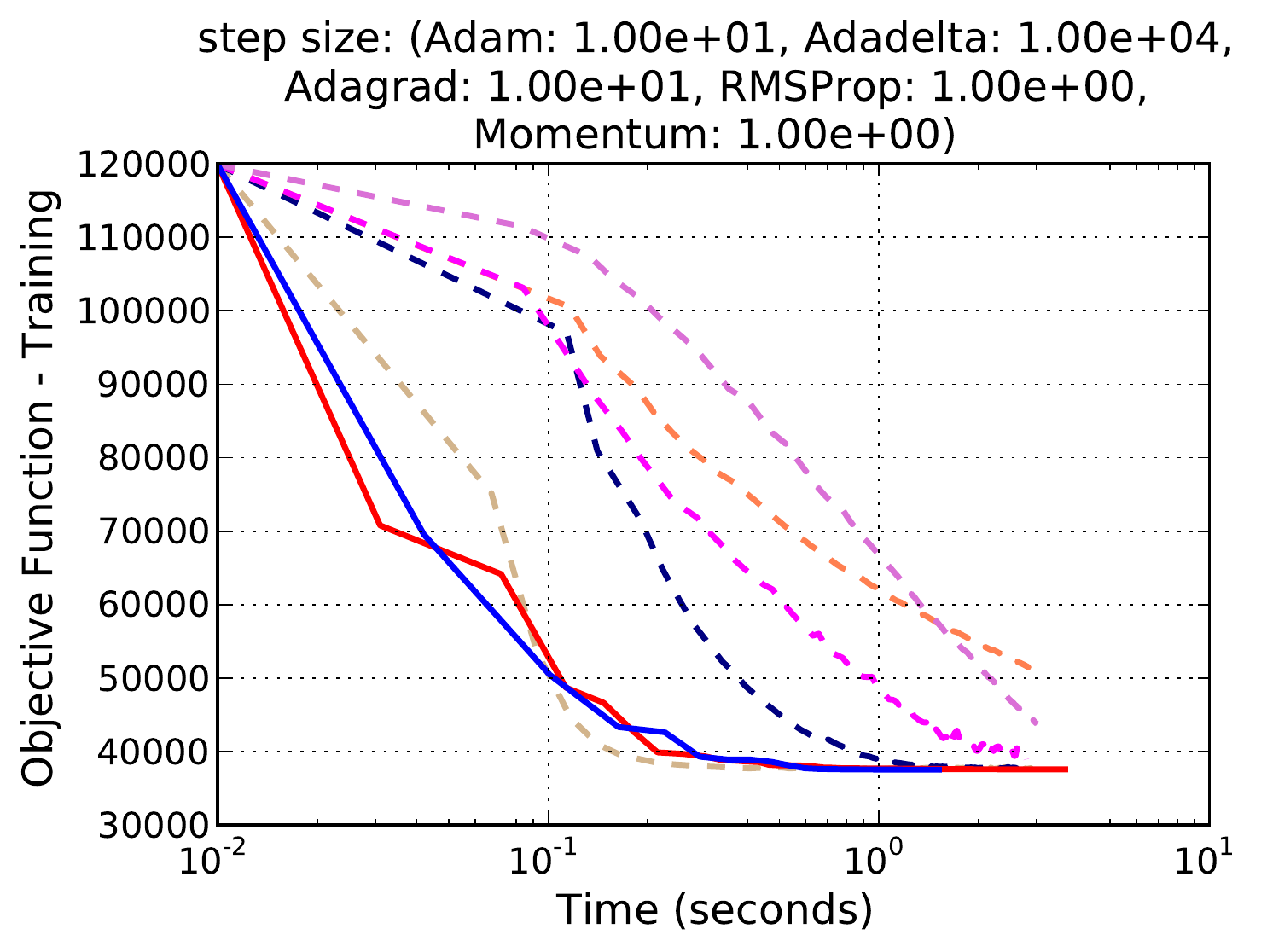}
      } \\
      \multicolumn{4}{c}{Drive Diagnostics} \\
      \parbox[c]{1.5in}{
      	\includegraphics[width=1.4in, height=0.9in]{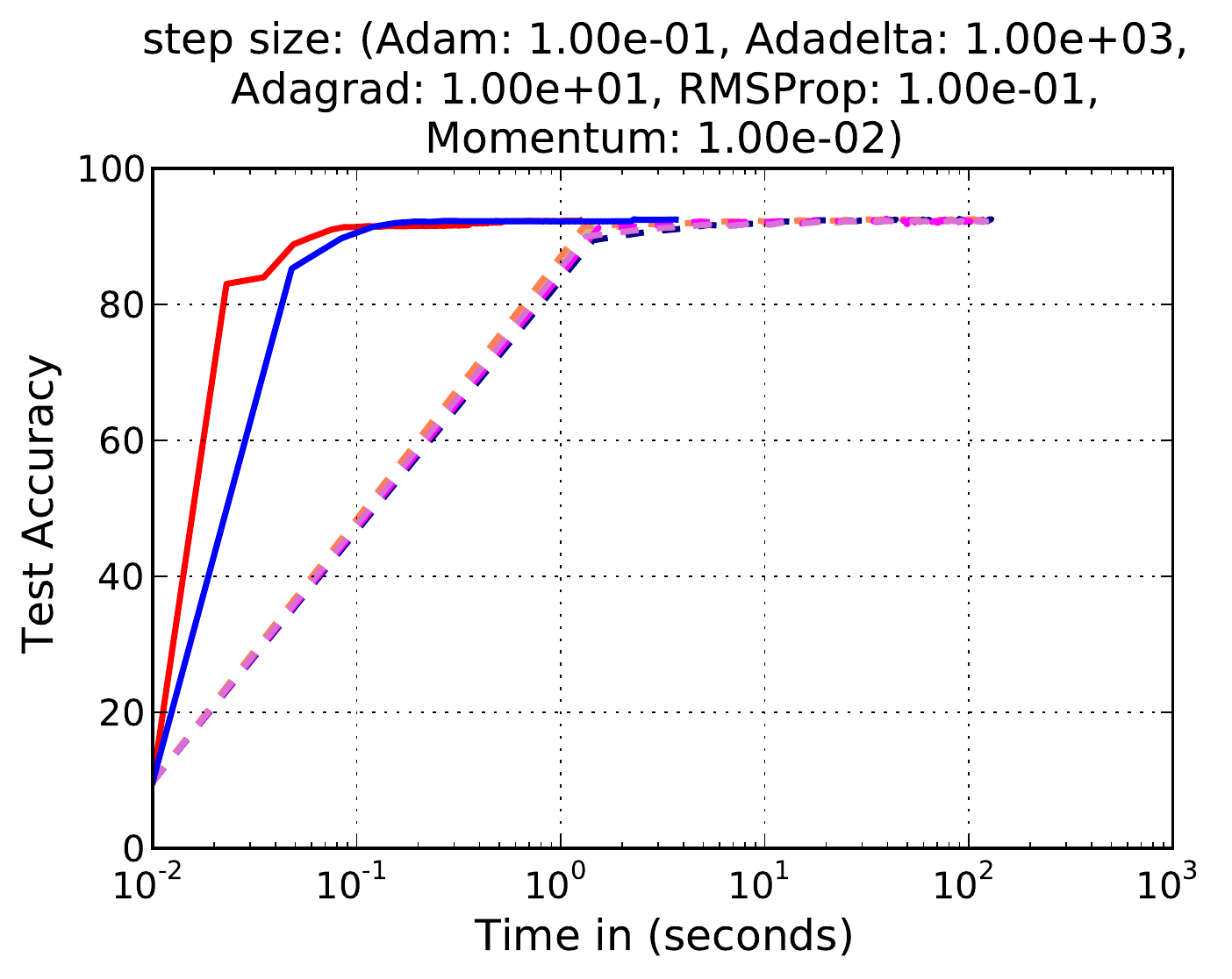}
      } & 
 	\parbox[c]{1.5in}{
      	\includegraphics[width=1.4in, height=0.9in]{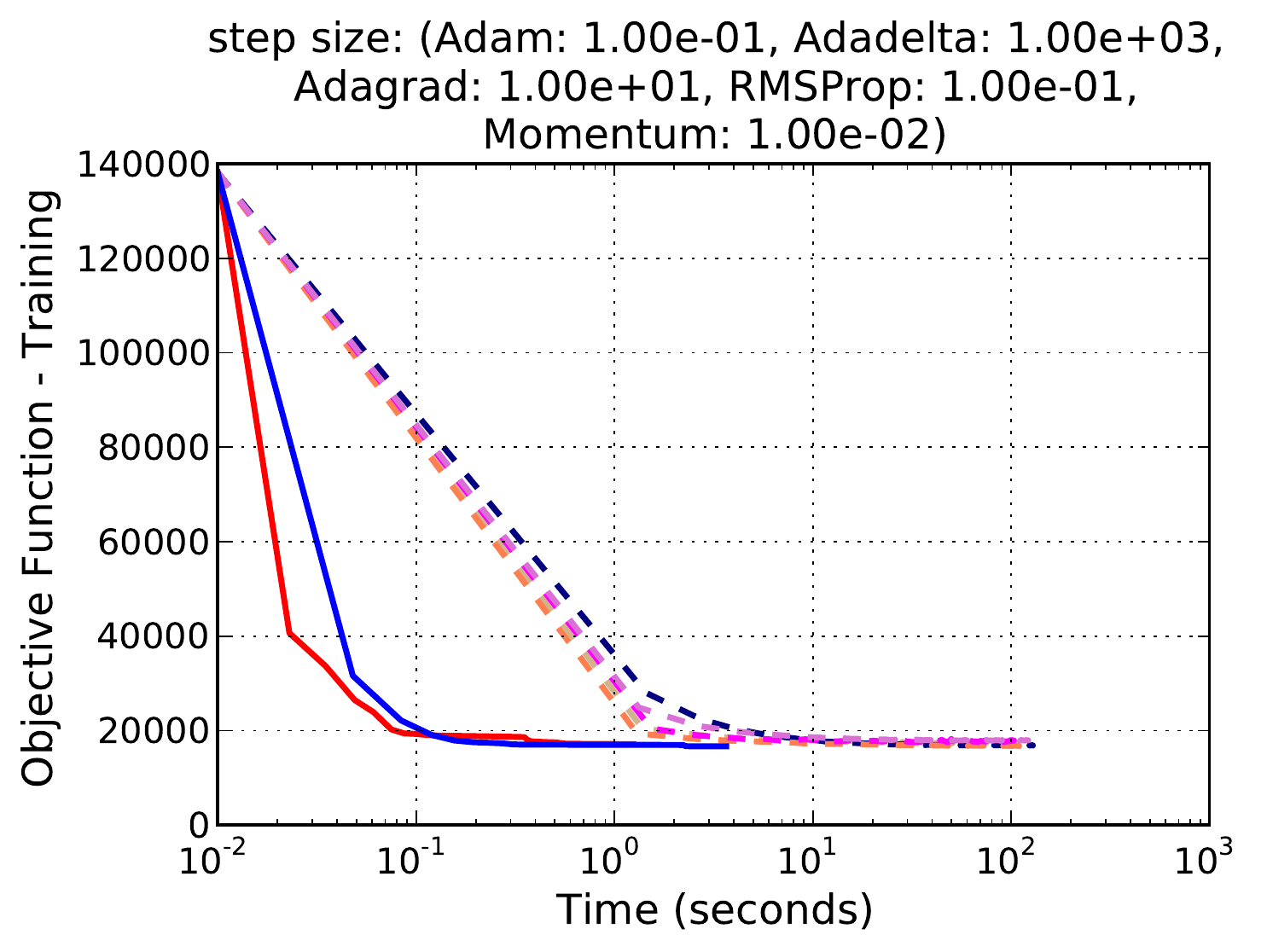}
      } & 
 	\parbox[c]{1.5in}{
      	\includegraphics[width=1.4in, height=0.9in]{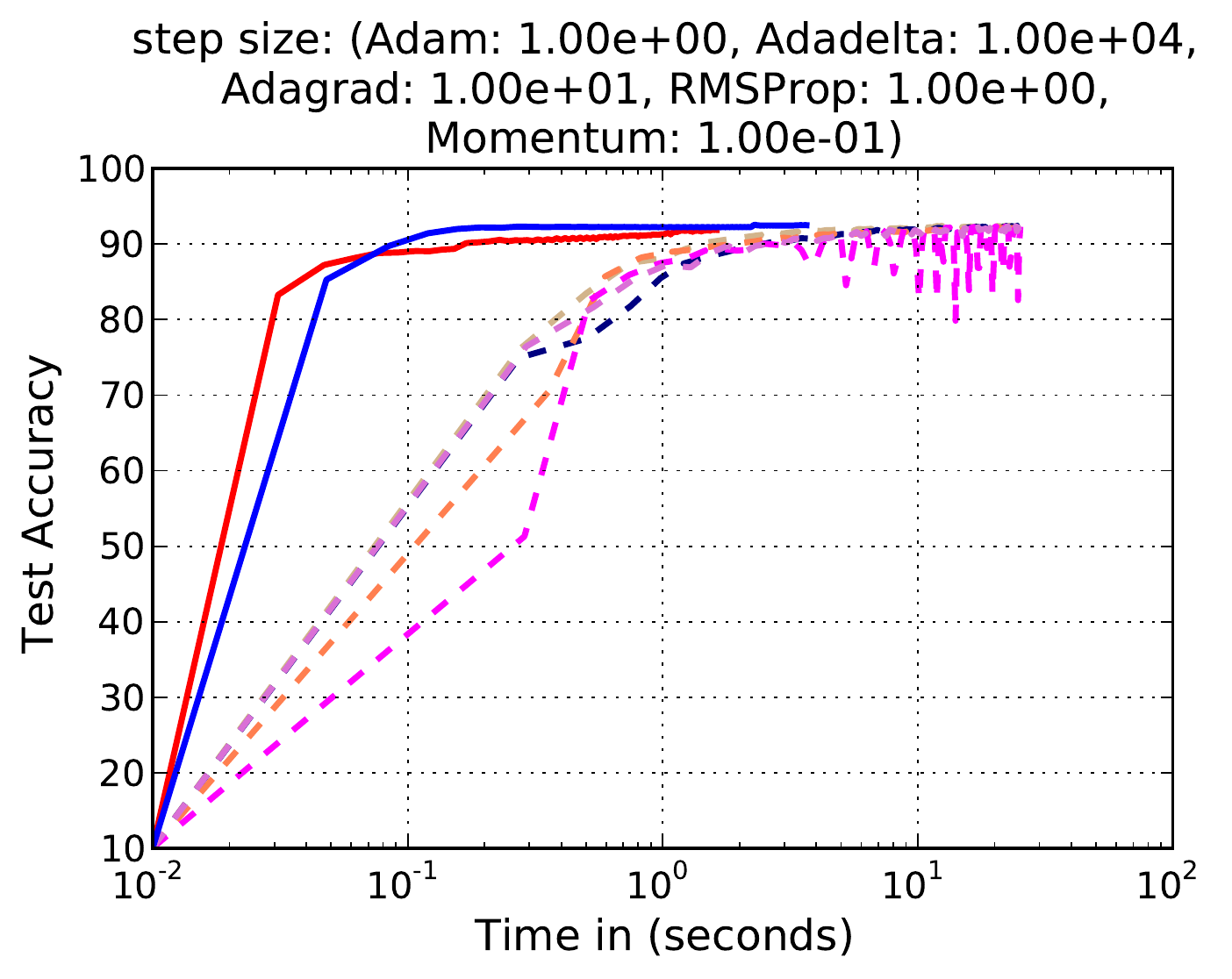}
      } & 
 	\parbox[c]{1.5in}{
      	\includegraphics[width=1.4in, height=0.9in]{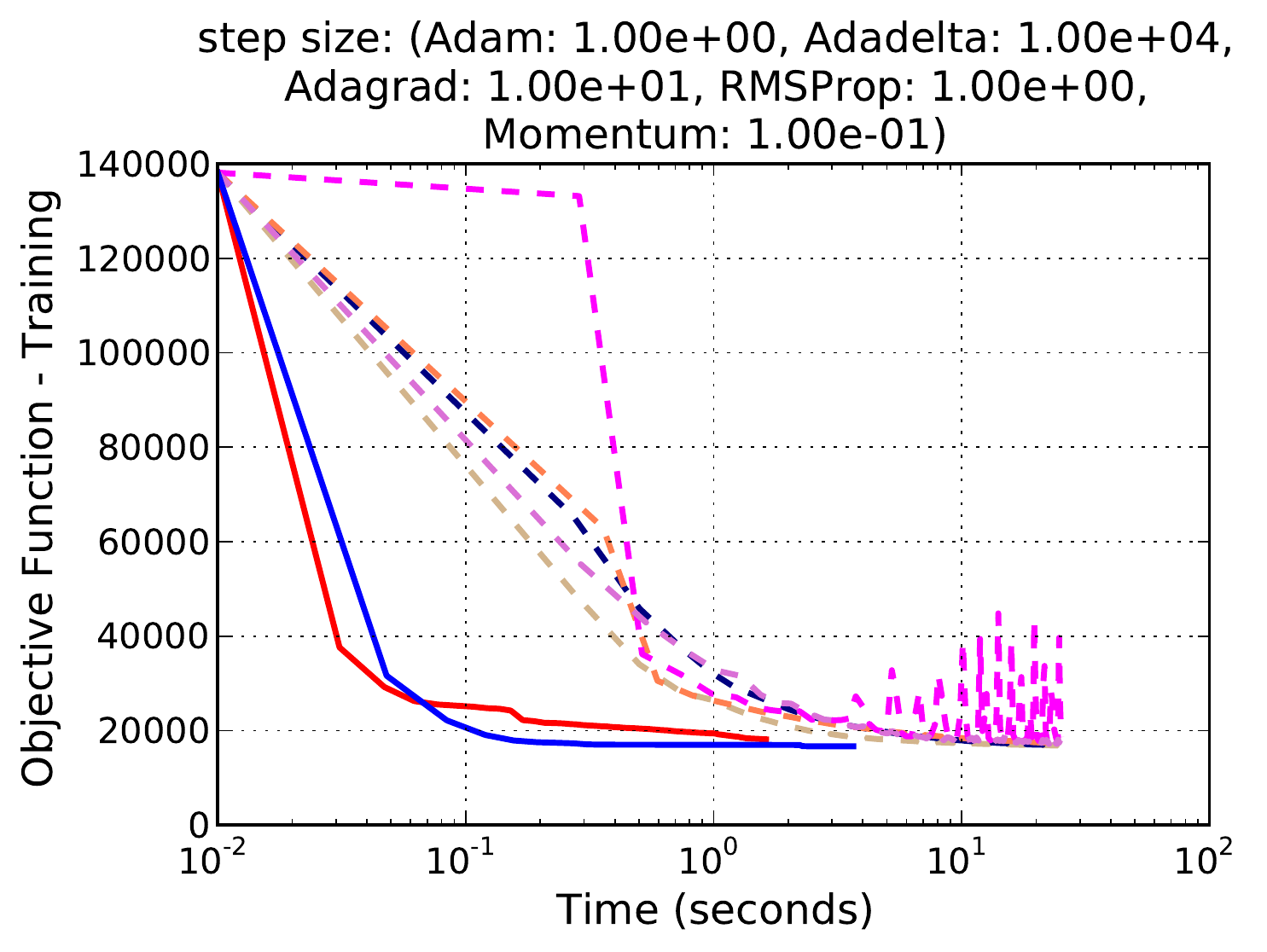}
      } \\
      \multicolumn{4}{c}{MNIST} \\
 	\parbox[c]{1.5in}{
      	\includegraphics[width=1.4in, height=0.9in]{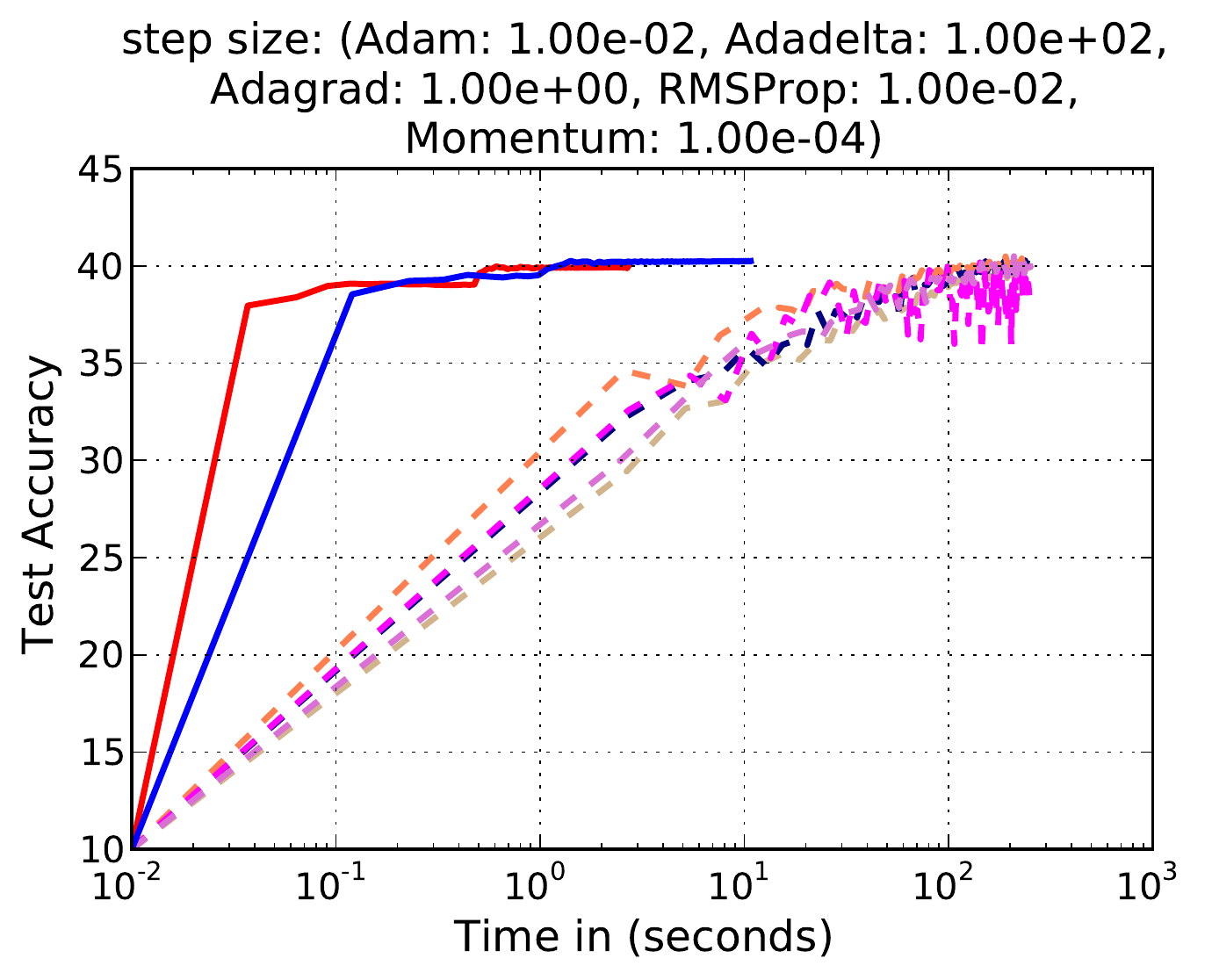}
      } & 
 	\parbox[c]{1.5in}{
      	\includegraphics[width=1.4in, height=0.9in]{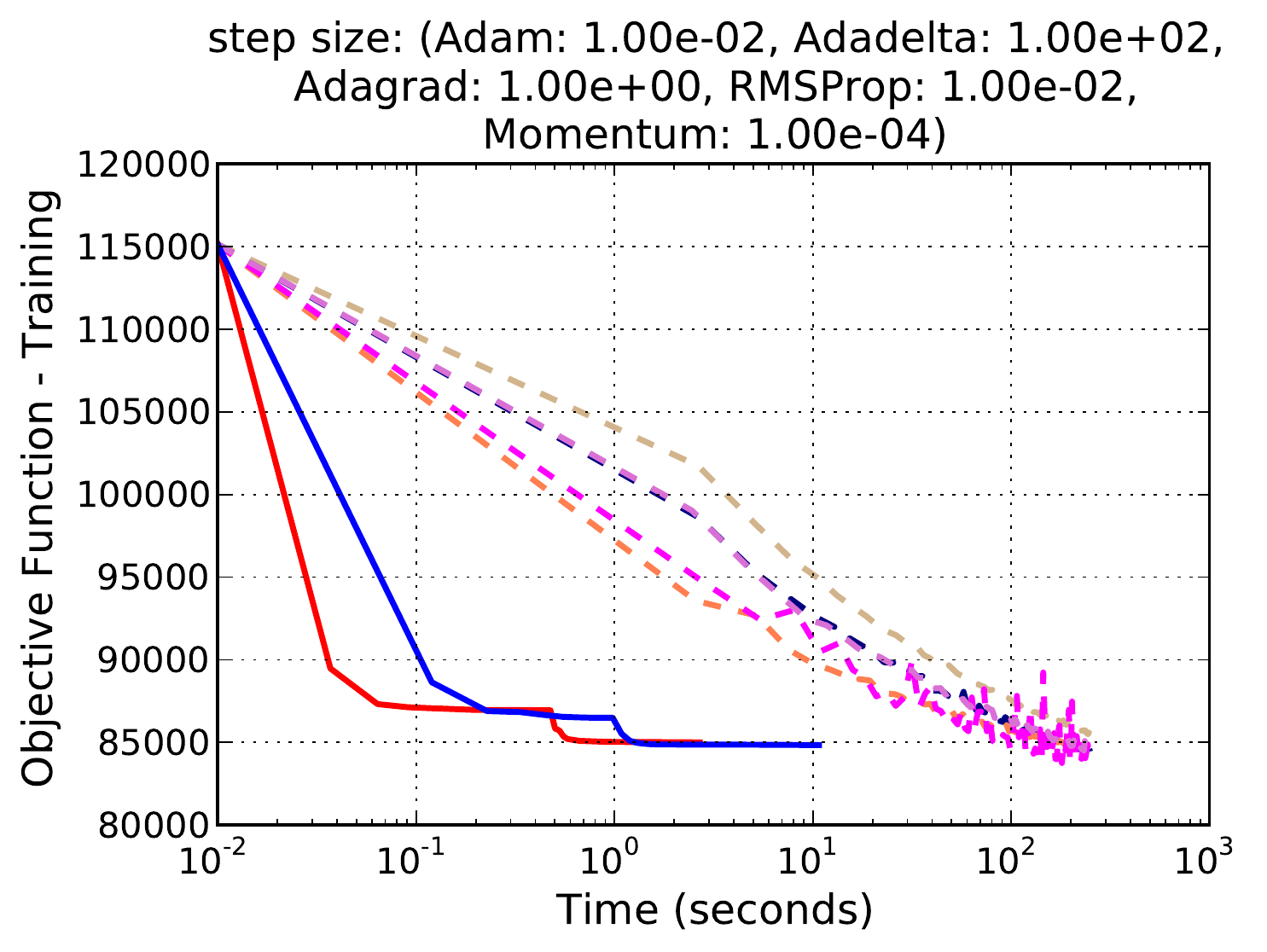}
      } & 
 	\parbox[c]{1.5in}{
      	\includegraphics[width=1.4in, height=0.9in]{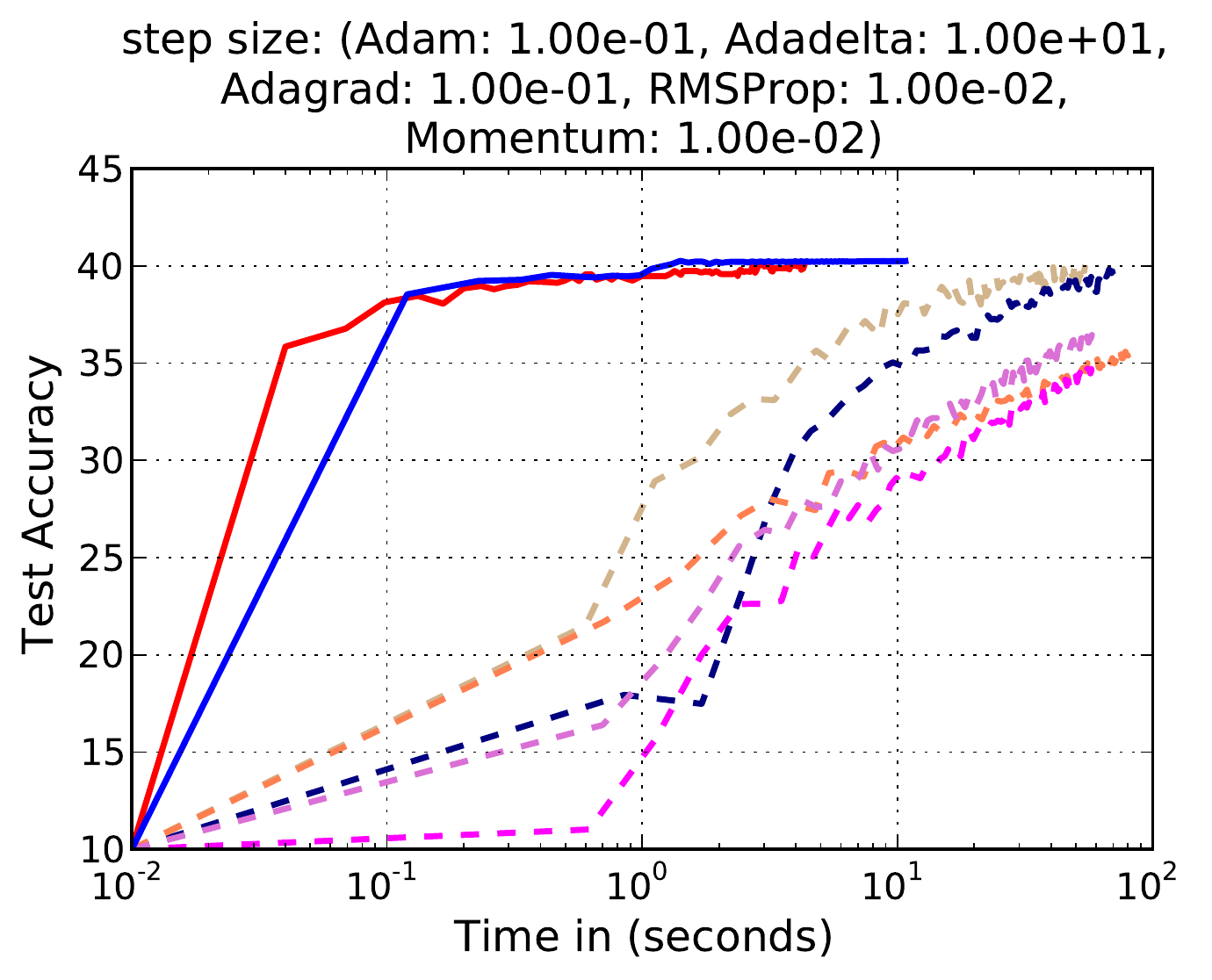}
      } & 
 	\parbox[c]{1.5in}{
      	\includegraphics[width=1.4in, height=0.9in]{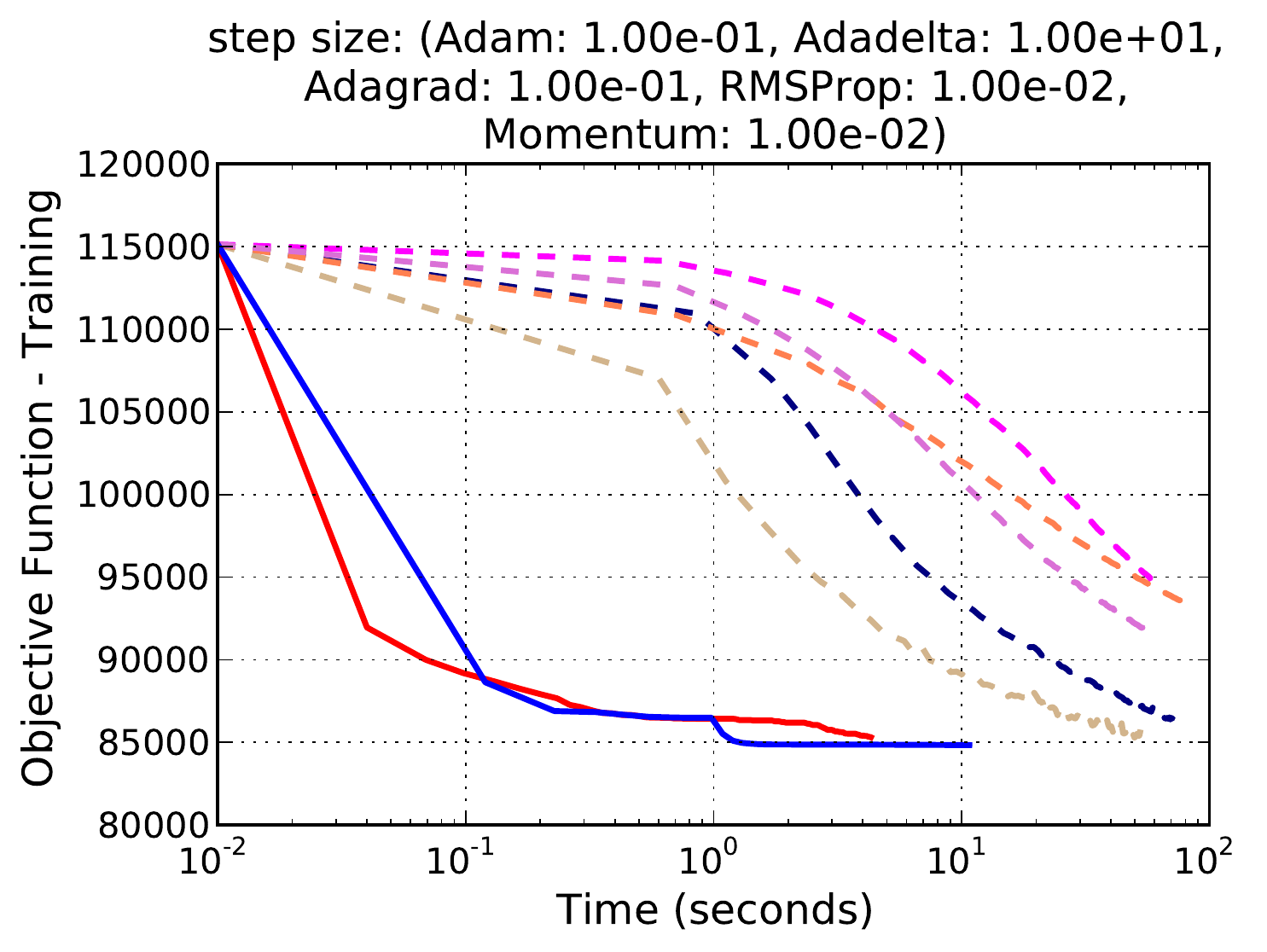}
      } \\
      \multicolumn{4}{c}{CIFAR-10} \\
 	\parbox[c]{1.5in}{
      	\includegraphics[width=1.4in, height=0.9in]{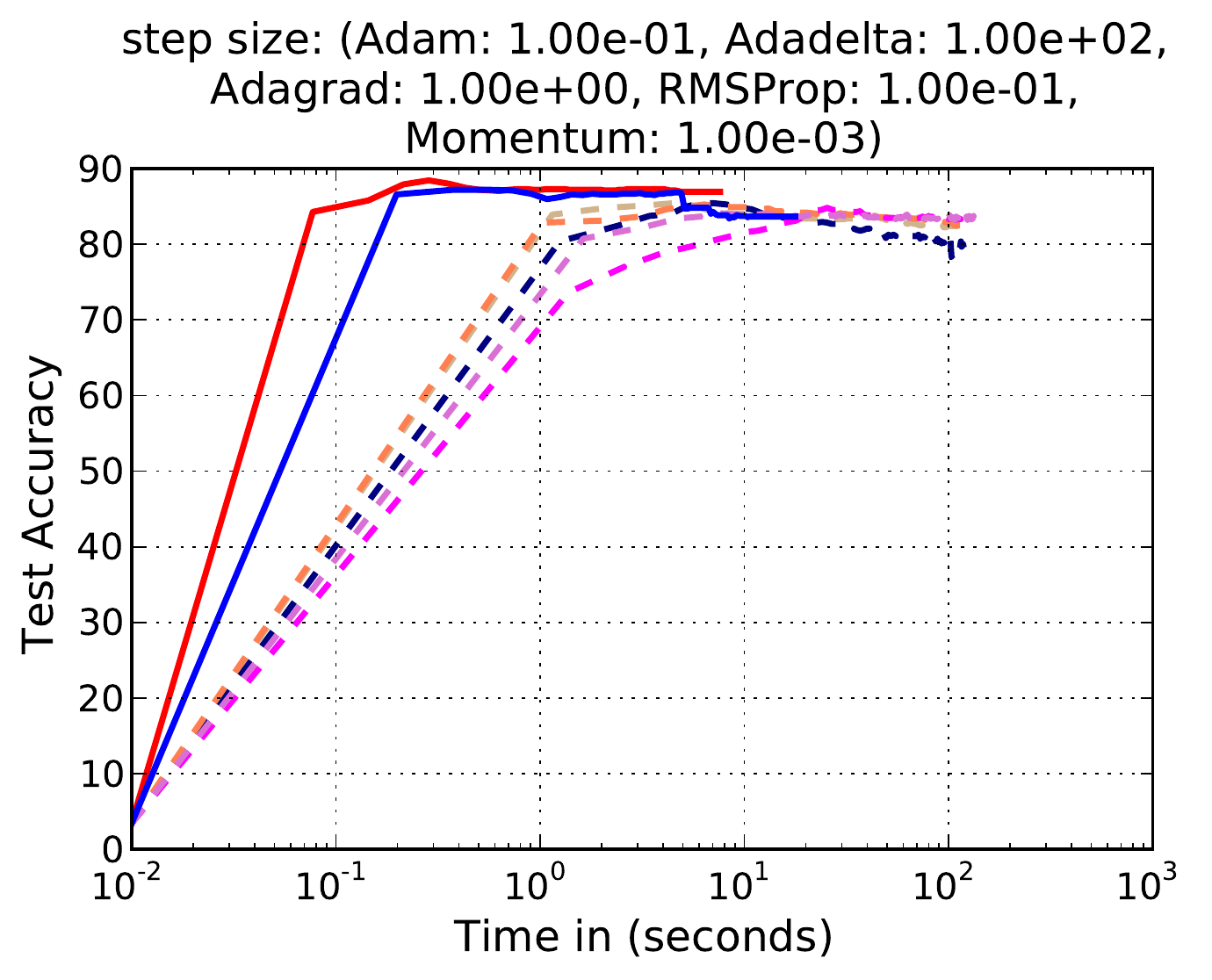}
      } & 
 	\parbox[c]{1.5in}{
      	\includegraphics[width=1.4in, height=0.9in]{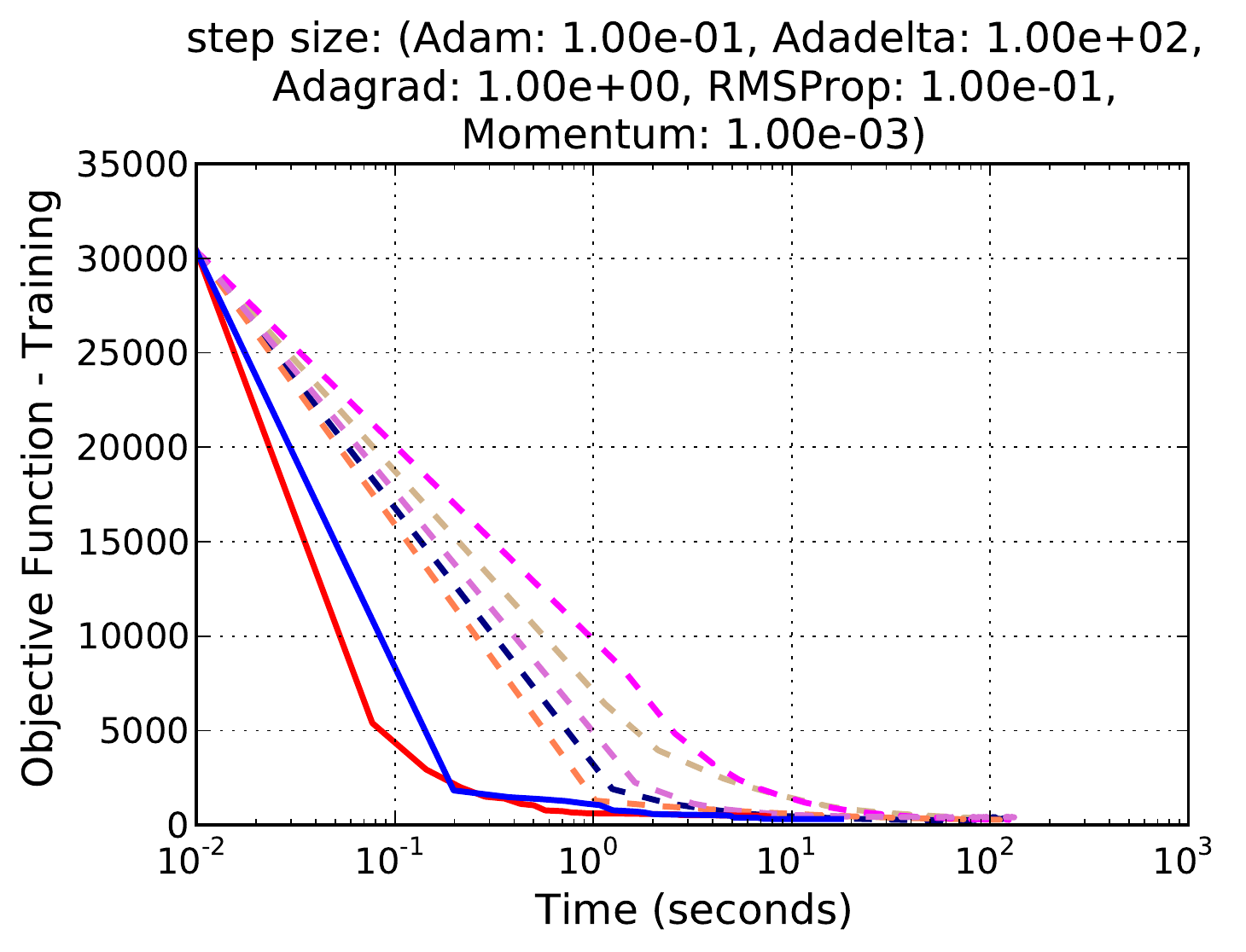}
      } & 
 	\parbox[c]{1.5in}{
      	\includegraphics[width=1.4in, height=0.9in]{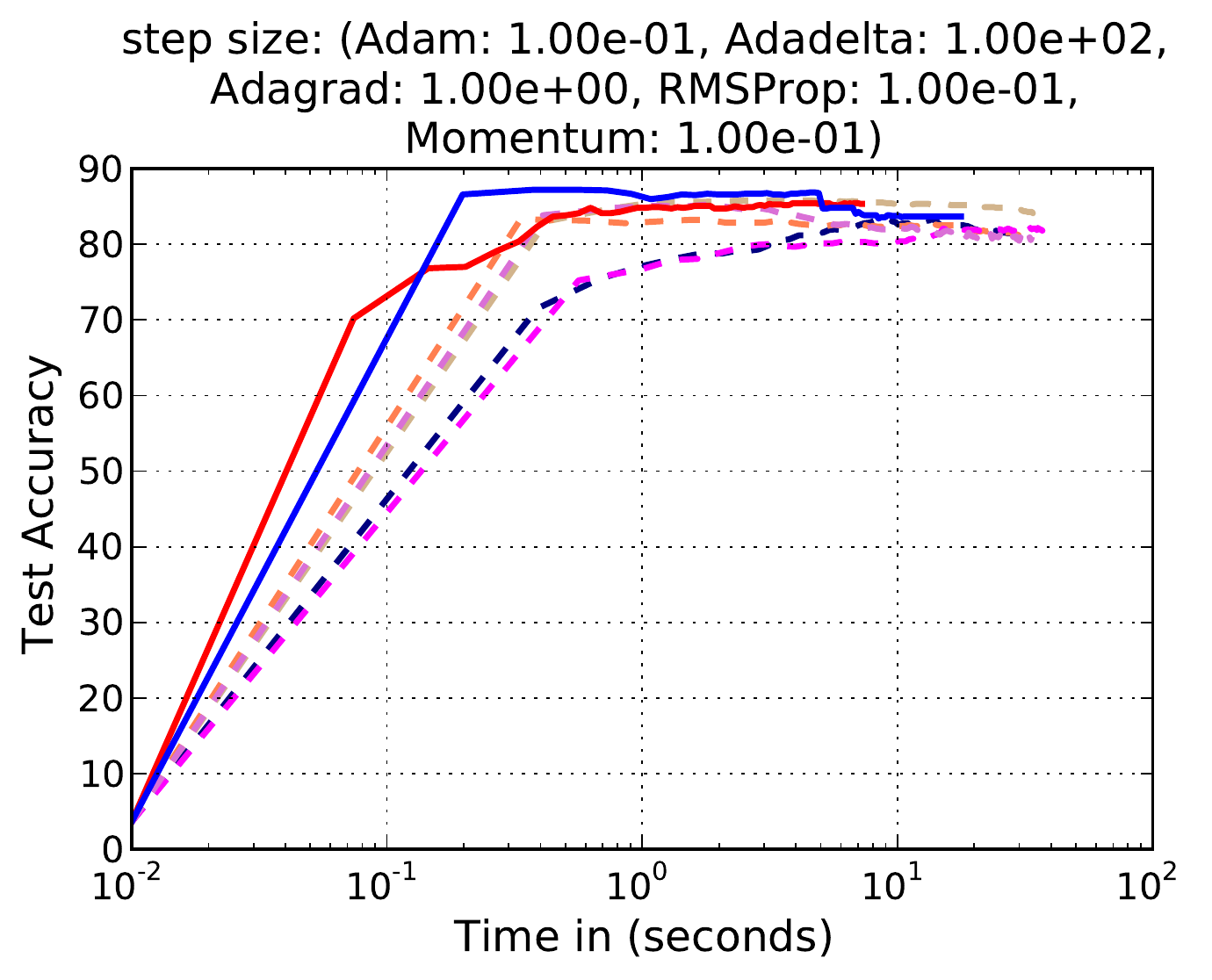}
      } & 
 	\parbox[c]{1.5in}{
      	\includegraphics[width=1.4in, height=0.9in]{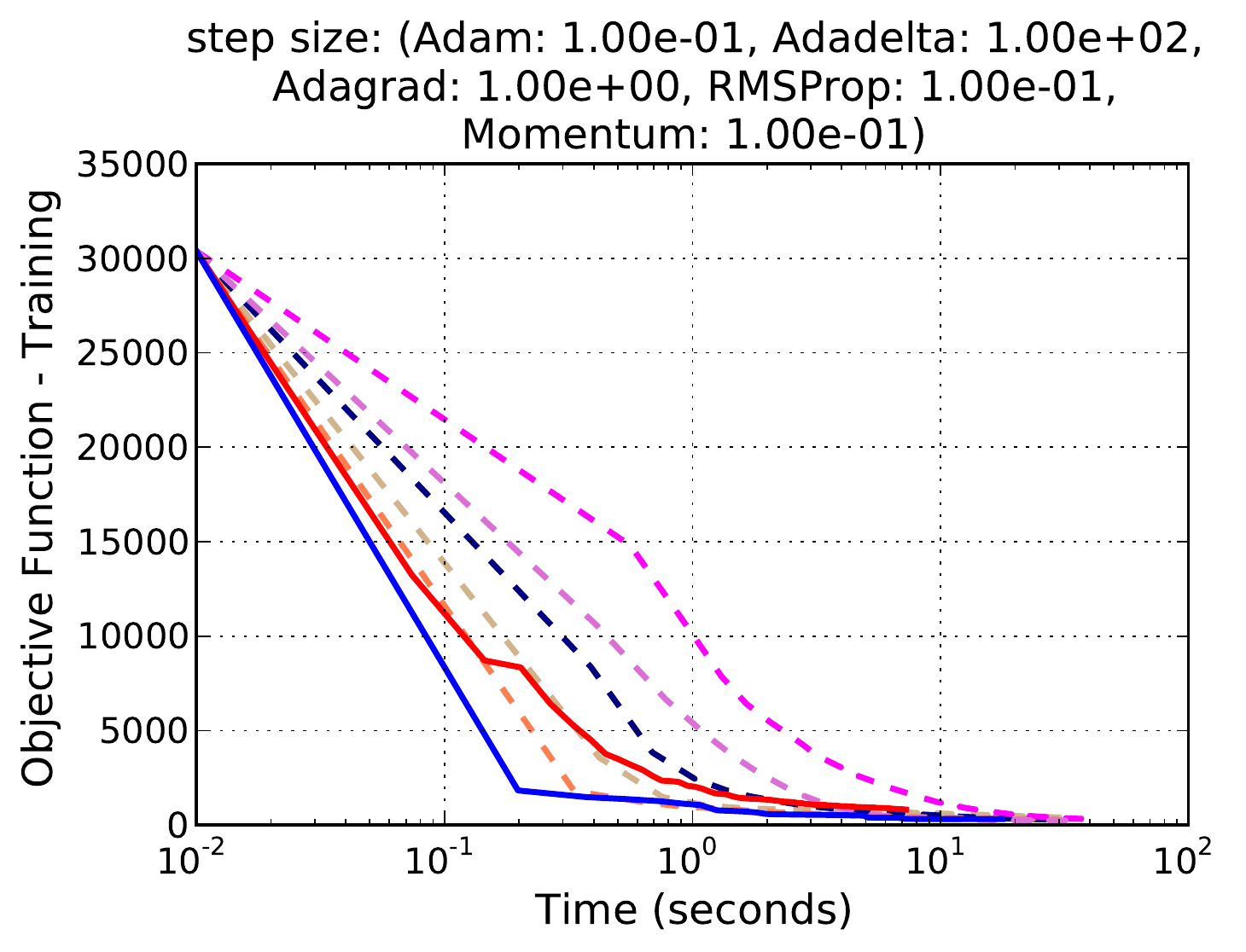}
      } \\
      \multicolumn{4}{c}{newsgroups} \\
 	\parbox[c]{1.5in}{
      	\includegraphics[width=1.4in, height=0.9in]{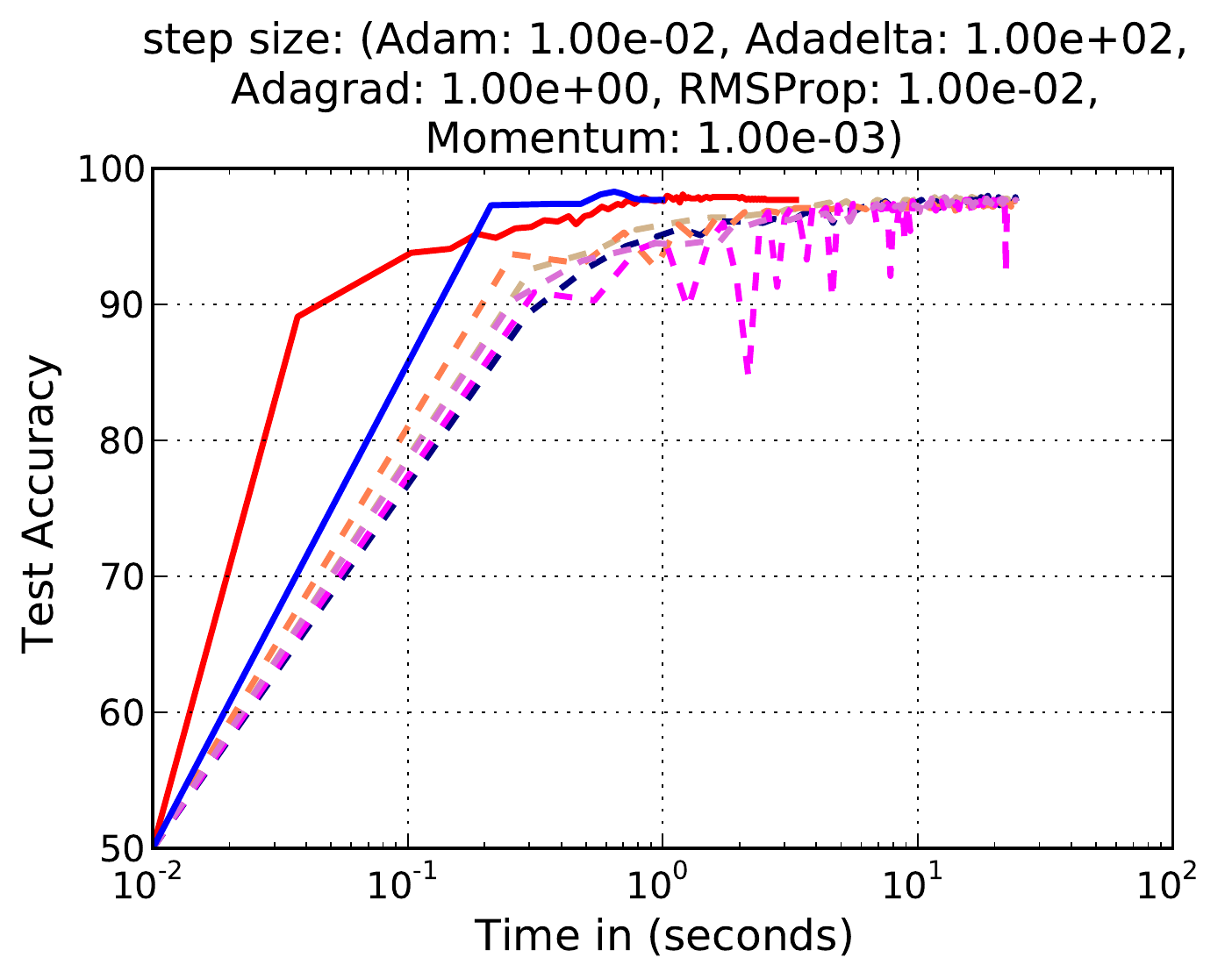}
      } & 
 	\parbox[c]{1.5in}{
      	\includegraphics[width=1.4in, height=0.9in]{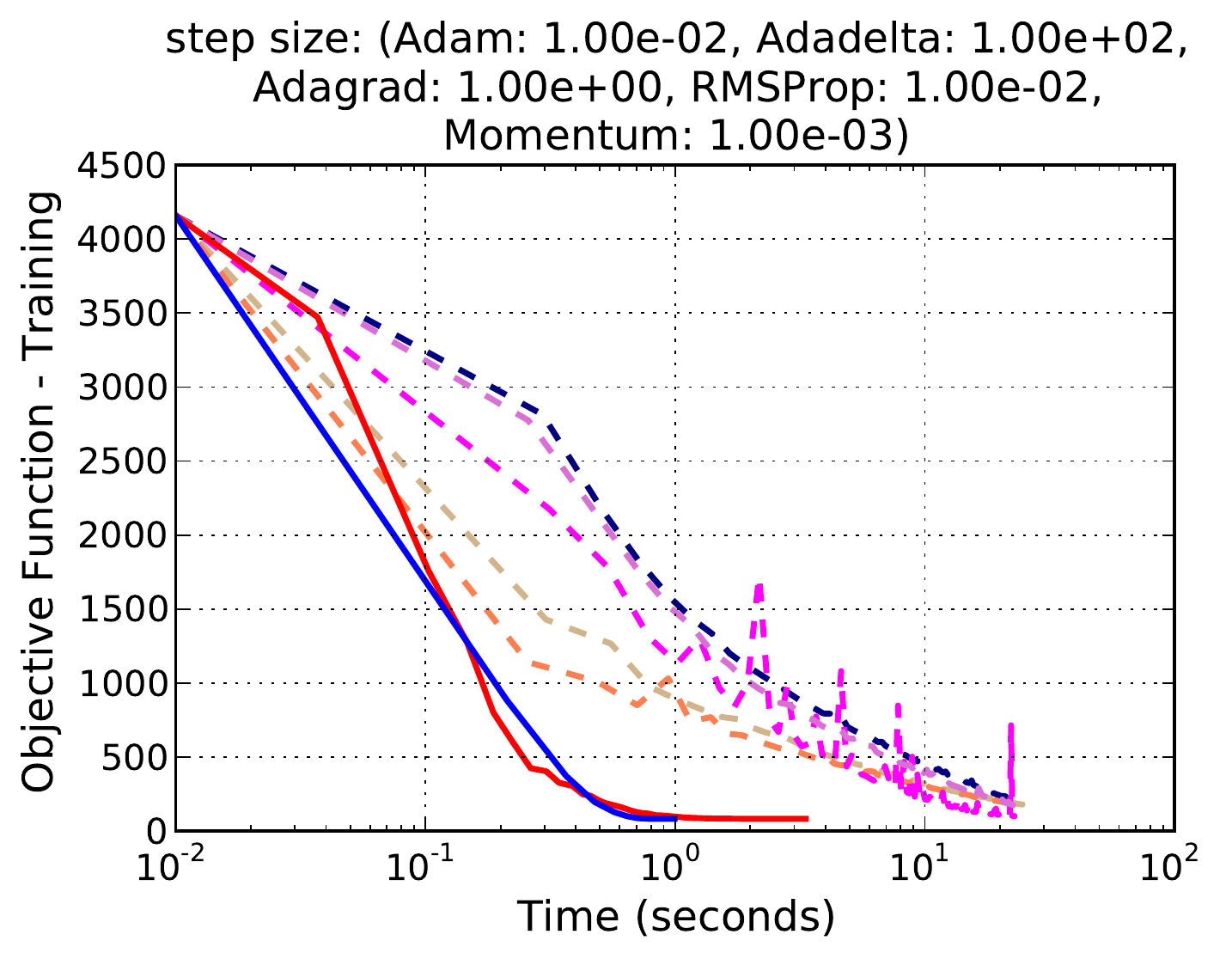}
      } & 
 	\parbox[c]{1.5in}{
      	\includegraphics[width=1.4in, height=0.9in]{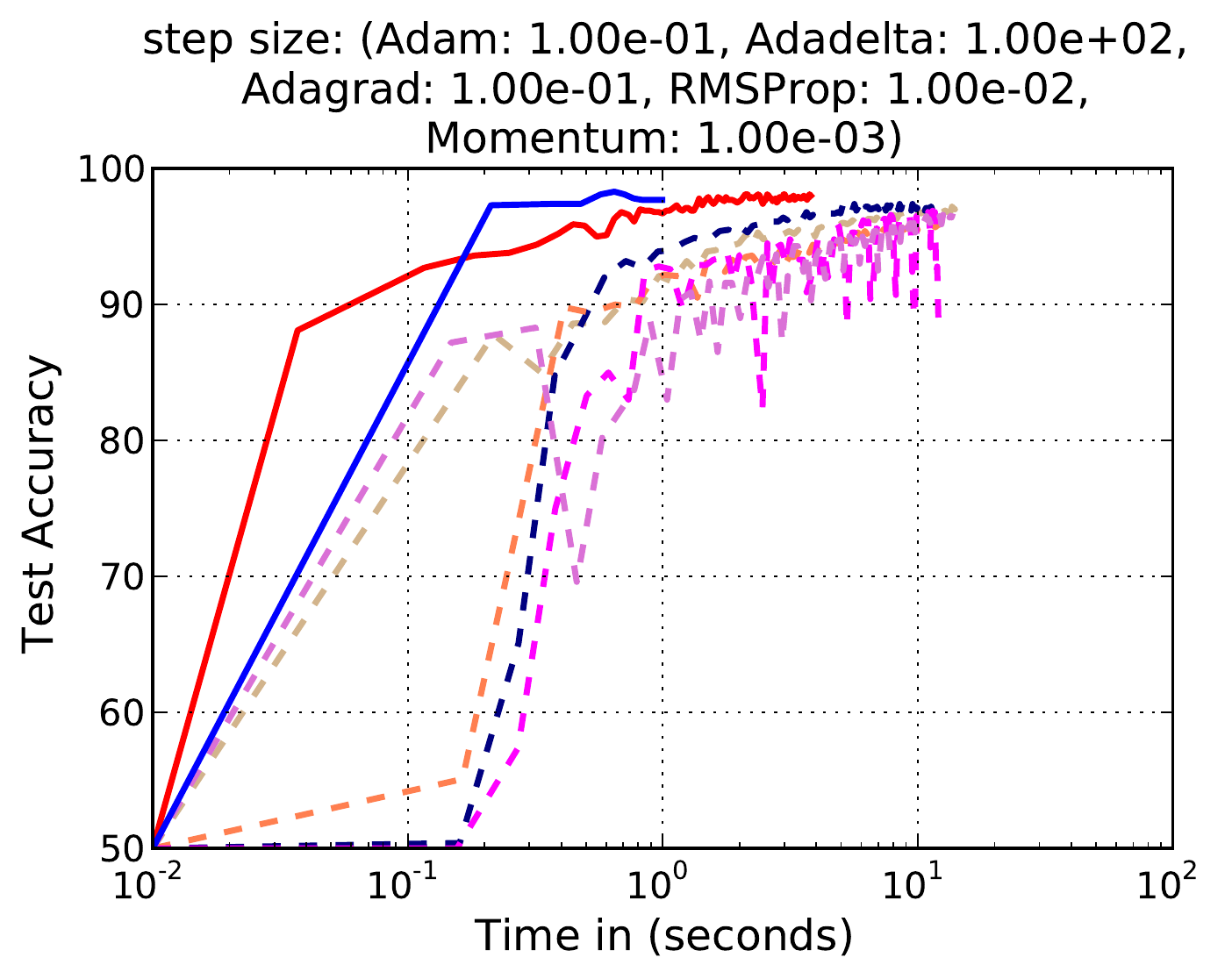}
      } & 
 	\parbox[c]{1.5in}{
      	\includegraphics[width=1.4in, height=0.9in]{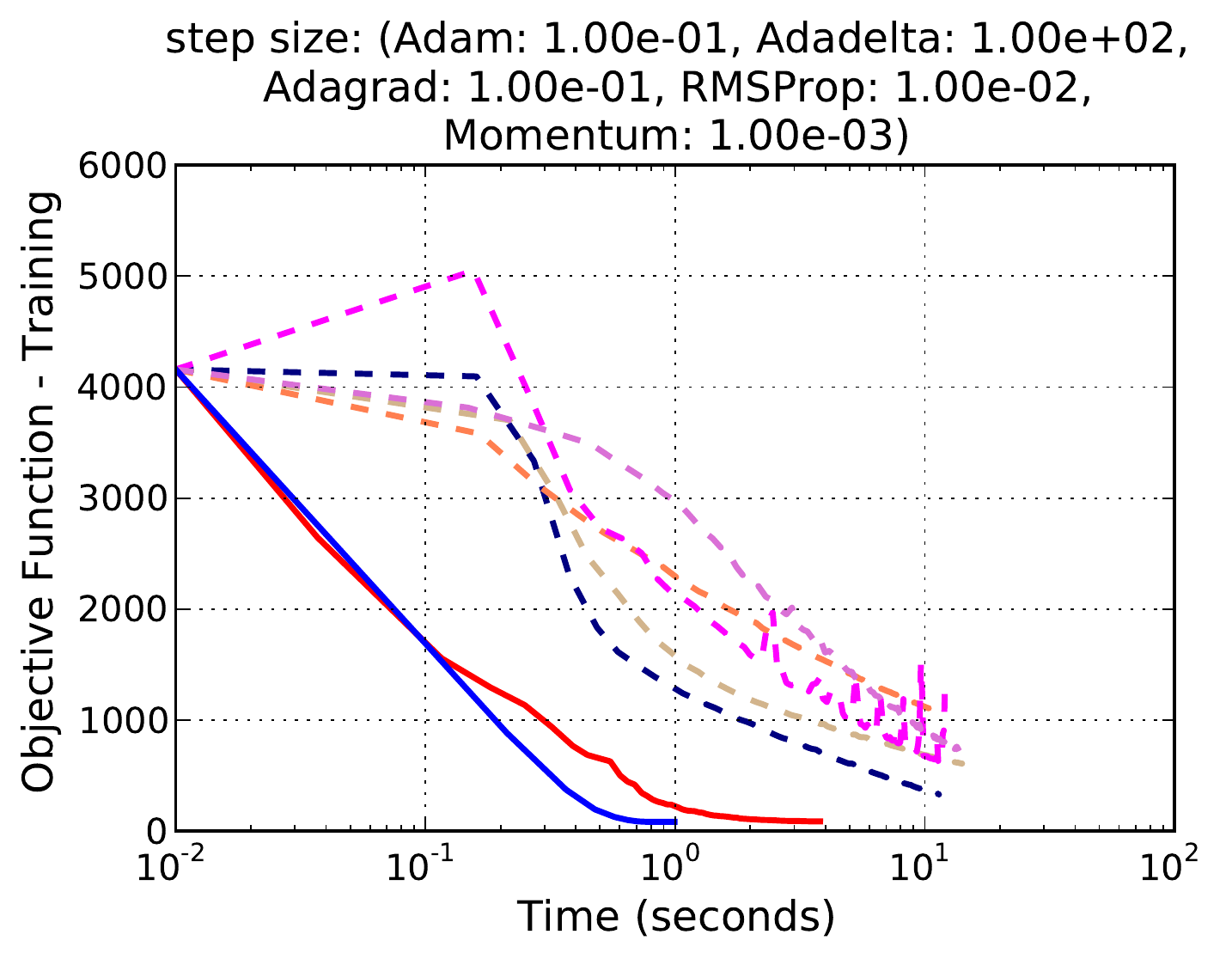}
      } \\
      \multicolumn{4}{c}{Gisette} \\
 	\parbox[c]{1.5in}{
      	\includegraphics[width=1.4in, height=0.9in]{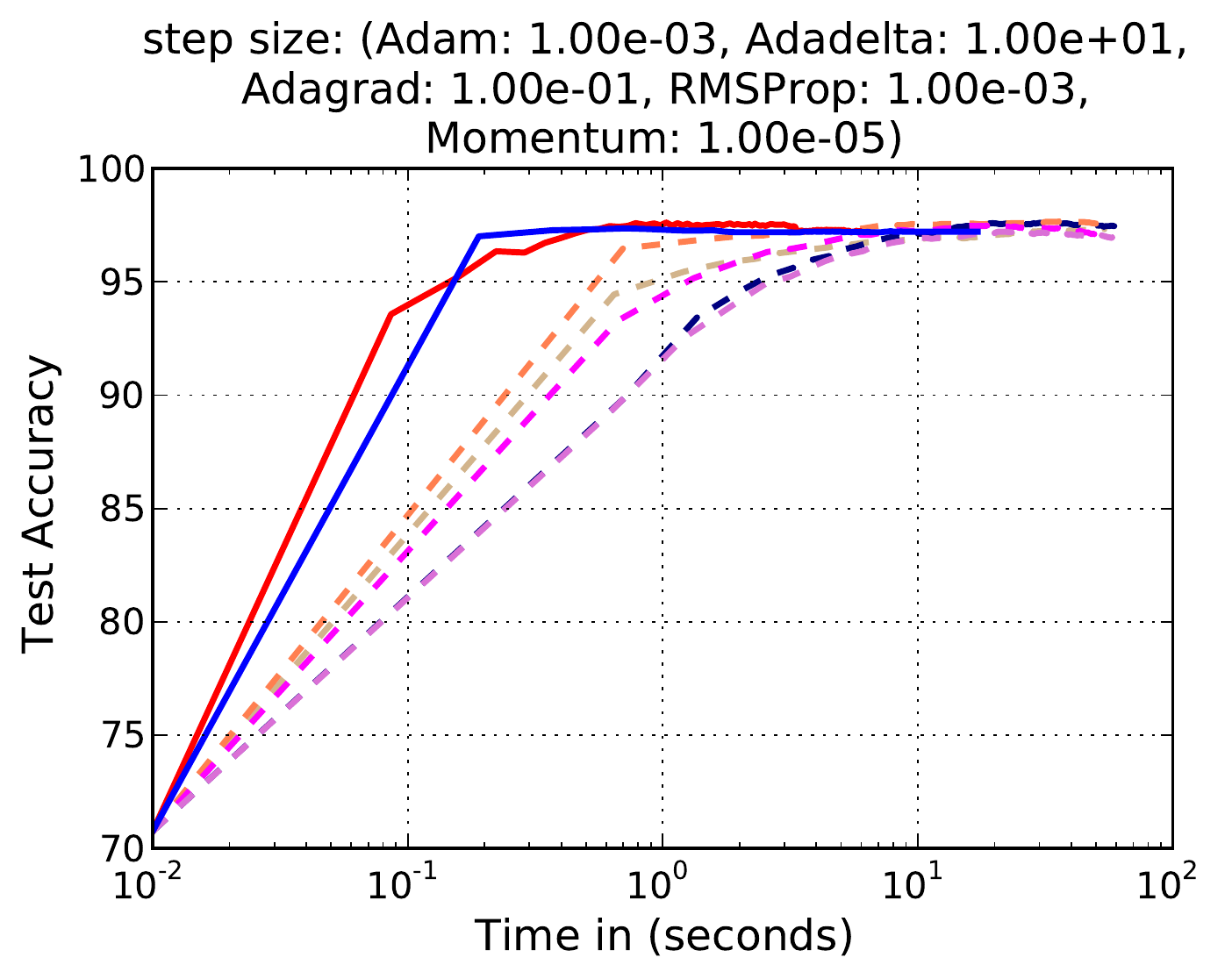}
      } & 
 	\parbox[c]{1.5in}{
      	\includegraphics[width=1.4in, height=0.9in]{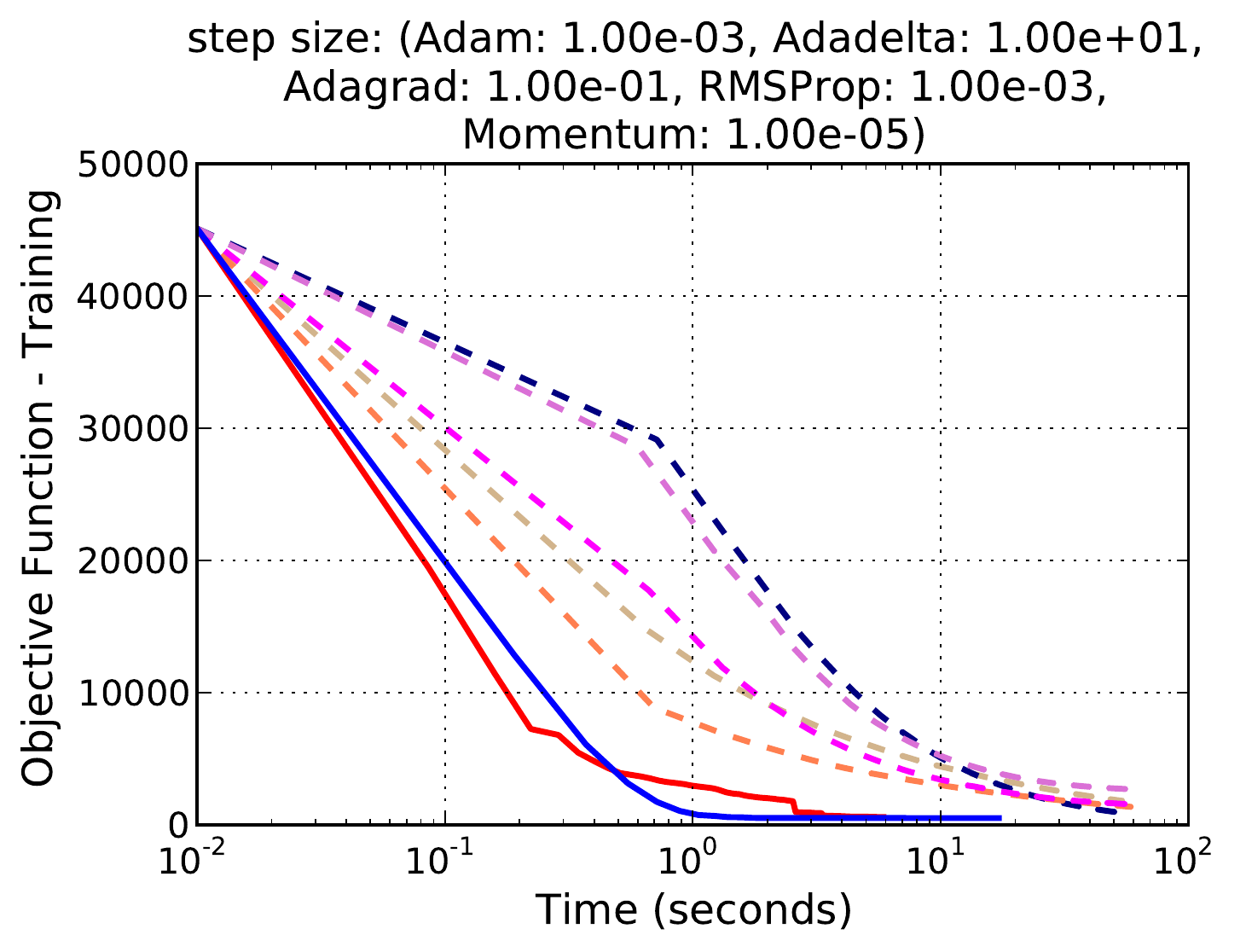}
      } & 
 	\parbox[c]{1.5in}{
      	\includegraphics[width=1.4in, height=0.9in]{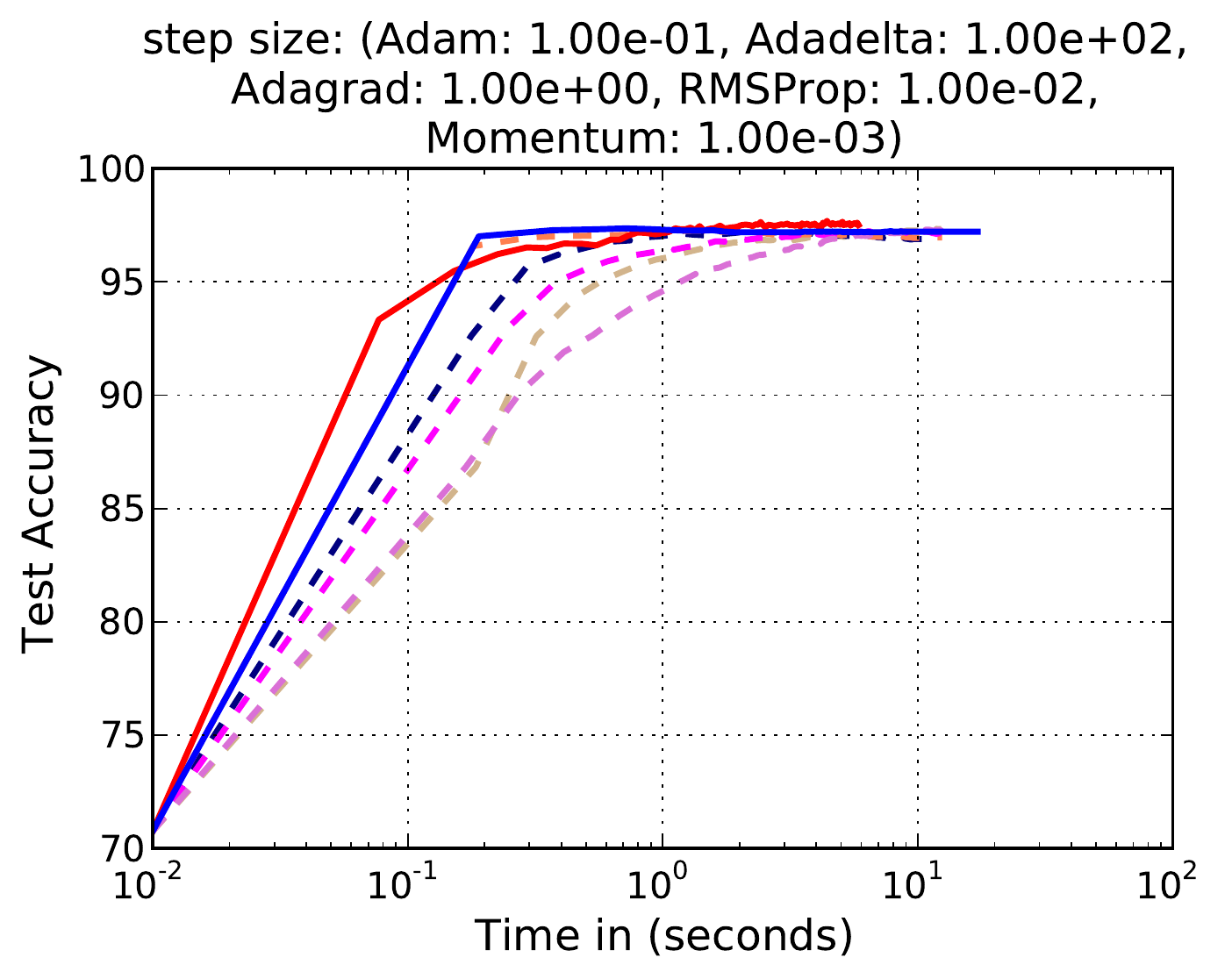}
      } & 
 	\parbox[c]{1.5in}{
      	\includegraphics[width=1.4in, height=0.9in]{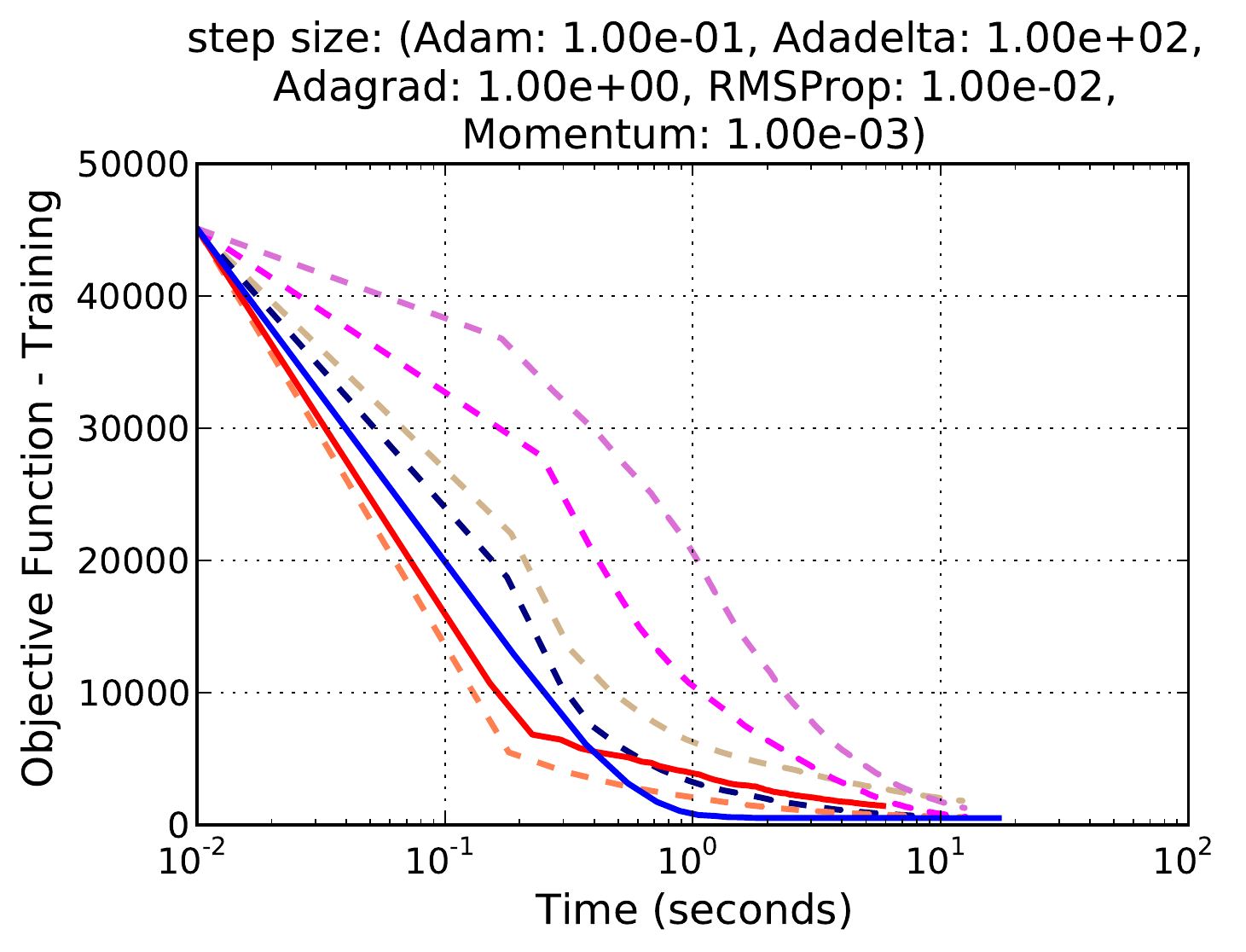}
      } \\
      \multicolumn{4}{c}{real-sim} \\
\end{tabular}
}
\end{table*}

Table~\ref{table-normalized-uniform} presents all the performance results. 
Columns \textit{1} and \textit{3} show the plots for \textit{cumulative-time vs. test-accuracy}
and columns \textit{2} and \textit{4} plot the numbers for \textit{cumulative-time vs. objective function (training)}. 
Please note that x-axis in all the plots is in ``log-scale''. 


%
%

%
%
%
%

\subsubsection{Covertype Dataset}

The first row in Table~\ref{table-normalized-uniform} shows the plots for \textit{Covertype} dataset. 
From the first two columns (batch size 128), we note the following: 
(i) Newton-type methods minimize the objective function to $\approx 3.4e5$ in a smaller
time interval (\textit{FullNewton}: 0.9 secs, \textit{SubsampledNewton-20}: 0.24 secs ), 
compared to first-order alternatives
(Adadelta - 91 secs, Adagrad - 183 secs, Adam - 57 secs, Momentum - 285 secs, RMSProp - 40 secs); 
(ii) Compared to first order algorithms, Newton-type methods achieve equivalent test accuracy, $ 68\% $, in a significantly shorter time interval, i.e.,  0.9 secs compared with tens of seconds for first order methods (Adadelta: 201 secs, Adagrad: 72 secs, Adam: 285 secs, Momentum: 128 secs, RMSProp: 111 secs);
(iii) \textit{SubsampledNewton-100} achieves relatively higher test accuracy
earlier compared to the \textit{FullNewton} method in a relatively short time interval (\textit{FullNewton}: 68\% in 1.5 secs, \textit{SubsampledNewton-100}: 68\% in 204 millisecs). 
For well-conditioned problems (such as this one), a relaxed \textit{CG-tolerance} and small sample sizes (5\% Hessian sample size) yield desirable results quickly.

Columns 3 and 4 present the performance of first-order methods with batch size 20\%.
Randomized Newton method, \textit{SubsampledNewton-20}, achieves higher test accuracy, 68\%, 
in a very short time, 1.05 secs, compared to any of the first order methods as shown in column 3
(Adadelta: 65\% in 21 secs, Adagrad: 65\% in 19 secs, Adam: 68\% in 20 secs, Momentum: 68\% in 18 secs, RMSProp: 65\% in 21 secs).
First order methods, with batch size 20\%, are executed on GPUs resulting in smaller time-per-epoch; see~\ref{sec:platform-cpu-gpu}. 
This can be attributed to processing larger batches of the dataset by the GPU-cores, yielding higher efficiency.

\subsubsection{Drive Diagnostics Dataset}

Results for the \textit{Drive Diagnostics} dataset are shown in the second row of Table~\ref{table-normalized-uniform}.
These plots clearly indicate that Newton-type methods achieve their lowest objective function value 
, 3.75e4, much earlier compared to first order methods (\textit{FullNewton} - 1.3 secs, 
\textit{SubsampledNewton-20} - 0.8 secs, \textit{SubsampledNewton-100} - 0.2 secs). Corresponding times
for batch size 128 for first order methods are : Adadelta - 16 secs, Adagrad - 34 secs, Adam - 25 secs, 
Momentum - 32 secs, RMSProp - 35 secs (lowest objective function value for these methods
are $ \approx $ 3.8e5). For batch size 20\%, except for Adadelta and Momentum, other first order methods achieve their lowest
objective function values, which are significantly higher compared to Newton-type methods, in $ \approx $ 3 seconds. 
Momentum is the only first order method that achieves almost equivalent objective function value, 
3.8e5 in 0.6 seconds, as Newton-type methods.

All first order methods, with batch size 128, achieve test accuracy of 87\% which is same as Newton-type methods 
but take much longer: \textit{FullNewton} - 0.2 secs, \textit{SubsampledNewton-20} - 0.3 secs, 
\textit{SubsampledNewton-100} - 0.15 secs vs. Adadelta - 30 secs,  Adagrad - 36 secs, Adam - 7 secs, Momentum - 32 secs, 
RMSProp - 7 secs. Here, except Momentum, none of the first order methods with 
batch size 20\% achieve 87\% test accuracy in 100 epochs. 

\subsubsection{MNIST and CIFAR-10 Datasets}


Rows 3 and 4 in Table~\ref{table-normalized-uniform} present plots for \textit{MNIST} and {CIFAR-10} datasets, 
respectively. 
Regardless of the batch size, Newton-type methods clearly outperform first-order methods.
For example, with \textit{MNIST} dataset, all the methods achieve a test accuracy of 92\%. However, Newton-type methods do so in $ \approx 0.2 $ seconds, compared to $ \approx 4$ seconds for first order methods with batch size of 128. 


\textit{CIFAR} results are shown in row 4 of Table~\ref{table-normalized-uniform}. We clearly notice that first order
methods, with batch size 128, make slow progress towards achieving their lowest objective function value (and 
test accuracy) taking almost 100 seconds to reach 8.4e4 (40\% test accuracy). Newton-type methods achieve 
these values in significantly shorter time (\textit{FullNewton} - 10 seconds, \textit{SubsampledNewton-20} - 4.2 seconds, 
\textit{SubsampledNewton-100} - 2.6 seconds). The slow progress of first order methods is much more pronounced 
when batch size is set to 20\%. Only Adam and Momentum methods achieve a test accuracy of $ \approx $ 40\% in 100 epochs (taking $ \approx $ 60 seconds). Note that \textit{CIFAR-10} represents a relatively
\textit{ill}-conditioned problem. As a result, in terms of lowering the objective function on \textit{CIFAR-10}, first-order methods are negatively affected by the ill-conditioning, whereas all Newton-type methods show a great degree of robustness. 
This demonstrates the versatility of  Newton-type methods for solving problems with various degrees of ill-conditioning.

\subsubsection{Newsgroups20 Dataset}
Plots in row 5 of Table~\ref{table-normalized-uniform} correspond to \textit{Newsgroups20} dataset.
This is a sparse dataset, and the largest in the scope of this work (the Hessian is $\approx$ 1e6 $\times$ 1e6).
Here, \textit{FullNewton} and \textit{SubsampledNewton-100} achieve, respectively, 87.22\% and 88.46\% test accuracy in the first few iterations. Smaller batch sized first order methods can only achieve a maximum test accuracy of 85\% in 100 epochs. Note that average
time per epoch for first order methods is $ \approx $ 1 sec compared to 75 millisecs for \textit{SubsampledNewton-100} iteration.
When 20\% gradient is used, as shown in column 3, we notice that the \textit{SubsampledNewton-20} method starts with a lower test accuracy of $ \approx $ 80\%  in the 5th iteration and slowly ramps up to 85.4\% as we near the allotted number of iterations. This can be attributed to a smaller gradient sample size, and sparse nature of this dataset.


%
%



\subsubsection{Gisette and Real-Sim Datasets}

Rows 6 and 7 in Table~\ref{table-normalized-uniform} show results for \textit{Gisette} and \textit{Real-Sim} datasets, respectively. 
\textit{FullNewton} method for \textit{Gisette} dataset converges in 11 iterations and yields 98.3\% test accuracy in 0.6 seconds.
\textit{SubsampledNewton-100} takes 34 iterations to reach 98\% test accuracy, whereas first order counterparts, except Momentum
method, can achieve 97\% test accuracy in 100 iterations. When batch size is set to 20\%, we notice that all first order methods
make slow progress towards achieving lower objective function values. Noticeably, none of the first order methods
can lower the objective function value to a level achieved by Newton-type methods, which can be attributed to the ill-conditioning of this problem; see Table \ref{table:datasets}. 

For \textit{Real-Sim} dataset, relative to first order methods and regardless of batch size, we clearly notice that Newton-type methods
achieve similar or lower objective function values, in a comparable or lower time interval. 
Further, \textit{FullNewton} achieves 97.3\% in the $ 2^{nd} $ iteration whereas it takes 11 iterations for \textit{SubsampledNewton-20}.

\subsection{Sensitivity to Hyper-Parameter Tuning}
The ``biggest elephant in the room'' in optimization using, almost all, first-order methods is that of fine-tuning of various underlying hyper-parameters, most notably, the step-size~\cite{berahas2017investigation,xuNonconvexEmpirical2017}. Indeed, the success of most such methods is tightly intertwined with many trial and error steps to find a proper parameter settings. It is highly unusual for these methods to exhibit acceptable performance on the first try, and it
often takes many trials and errors before one can see reasonable results. In fact, the ``true training time'', which almost
always includes the time it takes to appropriately tune these parameters, can be frustratingly long. 
In contrast, second-order optimization methods involve much less parameter tuning, and are less sensitive to specific choices of their hyper-parameters~\cite{berahas2017investigation,xuNonconvexEmpirical2017}.

\begin{figure*}[htbp]
	\centering
	\subfigure[SGD with Momentum]{
		\includegraphics[width=0.2\textwidth]
		{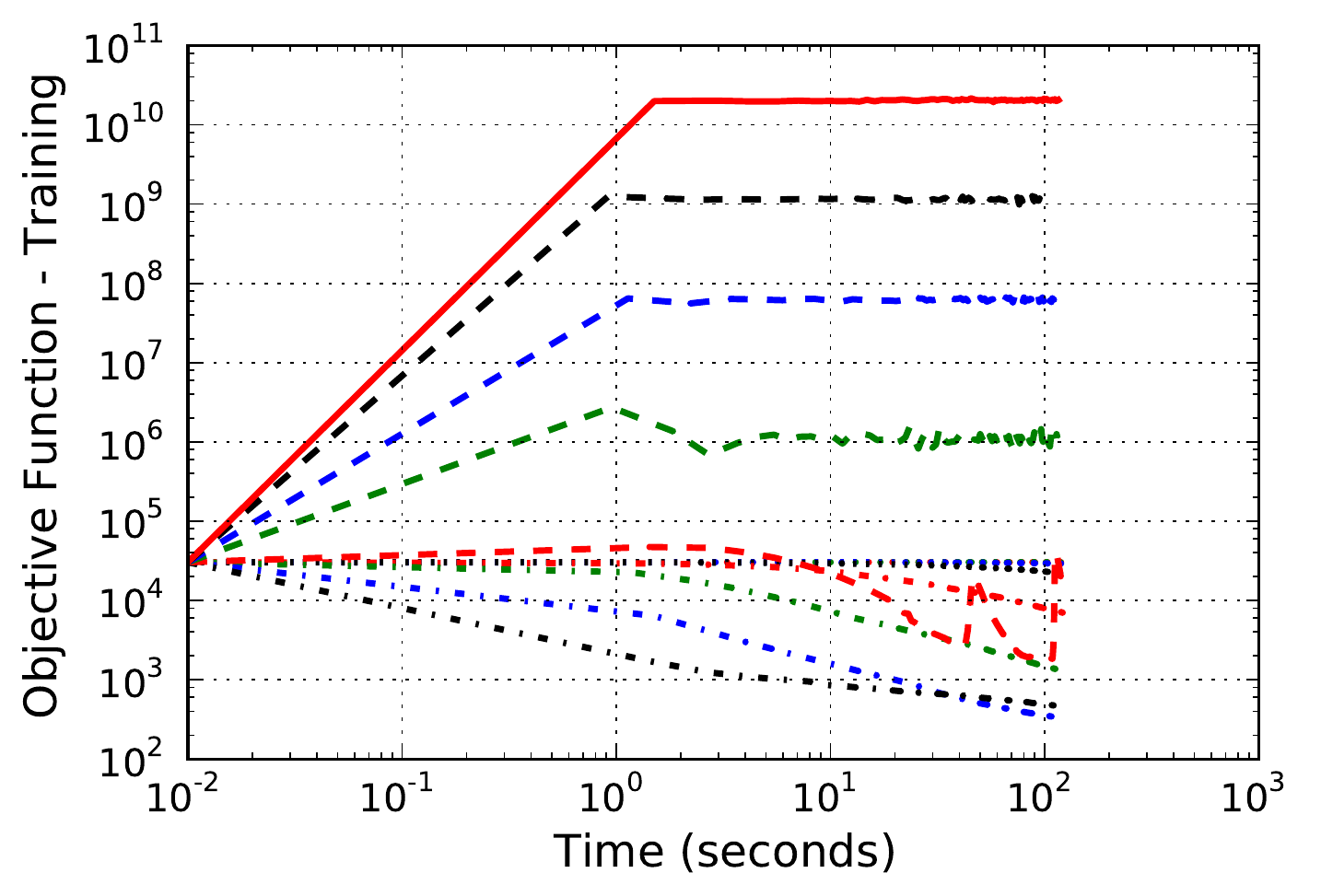}
	}
	\subfigure[Adagrad]{
		\includegraphics[width=0.2\textwidth]
		{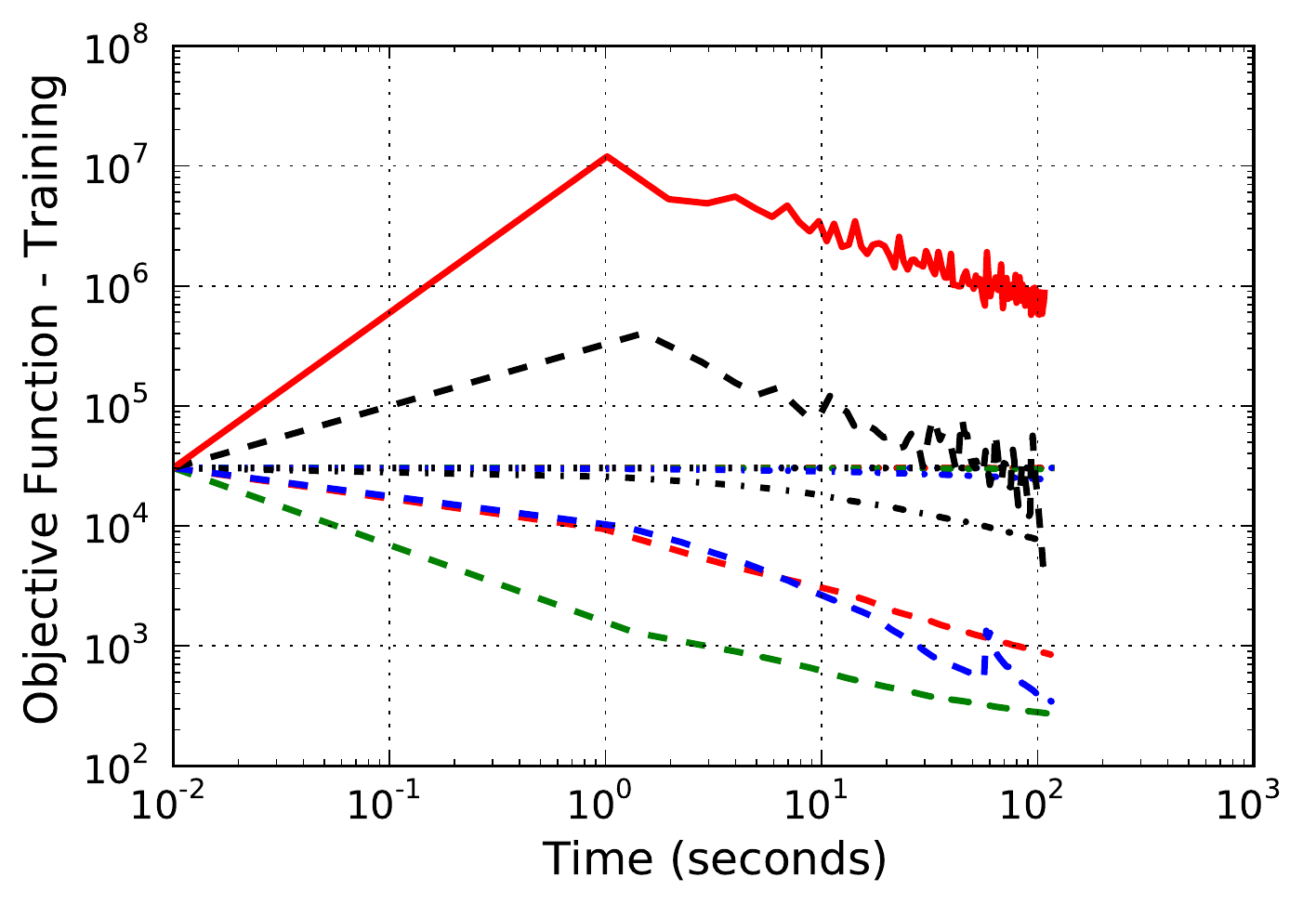}
	}
	\subfigure[RMSProp]{
		\includegraphics[width=0.2\textwidth]
		{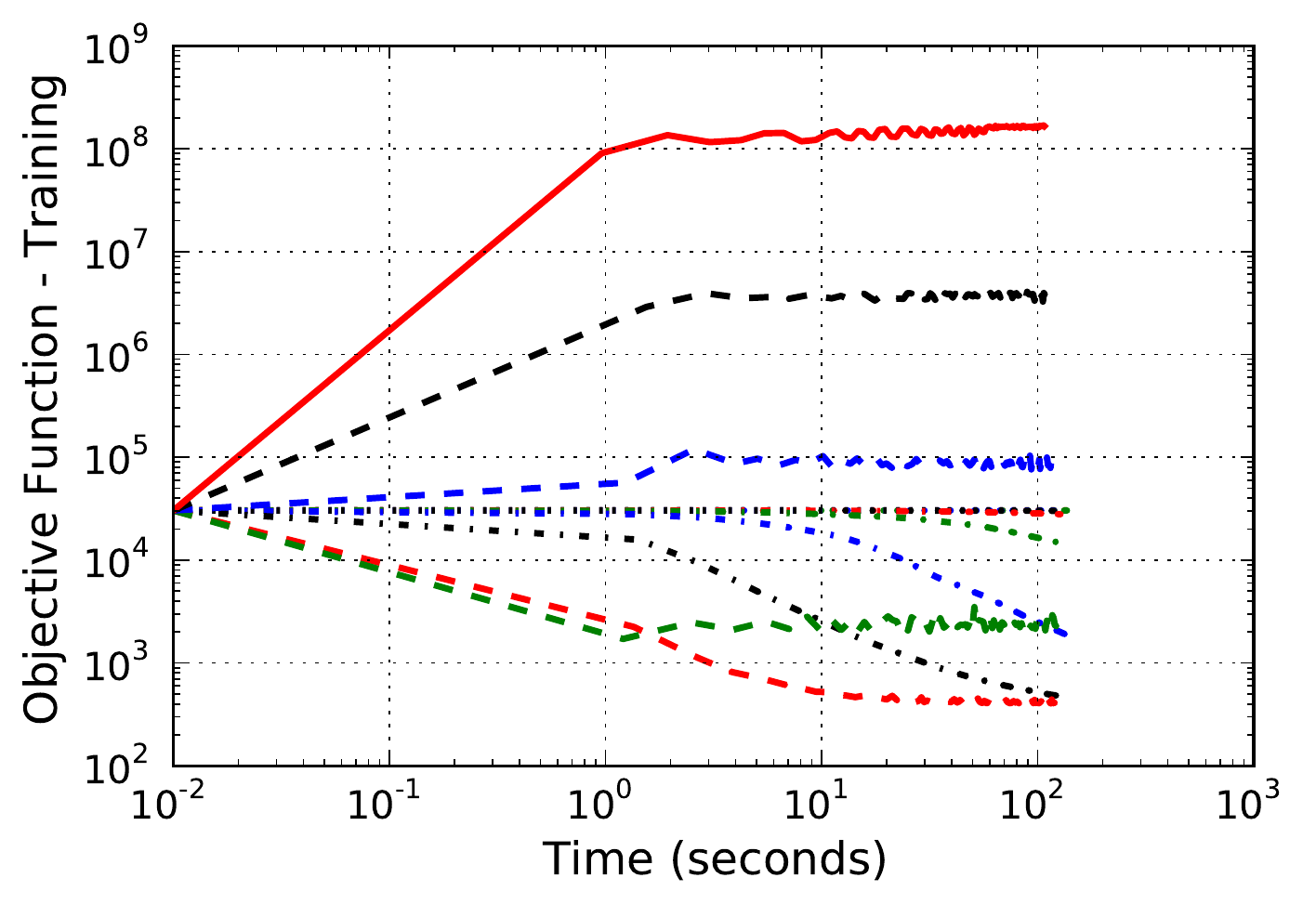}
	}
	\subfigure[Adam]{
		\includegraphics[width=0.2\textwidth]
		{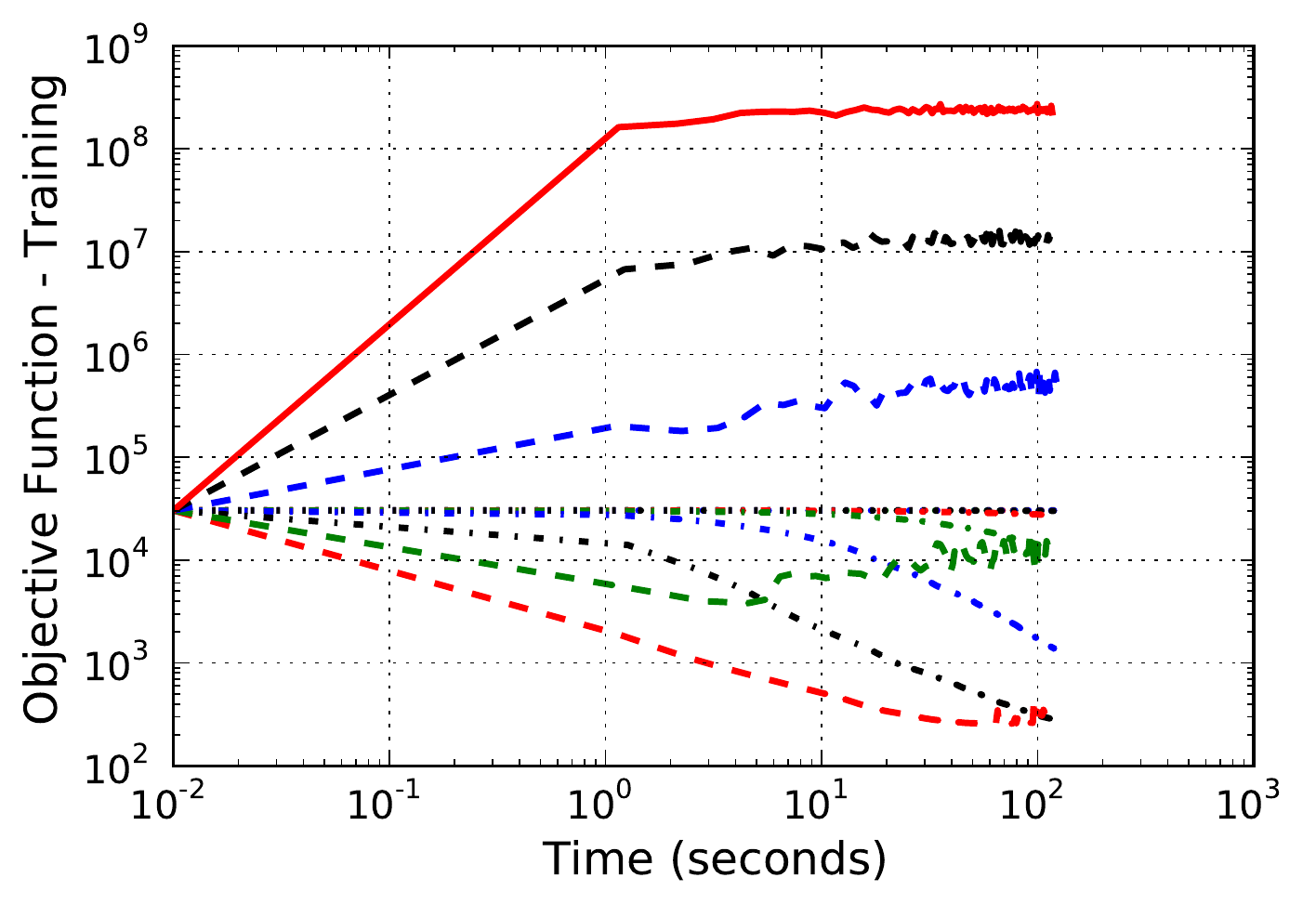}
	}
	\subfigure{
		\includegraphics[width=0.07\textwidth]
		{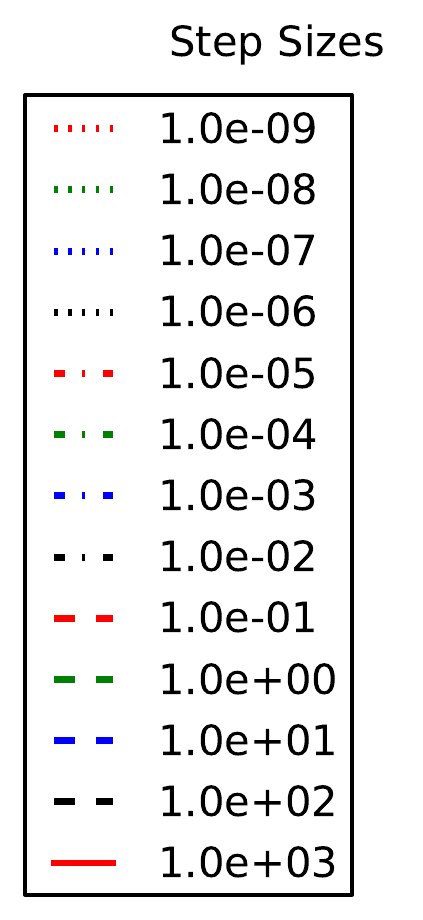}
	}
	\caption{Sensitivity of various first-order methods with respect to the choice of the step-size, i.e., learning-rate. It is clear that, too small a step-size can lead to slow convergence, while larger step-sizes cause the method to diverge. The range of step-sizes for which some of these methods perform reasonably, can be very narrow. This is contrast with Newton-type, which come with a priori ``natural'' step-size, i.e., $ \alpha = 1 $ , and only occasionally require the line-search to intervene}
	\label{fig:sensitivity}
\end{figure*}

Here, to further highlight such issues, we demonstrate the sensitivity of several first-order methods with respect to their learning rate. Figure~\ref{fig:sensitivity} shows the results of multiple runs of SGD with Momentum, Adagrad, RMSProp and Adam on \textit{Newsgroups20} dataset with several choices of step-size. Each method is run 13 times using step-sizes in the range $10^{-6}/L$ to $10^{6}/L$, in increments of $ 10 $, where $ L $ is the Lipschitz constant; see Table \ref{table:datasets}. 

It is clear that small step-sizes can result in stagnation, whereas large step sizes can cause the method to diverge. Only if the step-size is within a particular and often narrow range, which greatly varies across various methods, one can see reasonable performance. 

\begin{remark}
For some first-order methods, e.g., momentum based, line-search type techniques simply cannot be used. For others, the starting step-size for line-search is, almost always, a priori unknown. This is sharp contrast with randomized Newton-type methods considered here, which come with a priori ``natural'' step-size, i.e.,  $ \alpha = 1 $ , and furthermore, only occasionally require the line-search to intervene; see~\cite{roosta2016sub_global,roosta2016sub_local} for theoretical guarantees in this regard. 
\end{remark}

\section{Conclusions And Future Work}
\label{sec:conclusions}

In this paper, we demonstrate that sampled variants of Newton's method, when implemented appropriately, present compelling alternatives to popular first-order methods for solving convex optimization problems in machine learning and data analysis applications. We discussed, in detail, the GPU-specific implementation of Newton-type methods to achieve similar per-iteration costs as first-order methods. We experimentally showcased their advantages, including robustness to ill-conditioning and higher predictive performance. We also highlighted the sensitivity of various first-order methods with respect to their learning-rate. 

Extending our results and implementations to non-convex optimization problems and targeting broad classes of machine learning applications, is an important avenue for future work.

\bibliographystyle{plain}
\bibliography{newton-cg}

\begin{thebibliography}{10}

\bibitem{abadi2016tensorflow}
Mart{\'\i}n Abadi, Paul Barham, Jianmin Chen, Zhifeng Chen, Andy Davis, Jeffrey
  Dean, Matthieu Devin, Sanjay Ghemawat, Geoffrey Irving, Michael Isard, et~al.
\newblock Tensorflow: A system for large-scale machine learning.
\newblock In {\em OSDI}, volume~16, pages 265--283, 2016.

\bibitem{avron2010blendenpik}
Haim Avron, Petar Maymounkov, and Sivan Toledo.
\newblock Blendenpik: Supercharging {LAPACK}'s least-squares solver.
\newblock {\em SIAM Journal on Scientific Computing}, 32(3):1217--1236, 2010.

\bibitem{bengio1994learning}
Yoshua Bengio, Patrice Simard, and Paolo Frasconi.
\newblock Learning long-term dependencies with gradient descent is difficult.
\newblock {\em IEEE transactions on neural networks}, 5(2):157--166, 1994.

\bibitem{berahas2017investigation}
Albert~S Berahas, Raghu Bollapragada, and Jorge Nocedal.
\newblock {An Investigation of Newton-Sketch and Subsampled Newton Methods}.
\newblock {\em arXiv preprint arXiv:1705.06211}, 2017.

\bibitem{bollapragada2016exact}
Raghu Bollapragada, Richard Byrd, and Jorge Nocedal.
\newblock Exact and inexact subsampled {N}ewton methods for optimization.
\newblock {\em arXiv preprint arXiv:1609.08502}, 2016.

\bibitem{bottou2016optimization}
L{\'e}on Bottou, Frank~E Curtis, and Jorge Nocedal.
\newblock Optimization methods for large-scale machine learning.
\newblock {\em arXiv preprint arXiv:1606.04838}, 2016.

\bibitem{le2004large}
L{\'e}on Bottou and Yann Le{C}un.
\newblock Large scale online learning.
\newblock {\em Advances in neural information processing systems}, 16:217,
  2004.

\bibitem{byrd2012sample}
Richard~H. Byrd, Gillian~M. Chin, Jorge Nocedal, and Yuchen Wu.
\newblock Sample size selection in optimization methods for machine learning.
\newblock {\em Mathematical programming}, 134(1):127--155, 2012.

\bibitem{coates2009scalable}
Adam Coates, Paul Baumstarck, Quoc Le, and Andrew~Y Ng.
\newblock Scalable learning for object detection with gpu hardware.
\newblock In {\em Intelligent Robots and Systems, 2009. IROS 2009. IEEE/RSJ
  International Conference on}, pages 4287--4293. IEEE, 2009.

\bibitem{coates2013deep}
Adam Coates, Brody Huval, Tao Wang, David Wu, Bryan Catanzaro, and Ng~Andrew.
\newblock Deep learning with cots hpc systems.
\newblock In {\em International Conference on Machine Learning}, pages
  1337--1345, 2013.

\bibitem{doas12}
Kees van~den Doel and Uri Ascher.
\newblock Adaptive and stochastic algorithms for {EIT} and {DC} resistivity
  problems with piecewise constant solutions and many measurements.
\newblock {\em SIAM J. Scient. Comput.}, 34:DOI: 10.1137/110826692, 2012.

\bibitem{duchi2011adaptive}
John Duchi, Elad Hazan, and Yoram Singer.
\newblock Adaptive subgradient methods for online learning and stochastic
  optimization.
\newblock {\em The Journal of Machine Learning Research}, 12:2121--2159, 2011.

\bibitem{erdogdu2015convergence}
Murat~A. Erdogdu and Andrea Montanari.
\newblock Convergence rates of sub-sampled newton methods.
\newblock In {\em Advances in Neural Information Processing Systems 28}, pages
  3034--3042. 2015.

\bibitem{friedman2001elements}
Jerome Friedman, Trevor Hastie, and Robert Tibshirani.
\newblock {\em The elements of statistical learning}, volume~1.
\newblock Springer series in statistics Springer, Berlin, 2001.

\bibitem{gittens2016revisiting}
Alex Gittens and Michael~W Mahoney.
\newblock {Revisiting the Nystr{\"o}m method for improved large-scale machine
  learning}.
\newblock {\em The Journal of Machine Learning Research}, 17(1):3977--4041,
  2016.

\bibitem{kingma2014adam}
Diederik Kingma and Jimmy Ba.
\newblock Adam: A method for stochastic optimization.
\newblock {\em arXiv preprint arXiv:1412.6980}, 2014.

\bibitem{newton-cg-download}
Sudhir~B Kylasa.
\newblock Newton-cg cuda implementation download
  (scripts/code/tensorflow-python-scripts).
\newblock {\em https://github.com/kylasa/NewtonCG}, February 2018.

\bibitem{puremd-gpu}
Sudhir~B Kylasa, Hasan~Metin Aktulga, and Ananth~Y Grama.
\newblock Puremd-gpu: A reactive molecular dynamics simulation package for
  gpus.
\newblock {\em Journal of Computational Physics}, 272:343--359, September 2014.

\bibitem{arxiv-newton-cg}
Sudhir~B Kylasa, Farbod Roosta-Khorasani, Michael~W. Mahoney, and Ananth~Y
  Grama.
\newblock Gpu accelerated sub-sampled newton methods.
\newblock {\em
  https://www.cs.purdue.edu/homes/skylasa/papers/newton-cg-arXiv.pdf}, 2018.

\bibitem{mahoney2011randomized}
Michael~W Mahoney.
\newblock Randomized algorithms for matrices and data.
\newblock {\em Foundations and Trends{\textregistered} in Machine Learning},
  3(2):123--224, 2011.

\bibitem{meng2014lsrn}
Xiangrui Meng, Michael~A Saunders, and Michael~W Mahoney.
\newblock {LSRN: A parallel iterative solver for strongly over-or
  underdetermined systems}.
\newblock {\em SIAM Journal on Scientific Computing}, 36(2):C95--C118, 2014.

\bibitem{ngiam2011optimization}
Jiquan Ngiam, Adam Coates, Ahbik Lahiri, Bobby Prochnow, Quoc~V Le, and
  Andrew~Y Ng.
\newblock On optimization methods for deep learning.
\newblock In {\em Proceedings of the 28th international conference on machine
  learning (ICML-11)}, pages 265--272, 2011.

\bibitem{nocedal2006numerical}
Jorge Nocedal and Stephen Wright.
\newblock {\em Numerical optimization}.
\newblock Springer Science \& Business Media, 2006.

\bibitem{raina2009large}
Rajat Raina, Anand Madhavan, and Andrew~Y Ng.
\newblock Large-scale deep unsupervised learning using graphics processors.
\newblock In {\em Proceedings of the 26th annual international conference on
  machine learning}, pages 873--880. ACM, 2009.

\bibitem{roosta2016sub_global}
Farbod Roosta-Khorasani and Michael~W Mahoney.
\newblock {Sub-sampled Newton methods I: globally convergent algorithms}.
\newblock {\em arXiv preprint arXiv:1601.04737}, 2016.

\bibitem{roosta2016sub_local}
Farbod Roosta-Khorasani and Michael~W Mahoney.
\newblock {Sub-sampled Newton methods II: Local convergence rates}.
\newblock {\em arXiv preprint arXiv:1601.04738}, 2016.

\bibitem{rodoas2}
Farbod Roosta-Khorasani, Kees van~den Doel, and Uri Ascher.
\newblock Data completion and stochastic algorithms for {PDE} inversion
  problems with many measurements.
\newblock {\em Electronic Transactions on Numerical Analysis}, 42:177--196,
  2014.

\bibitem{rodoas1}
Farbod Roosta-Khorasani, Kees van~den Doel, and Uri Ascher.
\newblock Stochastic algorithms for inverse problems involving {PDE}s and many
  measurements.
\newblock {\em SIAM J. Scientific Computing}, 36(5):S3--S22, 2014.

\bibitem{shalev2014understanding}
Shai Shalev-Shwartz and Shai Ben-David.
\newblock {\em Understanding machine learning: From theory to algorithms}.
\newblock Cambridge university press, 2014.

\bibitem{sra2012optimization}
Suvrit Sra, Sebastian Nowozin, and Stephen~J Wright.
\newblock {\em Optimization for machine learning}.
\newblock Mit Press, 2012.

\bibitem{sutskever2013importance}
Ilya Sutskever, James Martens, George Dahl, and Geoffrey Hinton.
\newblock On the importance of initialization and momentum in deep learning.
\newblock In {\em International conference on machine learning}, pages
  1139--1147, 2013.

\bibitem{tijmen2012rmsprop}
Tijmen Tieleman and Geoffrey Hinton.
\newblock Lecture 6.5-rmsprop: Divide the gradient by a running average of its
  recent magnitude.
\newblock {\em COURSERA: Neural Networks for Machine Learning}, 4, 2012.

\bibitem{uci-repository}
UCI.
\newblock Uci machine learning repository.
\newblock {\em http://archive.ics.uci.edu/ml/index.php}, 02 2018.

\bibitem{xuNonconvexTheoretical2017}
Peng Xu, Farbod Roosta-Khorasani, and Michael~W. Mahoney.
\newblock {Newton-Type Methods for Non-Convex Optimization Under Inexact
  Hessian Information}.
\newblock {\em arXiv preprint arXiv:1708.07164}, 2017.

\bibitem{xuNonconvexEmpirical2017}
Peng Xu, Farbod Roosta-Khorasani, and Michael~W. Mahoney.
\newblock {Second-Order Optimization for Non-Convex Machine Learning: An
  Empirical Study}.
\newblock {\em arXiv preprint arXiv:1708.07827}, 2017.

\bibitem{xu2016sub}
Peng Xu, Jiyan Yang, Farbod Roosta-Khorasani, Christopher R{\'e}, and Michael~W
  Mahoney.
\newblock {Sub-sampled newton methods with non-uniform sampling}.
\newblock In {\em Advances in Neural Information Processing Systems}, pages
  3000--3008, 2016.

\bibitem{yang2016implementing}
Jiyan Yang, Xiangrui Meng, and Michael~W Mahoney.
\newblock Implementing randomized matrix algorithms in parallel and distributed
  environments.
\newblock {\em Proceedings of the IEEE}, 104(1):58--92, 2016.

\bibitem{zeiler2012adadelta}
Matthew~D Zeiler.
\newblock Adadelta: an adaptive learning rate method.
\newblock {\em arXiv preprint arXiv:1212.5701}, 2012.

\end{thebibliography}

\section{More Details On Softmax Function~\eqref{eq:softmax_log_likelihood}}
\subsection{Relationship to Logistic Regression with $ \pm1 $-labels}
\label{sec:logistics_pm1}
Sometimes, in the literature, for the two-class classification problem, instead of $ \{0,1\} $ the labels are marked as $ \pm1 $. In this case, the corresponding logistic regression is written as
\begin{align*}
&F(\xx)  = \sum_{i=1}^{n} \log \left( 1+e^{-b_{i} \xx^{T} \aa_{i} } \right).
\end{align*}
In this case, we have
\begin{align*}
&F(\xx) = \sum_{i=1}^{n} \log \left( e^{\frac{- \xx^{T} \aa_{i}}{2}} + e^{\frac{\xx^{T} \aa_{i}}{2}} \right) - \frac{b_{i} \xx^{T} \aa_{i}}{2} \\
&= \sum_{i=1}^{n} \log \left(e^{\frac{- \xx^{T} \aa_{i}}{2}} \left( 1 + e^{\xx^{T} \aa_{i}} \right) \right) - \frac{b_{i} \xx^{T} \aa_{i}}{2} \\
&= \sum_{i=1}^{n} \log \left( 1 + e^{\xx^{T} \aa_{i}} \right) - \frac{(1+b_{i}) \xx^{T} \aa_{i}}{2} \\
&= \sum_{i=1}^{n} \log \left( 1 + e^{\xx^{T} \aa_{i}} \right) - \tilde{b}_{i} \xx^{T} \aa_{i},
\end{align*}
where $ \tilde{b}_{i} \in \{0,1\} $. Hence this formulation co-incides with~\eqref{eq:softmax_log_likelihood}. 

\subsubsection{Softmax Multi-Class problem is (strictly) convex}
\label{sec:softmax_convex}
Consider the data matrix $X \in \mathbb{R}^{n \times d}$ where each row, $\aa_{i}^{T}$, is a row vector corresponding to the $i^{th}$ data point. The Hessian matrix can be written as
\begin{align*}
\nabla^{2} \mathcal{L} = \mathbf{X}^{T} \mathbf{W} \mathbf{X},
\end{align*}
where
\begin{align*}
\mathbf{X} &= \begin{bmatrix}
X & 0  & \ldots & 0 \\
0& X  & \ldots & 0 \\
\vdots & & \ddots & \vdots \\
0& 0  & \ldots & X
\end{bmatrix}_{(n\times (C-1)) \times (d \times (C-1))}, \\
\mathbf{W} &= \begin{bmatrix}
W_{1,1} & W_{1,2}  & \ldots & W_{1,C-1} \\
W_{2,1}& W_{2,2}  & \ldots & W_{2,C-1} \\
\vdots & & \ddots & \vdots \\
W_{C-1,1}& W_{C-1,2}  & \ldots & W_{C-1,C-1}
\end{bmatrix},
\end{align*}
and each $W_{c,c}$ and $W_{c,b}$ is a $n \times n$ diagonal matrix corresponding to~\eqref{eq:softmax_hessian_wc_wc} and~\eqref{eq:softmax_hessian_wc_wb}, respectively. 
Note that since 
\begin{align*}
&\left(\frac{e^{\lin{\aa_{i}, \xx_{c}}}}{1+\sum_{c' = 1}^{C-1} e^{\lin{\aa_{i}, \xx_{c'}}}} - \frac{e^{2\lin{\aa_{i}, \xx_{c}}}}{\left(1+\sum_{c' = 1}^{C-1} e^{\lin{\aa_{i}, \xx_{c'}}}\right)^{2}} \right) - \\
&\sum_{\substack{b = 1 \\ b \neq c}}^{C-1} \frac{e^{\lin{\aa_{i}, \xx_{\hat{c}} + \xx_{c}}}}{\left(1+\sum_{c' = 1}^{C-1} e^{\lin{\aa_{i}, \xx_{c'}}}\right)^{2}} \\
& = \left(\frac{e^{\lin{\aa_{i}, \xx_{c}}}}{1+\sum_{c' = 1}^{C-1} e^{\lin{\aa_{i}, \xx_{c'}}}} - \frac{e^{2\lin{\aa_{i}, \xx_{c}}}}{\left(1+\sum_{c' = 1}^{C-1} e^{\lin{\aa_{i}, \xx_{c'}}}\right)^{2}} \right) \\
&  \quad \quad  - \frac{e^{\lin{\aa_{i}, \xx_{c}}}}{1+\sum_{c' = 1}^{C-1} e^{\lin{\aa_{i}, \xx_{c'}}}} \left( \sum_{\substack{b = 1 \\ b \neq c}}^{C-1}  \frac{e^{\lin{\aa_{i}, \xx_{\hat{c}} }}}{1+\sum_{c' = 1}^{C-1} e^{\lin{\aa_{i}, \xx_{c'}}}} \right) \\
& = \left(\frac{e^{\lin{\aa_{i}, \xx_{c}}}}{1+\sum_{c' = 1}^{C-1} e^{\lin{\aa_{i}, \xx_{c'}}}} - \frac{e^{2\lin{\aa_{i}, \xx_{c}}}}{\left(1+\sum_{c' = 1}^{C-1} e^{\lin{\aa_{i}, \xx_{c'}}}\right)^{2}} \right) \\
&  \quad  \quad - \frac{e^{\lin{\aa_{i}, \xx_{c}}}}{1+\sum_{c' = 1}^{C-1} e^{\lin{\aa_{i}, \xx_{c'}}}} \left( 1-\frac{1 + e^{\lin{\aa_{i},\xx_{c}}}}{1+\sum_{c' = 1}^{C-1} e^{\lin{\aa_{i}, \xx_{c'}}}}  \right) \\
& = \frac{e^{\lin{\aa_{i}, \xx_{c}}}}{\left(1+\sum_{c' = 1}^{C-1} e^{\lin{\aa_{i}, \xx_{c'}}}\right)^{2}} > 0,
\end{align*}
the matrix $\mathbf{W}$ is strictly diagonally dominant, and hence it is symmetric positive definite. So the problem is convex (in fact it is strictly-convex if the data matrix $ X $ is full column rank).

\section{Tensorflow's Performance Comparison on Various Compute Platforms}
\label{sec:platform-cpu-gpu}
\begin{table}[!htb]
\centering
\caption{Performance comparison between first-order and second-order methods on CPU-only and 1-GPU-1-CPU-core compute platforms
for \textit{covertype} dataset. 
Batch-size 128 first order methods are compared with second order methods using full gradient and hessian sample size set to 5\%. 
Batch-size 20\% first order methods are compared with second order methods using sample sizes of 20\% and 5\% for gradient 
and hessian computations respectively. } 
\label{table-cpu-gpu-comparison}
\scalebox{0.9}{
\begin{tabular}
      {cccc} \hline 
      Time vs. Accuracy & Time vs. Misfit  & Time vs. Accuracy & Time vs. Misfit \\
      \multicolumn{2}{c}{Batch Size = 128} & \multicolumn{2}{c}{Batch Size = 20\%} \\       
       \multicolumn{2}{c}{Gradient Sample Size = 100\%} & \multicolumn{2}{c}{Gradient Sample Size = 20\%} \\       
      \multicolumn{2}{c}{Hessian Sample Size = 5\%} & \multicolumn{2}{c}{Hessian Sample Size = 5\%} \\       
	\hline
      \multicolumn{4}{c}{  	\parbox[c]{6in}{
      	\includegraphics[width=6in, height=0.5in]{./figures-uniform-cpu/uniform_legend.pdf}
      } }	\\
 	\parbox[c]{1.5in}{
      	\includegraphics[width=1.4in, height=1.1in]{./figures-uniform-cpu/normalized_batch-128_forest_time_test_accuracy_5_100_10.pdf}
      } & 
 	\parbox[c]{1.5in}{
      	\includegraphics[width=1.4in, height=1.1in]{./figures-uniform-cpu/normalized_batch-128_forest_time_train_function_5_100_10.pdf}
      } & 
 	\parbox[c]{1.5in}{
      	\includegraphics[width=1.4in, height=1.1in]{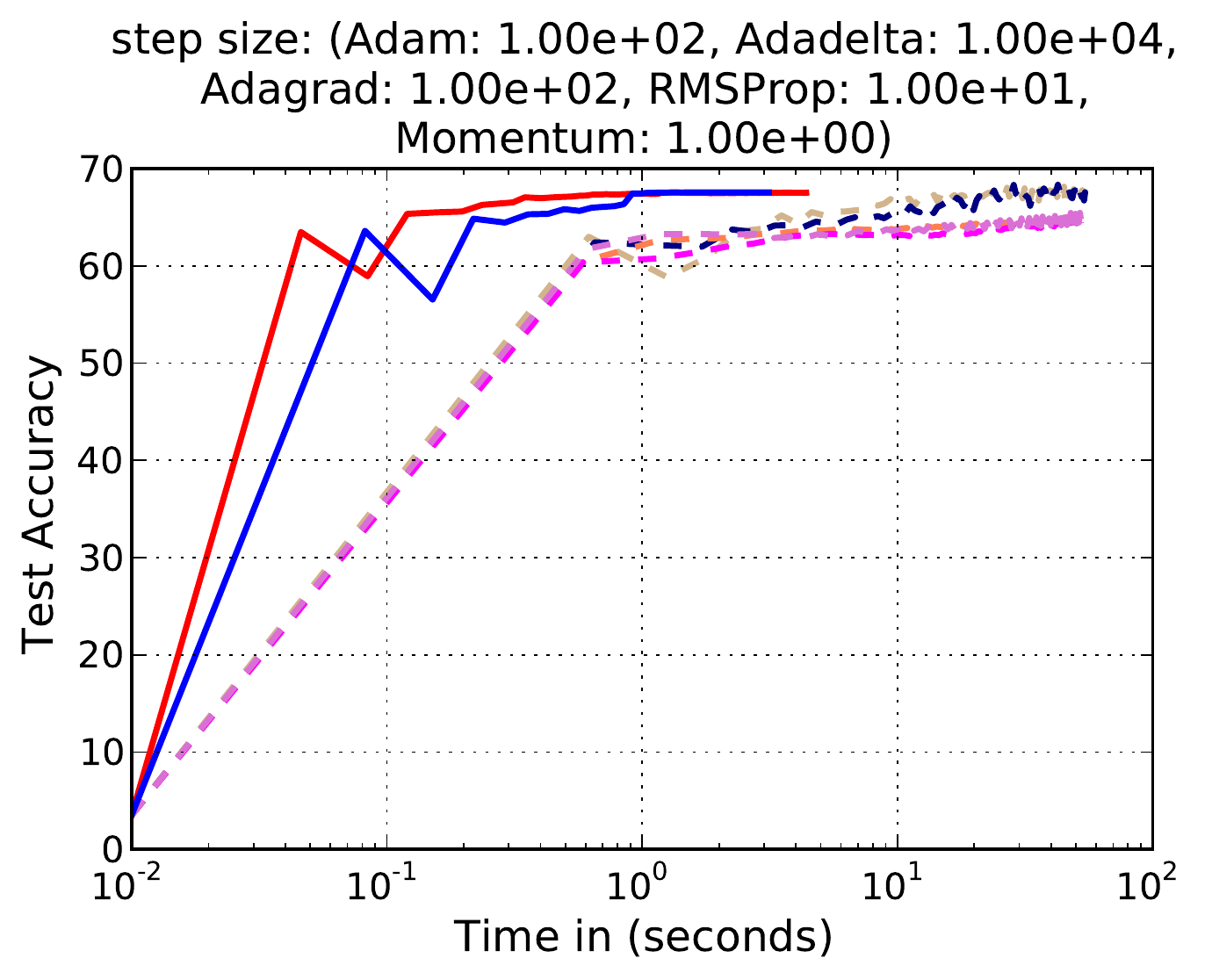}
      } & 
 	\parbox[c]{1.5in}{
      	\includegraphics[width=1.4in, height=1.1in]{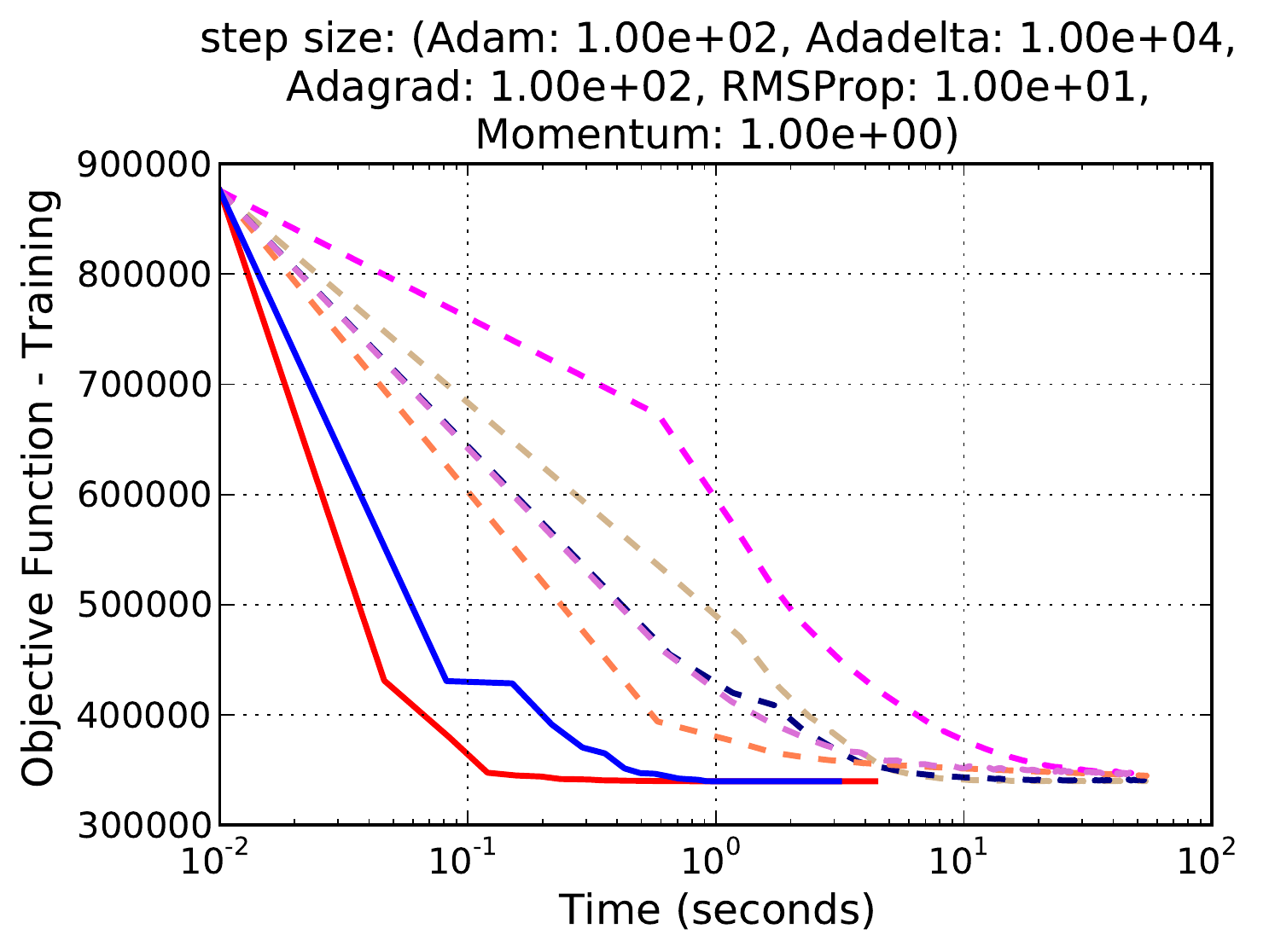}
      } \\
      \multicolumn{4}{c}{Using CPU-only cores for Tensorflow implementations. Newton-type methods use 1-GPU-1-CPU-core.} \\      
      \multicolumn{4}{c}{} \\      
      \multicolumn{4}{c}{} \\      
 	\parbox[c]{1.5in}{
      	\includegraphics[width=1.4in, height=1.1in]{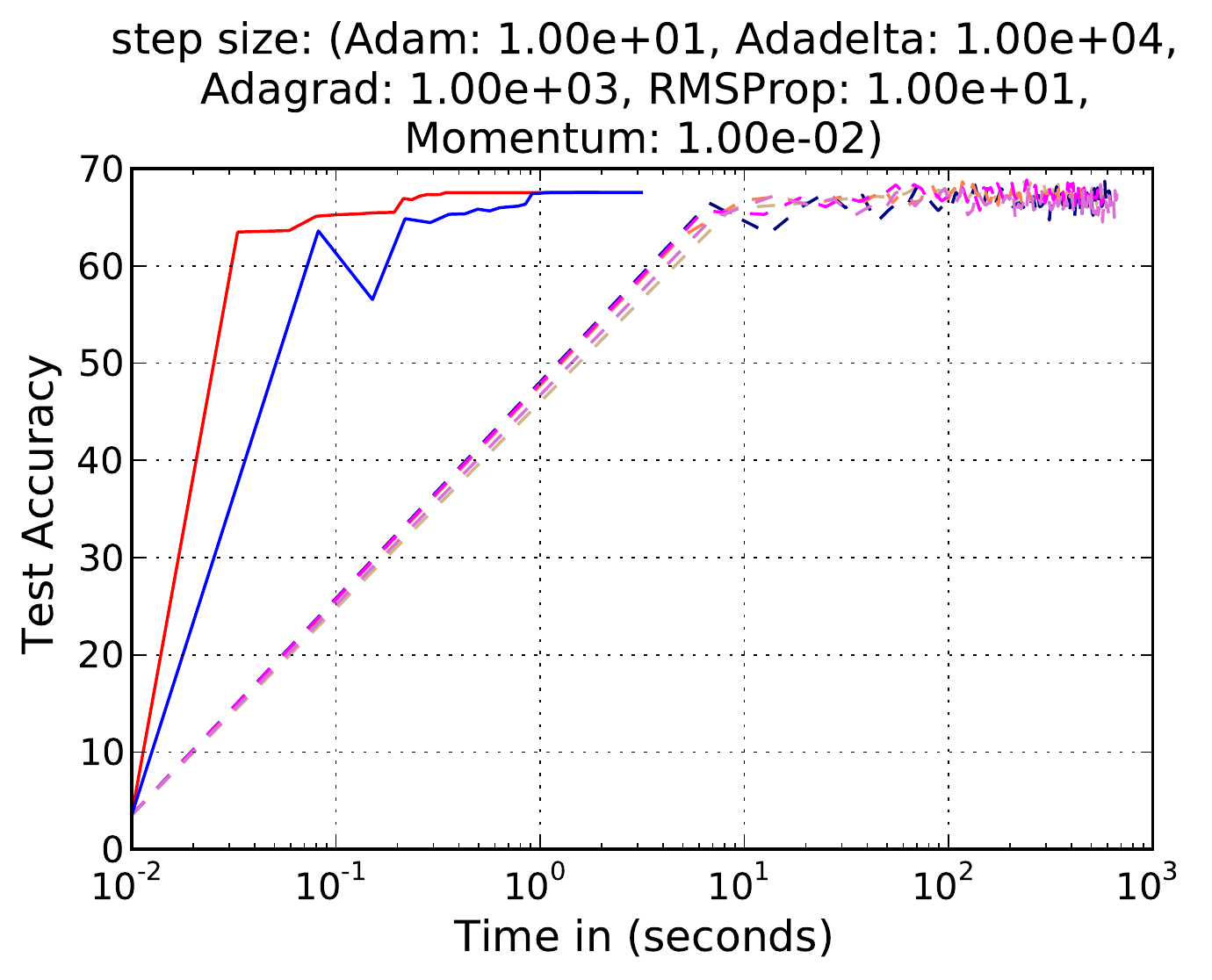}
      } & 
 	\parbox[c]{1.5in}{
      	\includegraphics[width=1.4in, height=1.1in]{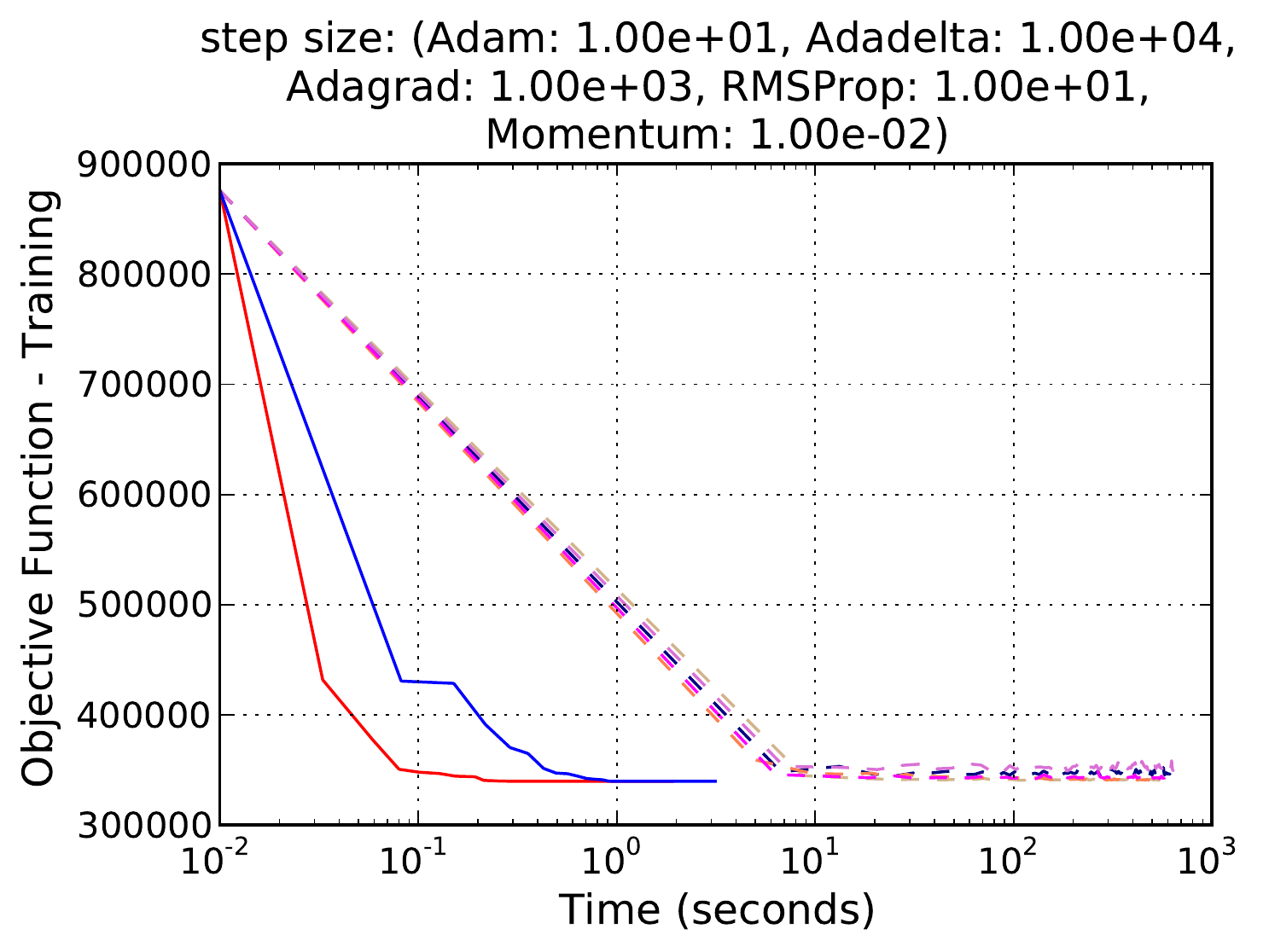}
      } & 
 	\parbox[c]{1.5in}{
      	\includegraphics[width=1.4in, height=1.1in]{./figures-uniform-gpu/normalized_batch-20_forest_time_test_accuracy_5_20_10.pdf}
      } & 
 	\parbox[c]{1.5in}{
      	\includegraphics[width=1.4in, height=1.1in]{./figures-uniform-gpu/normalized_batch-20_forest_time_train_function_5_20_10.pdf}
      } \\
      \multicolumn{4}{c}{Using 1-GPU-1-CPU-core for Tensorflow implementations. Newton-type methods use 1-GPU-1-CPU-core.} \\            
      \multicolumn{4}{c}{} \\      
      \multicolumn{4}{c}{} \\      \hline
\end{tabular}
}
\end{table}

%
%

Columns 1 and 2 of table~\ref{table-cpu-gpu-comparison} plots the results for~\textit{covertype} dataset,
when batch size is set to 128, using CPU-only cores (row 1) and 1-GPU-1-CPU-core (row 2) for first-order 
tensorflow implementations. Note that newton-type methods always use 1-GPU-1-CPU-core as the 
compute platform irrespective of any of the hyper-parameter settings. We clearly notice that the first-order 
methods takes $ \approx $ 600 seconds when GPU cores are used compared to $ \approx $
350 seconds when CPU cores are used. This can be attributed to the small batch size used for first-order
methods. Smaller batch size results in computing the gradient, a compute-intensive operation, much more
frequently compared to a large batch size. For the plots shown in table~\ref{table-cpu-gpu-comparison} 
training size for ~\textit{covertype} is set to 450,000. This means gradient is computed $ \approx $ 3516
times to complete each of the training epochs in this instance. Since the batch size is very small most
of the GPU cores are idle during every computation of the gradient resulting in low GPU occupancy (which 
is the ratio of active warps on an SM and maximum allowed warps). Also 
with each invocation of gradient computation there is CUDA kernel instantiation overhead which accumulates
as well. Because of above reasons small batch sizes yield high time per epoch for first-order methods. 

Columns 3 and 4 of table~\ref{table-cpu-gpu-comparison} plots for the results for \textit{covertype} dataset
using a large batch size, of 20\% of the dataset. Note that batch size for first-order methods is same as the 
gradient sample size for newton-type methods for these plots. We clearly notice that first-order tensorflow
methods takes $ \approx $ 55 seconds when CPU-only cores are used as the compute platform
compared to $ \approx $ 22.5 seconds when 1-GPU-1-CPU-core is used, a speedup of $2 \times$ over
CPU only compute platform. In this instance, during each epoch of first-order methods gradient is evaluated
only 5 times. Because of the large batch size, $ \approx $ 90,000 points, are processed by the GPU resulting
in higher utilization of the GPU cores (compared to the same computation using smaller batch size). 
This explains why GPU-cores yield shorter time per epoch when large
batch size are used for first-order methods.

\end{document}